\documentclass{article}

\usepackage[final]{neurips_2022}

\usepackage[utf8]{inputenc} 
\usepackage[T1]{fontenc}    
\usepackage{hyperref}       
\usepackage{url}            
\usepackage{booktabs}       
\usepackage{amsfonts}       
\usepackage{nicefrac}       
\usepackage{microtype}      
\usepackage{xcolor}         

\usepackage{microtype}
\usepackage{graphicx}
\usepackage{subfigure}
\usepackage{booktabs} 
\usepackage{wrapfig}

\usepackage{subfigure}
\usepackage{stfloats}
\usepackage{caption}
\usepackage{multirow}

\usepackage{amsmath}
\usepackage{amssymb}
\usepackage{mathtools}
\usepackage{amsthm}
\usepackage{soul}
\usepackage[linesnumbered,ruled]{algorithm2e} 
\usepackage{algorithmic}
\usepackage{float}
\title{GT-GAN: General Purpose Time Series Synthesis with Generative Adversarial Networks}

\author{%
  Jinsung Jeon \\
  Yonsei University\\
  \texttt{jjsjjs0902@yonsei.ac.kr} \\
   \And
   Jeonghak Kim \\
   Kakao Corp. \\
   \texttt{haggie.pro@kakaocorp.com} \\
   \And
   Haryong Song \\
   Linger Studio Corp. \\
   \texttt{harong@lingercorp.com} \\
   \AND
   Seunghyeon Cho \\
   Yonsei University\\
   \texttt{seunghyeoncho@yonsei.ac.kr} \\
   \And
   Noseong Park \\
   Yonsei University\\
   \texttt{noseong@yonsei.ac.kr} \\
}

\begin{document}

\maketitle

\begin{abstract}
 Time series synthesis is an important research topic in the field of deep learning, which can be used for data augmentation. Time series data types can be broadly classified into regular or irregular. However, there are no existing generative models that show good performance for both types without any model changes. Therefore, we present a general purpose model capable of synthesizing regular and irregular time series data. To our knowledge, we are the first designing a general purpose time series synthesis model, which is one of the most challenging settings for time series synthesis. To this end, we design a generative adversarial network-based method, where many related techniques are carefully integrated into a single framework, ranging from neural ordinary/controlled differential equations to continuous time-flow processes. Our method outperforms all existing methods.
\end{abstract}

\section{Introduction}
\label{Introduction}
Time series data occurs frequently in real-world applications~\citep{reinsel2003elements,fu2011review,li2018dcrnn_traffic,bing2018stgcn,wu2019graphwavenet,guo2019astgcn,bai2019STG2Seq,song2020stsgcn,huang2020lsgcn,ren2021deep,tekin2021spatio}. Among many tasks related to time series, synthesizing time series data is one of the most important tasks because real-world time series data is frequently imbalanced and/or insufficient. Since regular and irregular time series data have different characteristics, however, different model designs had been adopted for them. Therefore, existing time series synthesis work focuses on either regular or irregular time series synthesis~\citep{NEURIPS2019_c9efe5f2,alaa2021generative}. To our knowledge, there are no existing methods that work well for both types.

Regular time series means regularly sampled observations without any missing ones, and irregular times series means that some observations are missing from time to time. Irregular time series is much harder to process than regular time series. For instance, it is known that neural networks perform better after transforming time series data into its frequency domain, i.e., the Fourier transform, and some time series generative models use this approach~\citep{alaa2021generative}. However, it is not easy to observe pre-determined frequencies from highly irregular time series~\citep{DBLP:conf/nips/KidgerBASL19}. However, continuous-time models~\citep{chen2018neural,DBLP:conf/nips/KidgerMFL20,debrouwer2019gruodebayes} show good performance in processing both regular and irregular time series. By resorting to them, we propose a general purpose model that can synthesize both time series types without any model changes.

To achieve the goal, we design a sophisticated model which utilizes a diverse set of technologies, ranging from generative adversarial networks (GANs~\citep{NIPS2014_5423}), and autoencoders (AEs) to neural ordinary differential equations (NODEs~\citep{chen2018neural}),  neural controlled differential equations (NCDEs~\citep{DBLP:conf/nips/KidgerMFL20}), and continuous time-flow processes (CTFPs~\citep{DBLP:journals/corr/abs-2002-10516}), which reflects the difficulty of the problem.

\begin{figure}[t]
\centering
\includegraphics[width=0.8\columnwidth]{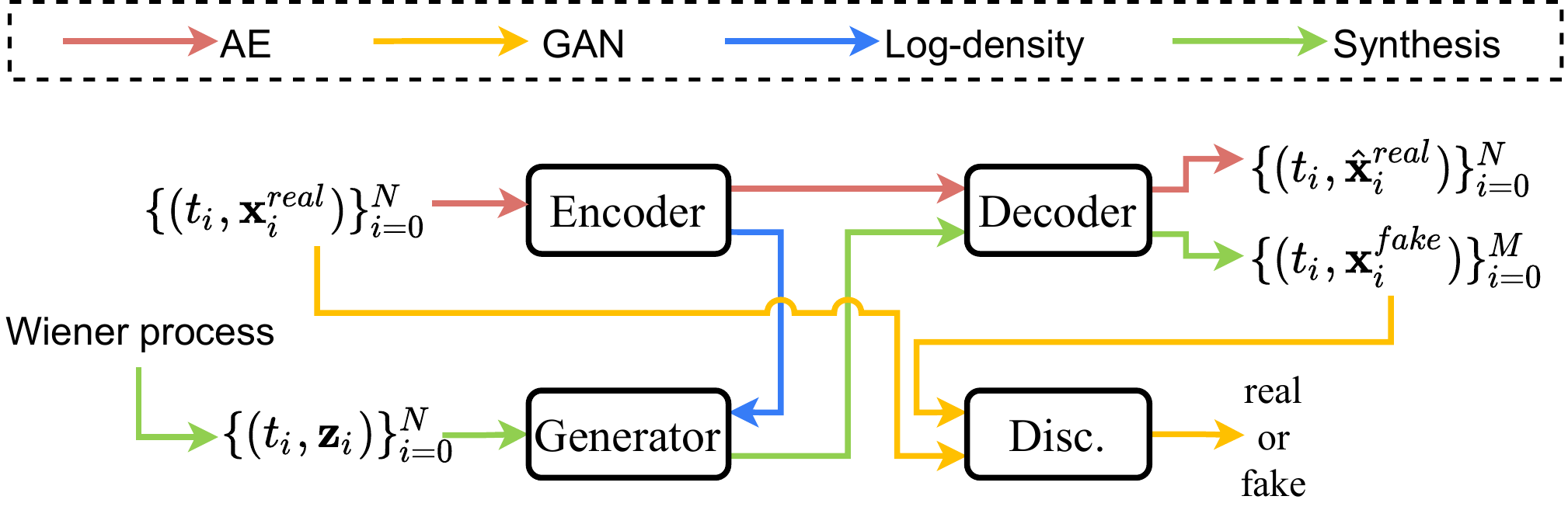}
\caption{The overall design of our method, GT-GAN. Each color means a separate workflow. The autoencoder and the GAN share the workload to synthesize  regular and irregular time series.}
\label{fig:archi}
\end{figure} 

Fig.~\ref{fig:archi} shows the overall design of our proposed method. One of the key points in our model design is that we combine the \emph{adversarial} training of GANs and the \emph{exact maximum likelihood} training of CTFPs into a single framework. However, the exact maximum likelihood training is applicable only to invertible mapping functions whose input and output sizes are the same. Therefore, we design an \emph{invertible} generator, and adopt an autoencoder, on whose hidden space our GAN performs the adversarial training.

\begin{wrapfigure}{l}{0.6\textwidth}
\vspace{-1em}
\centering
\includegraphics[width=0.6\columnwidth]{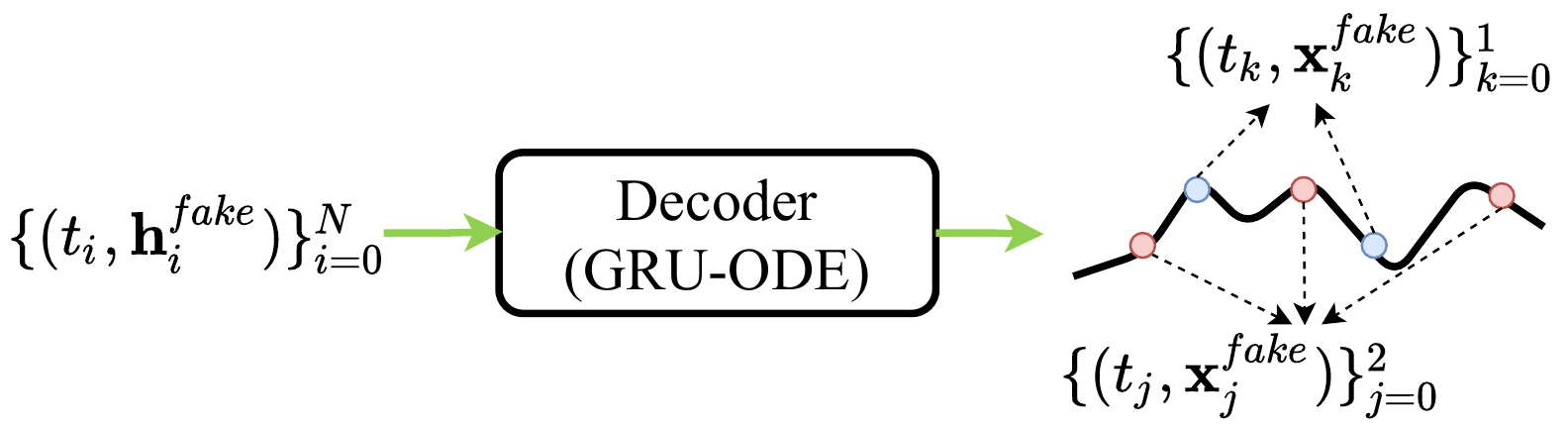}
\caption{An example of sampling two irregular time series from a fake continuous path represented by an ordinary differential equation (ODE). In other words, we solve ODE problems to sample regular/irregular time series.}
\label{fig:sampling}
\vspace{-1em}
\end{wrapfigure}

In other words, i) the hidden vector size of the encoder is the same as the noisy vector of the generator, ii) the generator produces a set of fake hidden vectors, iii) the decoder converts the set into a fake continuous path (cf. Fig.~\ref{fig:sampling}), and iv) the discriminator provides feedback after reading the sampled fake sample. We note that in the third step, a fake continuous path is created by the decoder. Therefore, we can sample any arbitrary regular/irregular time series sample from the fake path, which shows the flexibility in our method.

We conduct experiments with 4 datasets and 7 baselines. Since our method is able to support both the regular and irregular time series synthesis, we test for both of them. Our method outperforms other baselines in both environments. Our contributions can be summarized as follows:

\begin{enumerate}
    \item We design a model based on various state-of-the-art deep learning technologies. Our method is able to process any types of time series data, ranging from regular to irregular, without any model changes. 
    \item Our experimental results and visualization prove the efficacy of the proposed model.
    \item Since our task is one of the most challenging tasks for time series synthesis, the proposed model architecture is carefully designed. Our ablation studies show that our proposed model does not work well if any part is missing.
\end{enumerate}

\section{Related work and preliminaries}
GANs are one of the most representative generative technology. Ever since the first introduction in its seminal research paper, GANs have been adopted to main different domains. Recently, researchers focused on their synthesis for time series data. Therefore, there have been proposed several GANs for time series synthesis.
C-RNN-GAN~\citep{mogren2016crnngan} has a regular GAN framework that can be applied to sequential data by using LSTM in its generator and discriminator. Recurrent Conditional GAN (RCGAN~\citep{esteban2017realvalued}) took a similar approach except that its generator and discriminator take conditional input for better synthesis. WaveNet~\citep{oord2016wavenet} also generates time series data from the conditional probability of previous data by using the dilated casual convolution. WaveGAN~\citep{donahue2019adversarial} has a similar approach with DCGAN~\citep{radford2016unsupervised}, where its generator is based on WaveNet. We can modify the teacher-forcing (T-Forcing~\citep{graves2014generating}) and professor-forcing (P-Forcing~\citep{lamb2016professor}) models to generate time series data from noise vectors, although they are not GAN models, by using the forecasting characteristic of those models.

TimeGAN~\citep{NEURIPS2019_c9efe5f2} is yet another model for time series synthesis. This model aims mainly at synthesizing fake \emph{regular} time series samples. They proposed a framework where the adversarial training of GANs and the supervised training of predicting $\mathbf{x}_{i+1}$ from $\mathbf{x}_{i}$, where $\mathbf{x}_{i}$ and $\mathbf{x}_{i+1}$ mean two multivariate time series values at time $t_i$ and $t_{i+1}$, respectively.

\section{Proposed method}
\label{proposed_method}
In this section, we describe our design. Since our general purpose time series synthesis is a challenging task, the proposed design is much more complicated than other baselines. 

\begin{figure*}[t]
\centering
\includegraphics[width=\textwidth]{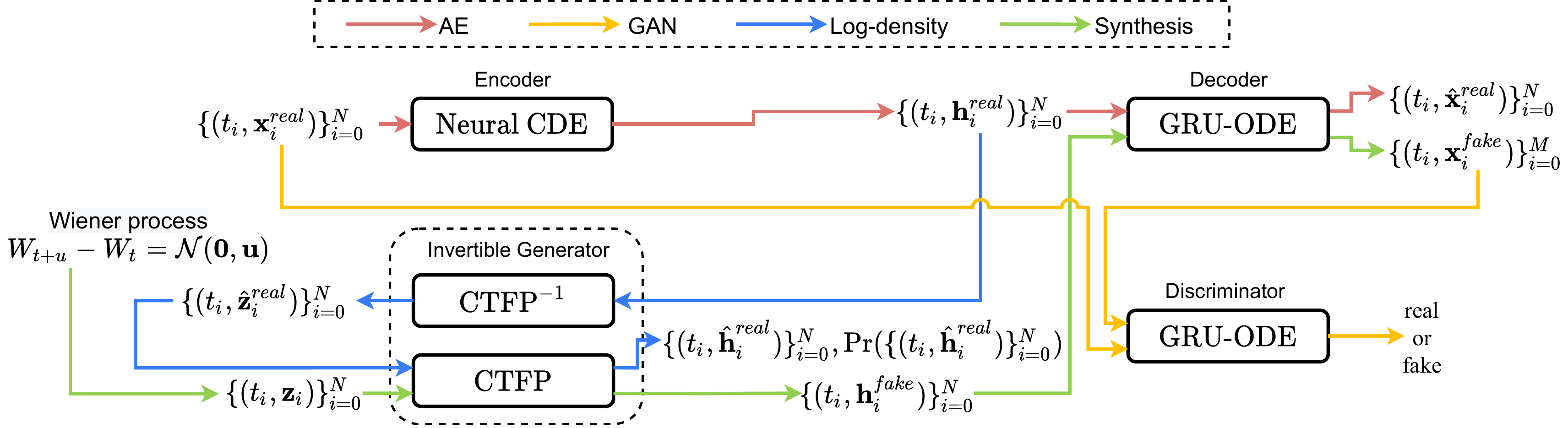}
\caption{The detailed architecture of our proposed method. Neural CDEs (or NCDEs) are a recent breakthrough for processing time series. GRU-ODEs are a continuous interpretation of gated recurrent units (GRUs) based on NODEs. CTFPs are a flow-based concept to convert an input time series process into a target process. CTFPs are not a GAN-based concept but we integrate them into our framework, considering the challenging nature of the general purpose time series synthesis.}
\vspace{-1em}
\label{fig:detail}
\end{figure*} 

\subsection{Overall workflow}
We first describe the overall workflow in our model design, which consists of several different data paths (and several different training methods based on the data paths) as follows:
\begin{enumerate}
    \item \textbf{Autoencoder path:} Given an time series sample $\{(t_i,\mathbf{x}^{real}_i)\}_{i=0}^N$, the encoder produces a set of hidden vectors $\{\mathbf{h}^{real}_i\}_{i=0}^N$. The decoder recovers a continuous path $\hat{X}^{real}$, which enhances the flexibility of our proposed method. From the path $\hat{X}^{real}$, we sample $\{(t_i,\hat{\mathbf{x}}^{real}_i)\}_{i=0}^N$. We train the encoder and the decoder using the standard autoencoder (AE) loss to match $\mathbf{x}^{real}_i$ and $\hat{\mathbf{x}}^{real}_i$ for all $i$.
    \item \textbf{Adversarial path:} Given a set of noisy vectors $\{\mathbf{z}_i\}_{i=0}^N$, our generator produces a set of fake hidden vectors $\{\mathbf{h}^{fake}_i\}_{i=0}^N$. The decoder recovers a fake continuous path $\hat{X}^{fake}$ from $\{\mathbf{h}^{fake}_i\}_{i=0}^N$. We sample $\{(t_j,\mathbf{x}^{fake}_j)\}_{j=0}^M$ from $\hat{X}^{fake}$ and feed it into the discriminator. For irregular time series synthesis, we sample $t_j$ in $[0,T]$. We train the generator, the decoder, and the discriminator using the standard adversarial loss.
    \item \textbf{Log-density path:} Given a set of hidden vectors $\{\mathbf{h}^{real}_i\}_{i=0}^N$ for an time series sample $\{(t_i,\mathbf{x}^{real}_i)\}_{i=0}^N$, the inverse path of the generator reproduces a set of noisy vectors $\{\hat{\mathbf{z}}_i\}_{i=0}^N$. We feed $\{\hat{\mathbf{z}}_i\}_{i=0}^N$ into its forward path again to reproduce $\{\hat{\mathbf{h}}^{real}_i\}_{i=0}^N$, where $\hat{\mathbf{h}}^{real}_i = \mathbf{h}^{real}_i$ for all $i$. During the forward pass, we calculate the negative log probability of $- \log p(\hat{\mathbf{h}}^{real}_i)$ for all $i$ with the change of variable theorem and minimize it for training, being inspired by~\citet{grover2018flow} and ~\citet{DBLP:journals/corr/abs-2002-10516}.
\end{enumerate}

In particular, we note that the dimensionality of the hidden space in the autoencoder is the same as that of the latent input space of the generator, i.e., $\dim(\mathbf{h}) = \dim(\mathbf{z})$. This is needed for the exact likelihood training in the generator --- the change of variable theorem requires that the input and output sizes are the same to estimate the exact likelihood. In addition to it, we let the autoencoder and the generator share the workload to synthesize fake time series by combining them into a single framework, i.e., the generator synthesizes fake hidden vectors and the decoder reproduces human-readable fake time series from them.

\subsection{Autoencoder}
\paragraph{Encoder} General NCDEs, which are considered as a continuous analogue to recurrent neural networks (RNNs), are defined as follows:
\begin{align}\label{eq:ncde}
\begin{split}
\mathbf{h}(t_{i+1}) &= \mathbf{h}(t_i) + \int_{t_i}^{t_{i+1}} f(\mathbf{h}(t);\mathbf{\theta}_f) dX(t)\\
&= \mathbf{h}(t_i) + \int_{t_i}^{t_{i+1}} f(\mathbf{h}(t);\mathbf{\theta}_f) \frac{dX(t)}{dt} dt,
\end{split}
\end{align}
where $X(t)$ is a continuous path created from a raw discrete time series sample $\{(t_i,\mathbf{x}^{real}_i)\}_{i=0}^N$ by an interpolation algorithm --- we note that $X(t_i) = (t_i, \mathbf{x}^{real}_i)$ for all $i$, and for other non-observed time-points the interpolation algorithm fills out values. Note that NCDEs keep reading the time-derivative of $X(t)$, denoted $\dot{X}(t) \stackrel{\text{def}}{=} \frac{dX(t)}{dt}$. In our case, we collect $\{\mathbf{h}^{real}_i\}_{i=0}^N$ as follows:
\begin{align}
\mathbf{h}^{real}_{i+1} &= \mathbf{h}^{real}_i + \int_{t_i}^{t_{i+1}} f(\mathbf{h}(t);\mathbf{\theta}_f) \frac{dX(t)}{dt} dt,
\end{align}where $\mathbf{h}^{real}_0 = \mathtt{FC}_{\dim(\mathbf{x}) \rightarrow \dim(\mathbf{h})}(\mathbf{x}^{real}_0)$ and $\mathtt{FC}_{input\_size \rightarrow output\_size}$ is a fully-connected layer with specific input and output sizes. We refer to Appendix~\ref{a:model} for the ODE function $f$ definition.

Therefore, the input time series $\{(t_i,\mathbf{x}^{real}_i)\}_{i=0}^N$ is represented by a set of hidden vectors $\{(t_i, \mathbf{h}^{real}_i)\}_{i=0}^N$. Because NCDEs are a continuous analogue to RNNs, it shows the best fit to processing irregular time series~\citep{DBLP:conf/nips/KidgerMFL20}.

\paragraph{Decoder} Our decoder, which reproduces a time series from its hidden representations, is based on GRU-ODEs~\citep{debrouwer2019gruodebayes} and is defined as follows:
\begin{align}
\bar{\mathbf{d}}(t_{i+1}) &= \mathbf{d}(t_i) + \int_{t_i}^{t_{i+1}} g(\mathbf{d}(t),t;\mathbf{\theta}_g) dt,\\
\mathbf{d}(t_{i+1}) &= \mathtt{GRU}(\mathbf{h}_{i+1}, \bar{\mathbf{d}}(t_{i+1})),\label{eq:dec:jump}\\
(t_{i+1}, \hat{\mathbf{x}}_{i+1}) &= (t_{i+1}, \mathtt{FC}_{\dim(\mathbf{d}) \rightarrow \dim(\mathbf{x})}(\mathbf{d}(t_{i+1}))),
\end{align}where $\mathbf{d}(t_0) = \mathtt{FC}_{\dim(\mathbf{h}) \rightarrow \dim(\mathbf{d})}(\mathbf{h}_0)$ and $\mathbf{h}_i$ means either the $i$-th real or fake hidden vector, i.e., $\mathbf{h}^{real}_i$ or $\mathbf{h}^{fake}_i$ --- recall that in Fig.~\ref{fig:detail}, the decoder is involved in both the autoencoder and the synthesis processes. $\hat{\mathbf{x}}$ means a reproduced copy of $\mathbf{x}$. GRU-ODEs uses the technology called neural ordinary differential equations (NODEs) to \emph{continuously} interpret GRUs and we refer to Appendix~\ref{a:model} for the ODE function $g$ definition.

In particular, the gated recurrent unit (GRU) at Eq.~\eqref{eq:dec:jump} is called as \emph{jump} which is known to be effective in processing time series with NODEs~\citep{debrouwer2019gruodebayes,NIPS2019_9177}. We train the encoder-decoder using the standard reconstruction loss between $\mathbf{x}^{real}_i$ and $\hat{\mathbf{x}}^{real}_i$ for all $i$ in all training time series samples.

\subsection{Generative adversarial network}

\paragraph{Generator} Whereas generators typically read a noisy vector to generate a fake sample in standard GANs, our generator reads a continuous path (or time series) sampled from a Wiener process to generate a fake time series sample --- this generation concept is known as continuous time flow processes (CTFPs~\citep{DBLP:journals/corr/abs-2002-10516}). Appendix.~\ref{a:generator} shows an example of our generation process. The input to our generation process is a random path sampled from a Wiener process, which is represented by a time series of latent vectors $\{(t_i, \mathbf{z}_i)\}_{i=0}^N$ in the path, and the output is a path of hidden vectors which is also represented by a time series of hidden vectors $\{(t_i, \mathbf{h}^{fake}_i)\}_{i=0}^N$. Therefore, our generator can be written as follows:
\begin{align}\label{eq:ctfp}
    \mathbf{h}^{fake}_{i} = \mathbf{w}_i(1) = \mathbf{w}_i(0) + \int_{0}^{1}r(\mathbf{w}_i(\tau), a_i(t), t; \mathbf{\theta}_r)d\tau,
\end{align}where $\mathbf{w}_i(0) = \mathbf{z}_i$, $a_i(0) = t_i$. Here, $\tau$ means a virtual time variable of the integral problem, and $t_i$ is a real physical time contained in a time series sample $\{(t_i, \mathbf{x}^{real}_i)\}_{i=0}^M$. We note that this design corresponds to a NODE model augmented with $a_i(t)$. We refer to Appendix~\ref{a:model} for the ODE function $r$ definition.

Owing to the invertible nature of NODEs, we can calculate the exact log-density of $\mathbf{h}^{real}_i$, i.e., the probability that $\mathbf{h}^{real}_i$ is generated by the generator, using the change of variable theorem and the Hutchinson's stochastic trace estimator as follows~\citep{grathwohl2019ffjord,DBLP:journals/corr/abs-2002-10516}:
\begin{align}
\hat{\mathbf{w}}(0) &= \mathbf{h}^{real}_i + \int_{1}^{0}r(\mathbf{w}(\tau), a_i(\tau), t; \mathbf{\theta}_r)d\tau,\label{eq:inv}\\
\begin{split}
    \log\Pr(\hat{\mathbf{h}}^{real}_i) &= \log\Pr(\hat{\mathbf{w}}(0)) \\
    &+ \int_{0}^{1}tr\Big( \frac{\partial r(\mathbf{w}(\tau), a_i(\tau), t; \mathbf{\theta}_r)}{\partial \mathbf{w}(\tau)} \Big)d\tau, \label{eq:for}
\end{split}
\end{align}where $\hat{\mathbf{w}}(0)$ means $\hat{\mathbf{z}}^{real}_i$ in Fig.~\ref{fig:detail}. $\hat{\mathbf{h}}^{real}_i$ means a reproduced copy of $\mathbf{h}^{real}_i$ by our generator. Eq.~\eqref{eq:inv} corresponds to ``$\textrm{CTFP}^{-1}$'', and Eqs.~\eqref{eq:ctfp} and~\eqref{eq:for} to ``$\textrm{CTFP}$'' in Fig.~\ref{fig:detail}. We note that in Eq.~\eqref{eq:inv}, the integral time is reversed to solve the reverse-mode integral problem.

Therefore, we minimize the negative log-density, denoted $-\log\Pr(\hat{\mathbf{h}}^{real}_i)$, for each $t_i$, and our generator is trained by the two different training paradigms: i) the adversarial training against the discriminator, and ii) the maximum likelihood estimator (MLE) training with the log-density.

\paragraph{Discriminator} We design our discriminator based on the GRU-ODE technology as follows:
\begin{align}
\bar{\mathbf{c}}(t_{i+1}) &= \mathbf{c}(t_i) + \int_{t_i}^{t_{i+1}} q(\mathbf{c}(t),t;\mathbf{\theta}_q) dt,\\
\mathbf{c}(t_{i+1}) &= \mathtt{GRU}(\mathbf{x}_{i+1}, \bar{\mathbf{c}}(t_{i+1})),\label{eq:disc:jump}
\end{align}where $\mathbf{c}(t_0) = \mathtt{FC}_{\dim(\mathbf{x}) \rightarrow \dim(\mathbf{c})}(\mathbf{x}_0)$, and $\mathbf{x}_i$ means the $i$-th time series value, i.e., $\mathbf{x}^{real}_i$ or $\mathbf{x}^{fake}_i$. The ODE function $q$ has the same architecture as $g$ but with its own parameters $\mathbf{\theta}_q$. After that, we calculate the real or fake classification $\mathbf{y} = \sigma(\mathtt{FC}_{\dim(\mathbf{c}) \rightarrow 2}(\mathbf{c}(t_{N})))$, where $\sigma$ is a softmax activation.

The role of each part of our proposed model is in Appendix~\ref{a:role}.

\subsection{Training method}
We use the mean squared reconstruction loss, i.e., the mean of $\|\mathbf{x}^{real}_i - \hat{\mathbf{x}}^{real}_i\|^2_2$ for all $i$, to train the encoder-decoder architecture. Then, we use the standard GAN loss to train the generator and the discriminator. In our preliminary experiments, we found that the original GAN loss is suitable for our task. Instead of other variations, such as WGAN-GP~\citep{gulrajani2017improved}, therefore, we use the standard GAN loss. We train our model in the following sequence:


\begin{enumerate}
    \item We pre-train the encoder-decoder networks the reconstruction loss for $K_{AE}$ iterations.
    \item After the above pre-training step, we start to jointly train all networks in the following sequence for $K_{JOINT}$ iterations: i) training the encoder-decoder networks with the reconstruction loss, ii) training the discriminator-generator networks with the GAN loss, iii) training the decoder to improve the discriminator's classification output with the discriminator loss and iv) the generator with the MLE loss every $P_{MLE}$ iteration. We found that too frequent MLE training incurs mode-collapse so we use it every $P_{MLE}$ iteration.
\end{enumerate}

In particular, the 2-ii step to train the decoder to help the discriminator out is one additional point where the autoencoder and the GAN are integrated into a single framework. In other words, the generator should deceive both the decoder and the discriminator. Our training algorithm refer to Appendix~\ref{alg:train}

The well-posedness\footnote{A well-posed problem means i) its solution uniquely exists, and ii) its solution continuously changes as input data changes.} of NCDEs and GRU-ODEs was already proved in \citet[Theorem 1.3]{lyons2007differential} and~\citet{debrouwer2019gruodebayes} under the mild condition of the Lipschitz continuity. We show that our NCDE layers are also well-posed problems. Almost all activations, such as ReLU, Leaky ReLU, SoftPlus, Tanh, Sigmoid, ArcTan, and Softsign, have a Lipschitz constant of 1. Other common neural network layers, such as dropout, batch normalization and other pooling methods, have explicit Lipschitz constant values. Therefore, the Lipschitz continuity of ODE/CDE functions can be fulfilled in our case. In other words, it is a well-posed training problem. As a result, our training algorithm solves a well-posed problem so its training process is stable in practice.

\section{Experimental evaluations}
\label{Experimental}
Our software and hardware environments are as follows: \textsc{Ubuntu} 18.04 LTS, \textsc{Python} 3.8.10, \textsc{Pytorch} 1.8.1, \textsc{Tensorflow} 2.5.0, \textsc{CUDA} 11.2, and \textsc{NVIDIA} Driver 417.22, i9 CPU, and \textsc{NVIDIA RTX 3090}. The mean and variance of 10 runs are reported for model evaluation.
\subsection{Experimental environments}

\paragraph{Datasets} We conduct experiments with 2 simulated and 2 real-world datasets. Sines has 5 features where each feature is created with different frequencies and phases independently. For each feature, $i\in\{1,...,5\}$,   $x_i(t)=sin(2\pi f_it+\theta_i)$, where$ f_i\sim \mathcal{U}[0,1]$ and $\theta_i\sim\mathcal{U}[-\pi,\pi]$. MuJoCo is multivariate physics simulation time series data with 14 features. Stocks is the Google stock price data from 2004 to 2019. Each observation represents one day and has 6 features. Energy is a UCI appliance energy prediction dataset with 28 values. To create the challenging irregular environments, 30, 50, 70\% of observations from each time series sample $\{(t_i,\mathbf{x}^{real}_i)\}_{i=0}^N$ is randomly dropped --- in other words, $N$ decreases to $0.7N, 0.5N, 0.3N$. Dropping random values has been mainly used to create irregular time series environments in the literature ~\citep{DBLP:conf/nips/KidgerBASL19,xu2020conformal,huang2020learning,tang2020joint,zhang2021graph,jhin2021attentive,DBLP:journals/corr/abs-2106-15580}. Therefore, we conduct experiments with both the regular and the irregular environments.

\paragraph{Baselines} We consider the following baselines for the regular time series experiments: TimeGAN, RCGAN, C-RNN-GAN, WaveGAN, WaveNet, T-Forcing, and P-Forcing. For the irregular experiments, we exclude WaveGAN and WaveNet, which cannot handle irregular time series, and redesign other baselines by replacing their GRU with GRU-$\bigtriangleup t$ and GRU-Decay (GRU-D)~\citep{che2018recurrent}. GRU-$\bigtriangleup t$ and GRU-D are effective models for processing irregular time series data. GRU-$\bigtriangleup t$ additionally uses the time difference between observations as input. GRU-D is a modification of GRU-$\bigtriangleup t$ to learnt exponential decays between observations. TimeGAN-$\bigtriangleup t$, RCGAN-$\bigtriangleup t$, C-RNN-GAN-$\bigtriangleup t$, T-Forcing-$\bigtriangleup t$, and P-Forcing-$\bigtriangleup t$ (resp. TimeGAN-D, RCGAN-D, C-RNN-GAN-D, T-Forcing-D, and P-Forcing-Decay) are modified with GRU-$\bigtriangleup t$ (resp. GRU-D) and can handle irregular data.

Our ablation studies also involve many advanced methods, based on NODEs, VAEs, flow models, and so forth. We intentionally leave these advanced methods for our ablation studies since our proposed method internally has them as sub-parts.

\paragraph{Evaluation metrics} For quantitative evaluation of synthesized data, it is evaluated with the discriminative score and the predictive score used in TimeGAN~\citep{NEURIPS2019_c9efe5f2}. The discriminative score measures the similarity between the original data and the synthesized data. After learning a model that classifies the original data and the synthesized data using a neural network, it is tested whether the original data and the synthesized data are classified well. The discriminative score is $|$Accuracy-0.5$|$, and if the score is low, classification is difficult, so the original data and the synthesized data are decided to be similar. The predictive score measures the effectiveness of the synthesized data using the train-synthesis-and-test-real (TSTR) method. After training a model that predicts the next step using the synthesized data, the mean absolute error (MAE) is calculated between the predicted values and the ground-truth values in test data. If the MAE is small, the model trained using the synthesized data is decoded to be similar to the original data. For qualitative evaluation, the synthetic data is visualized with the original data. There are two methods for visualization. One is to project original and synthetic data in a two dimensional space using t-SNE~\citep{van2008visualizing}. The other one is the kernel density estimation to draw data distributions.

\subsection{Experimental results}

\begin{table*}[t]
\begin{minipage}{.5\linewidth}
\centering
\scriptsize
\setlength{\tabcolsep}{1pt}
\captionof{table}{\label{tab:regular}Regular time series}
\begin{tabular}{cccccc}
\hline
 & Method    & Sines & Stocks & Energy & MuJoCo\\ \hline
\multirow{8}{*}{\rotatebox[origin=c]{90}{Discriminative Score}} & GT-GAN       &   .012$\pm$.014  &\textbf{.077$\pm$.031}       &   \textbf{.221$\pm$.068}  &  \textbf{.245$\pm$.029}     \\   \cline{2-6}
& TimeGAN   & \textbf{.011$\pm$.008}  & .102$\pm$.021   &  .236$\pm$.012  &  .409$\pm$.028   \\ 
& RCGAN     & .022$\pm$.008  & .196$\pm$.027   & .336$\pm$.017    & .436$\pm$.012     \\ 
& C-RNN-GAN & .229$\pm$.040  & .399$\pm$.028   &.499$\pm$.001    &   .412$\pm$.095   \\  
& T-Forcing   & .495$\pm$.001  & .226$\pm$.035   &.483$\pm$.004    &      .499$\pm$.000 \\ 
& P-Forcing   & .430$\pm$.227  & .257$\pm$.026   &.412$\pm$.006    &   .500$\pm$.000   \\
& WaveNet   & .158$\pm$.011  & .232$\pm$.028   &.397$\pm$.010    &      .385$\pm$.025 \\ 
& WaveGAN   & .277$\pm$.013  & .217$\pm$.022   &.363$\pm$.012    &   .357$\pm$.017   \\ \hline \hline
\multirow{9}{*}{\rotatebox[origin=c]{90}{Predictive Score}}           
&GT-GAN     & .097$\pm$.000    &  .040$\pm$.000     & .312$\pm$.002    &  \textbf{.055$\pm$.000}    \\ \cline{2-6}
& TimeGAN   & \textbf{.093$\pm$.019}  & .038$\pm$.001   &\textbf{.273$\pm$.004}    & .082$\pm$.006     \\ 
& RCGAN     & .097$\pm$.001  & .040$\pm$.001  &  .292$\pm$.005  &  .081$\pm$.003     \\ 
& C-RNN-GAN & .127$\pm$.004  & \textbf{.038$\pm$.000}   &.483$\pm$.005    &  .055$\pm$.004 \\
& T-Forcing   & .150$\pm$.022  & .038$\pm$.001   &.315$\pm$.005    &      .142$\pm$.014 \\ 
& P-Forcing   & .116$\pm$.004  & .043$\pm$.001  &.303$\pm$.006    &    .102$\pm$.013  \\  
& WaveNet   & .117$\pm$.008  & .042$\pm$.001   &.311$\pm$.005    &      .333$\pm$.004 \\ 
& WaveGAN   & .134$\pm$.013  & .041$\pm$.001   &.307$\pm$.007    &   .324$\pm$.006   \\ \cline{2-6}
& Original  & .094$\pm$.001  & .036$\pm$.001  & .250$\pm$.003   &  .031$\pm$.003    \\ \hline
\end{tabular}
\end{minipage}
\begin{minipage}{.5\linewidth}
\centering
\scriptsize
\setlength{\tabcolsep}{1pt}
\captionof{table}{\label{tab:irregular30}Irregular time series (30\% dropped)}
\begin{tabular}{ccccccc}
\hline
 &  \multicolumn{2}{c}{Method}   & Sines & Stocks & Energy & MuJoCo \\ \hline
\multirow{11}{*}{\rotatebox[origin=c]{90}{Discriminative Score}} 
& \multicolumn{2}{c}{GT-GAN}      &  \textbf{.363$\pm$.063}   & \textbf{.251$\pm$.097}   & \textbf{.333$\pm$.063}   &  \textbf{.249$\pm$.035}    \\   \cline{2-7} 
&\multirow{5}{*}  & TimeGAN-$\bigtriangleup t$   & .494$\pm$.012 & .463$\pm$.020& .448$\pm$.027   & .471$\pm$.016      \\ 
& & RCGAN-$\bigtriangleup t$ & .499$\pm$.000 &  .436$\pm$.064 & .500$\pm$.000 & .500$\pm$.000  \\ 
& & C-RNN-GAN -$\bigtriangleup t$  & .500$\pm$.000  & .500$\pm$.001  &   .500$\pm$.000  & .500$\pm$.000     \\ 
& & T-Forcing-$\bigtriangleup t$     &  .395$\pm$.063 & .305$\pm$.002  &  .477$\pm$.011   & .348$\pm$.041     \\ 
& & P-Forcing-$\bigtriangleup t$     & .344$\pm$.127  & .341$\pm$.035    &  .500$\pm$.000   &    .493$\pm$.010   \\ \cline{2-7} 
&\multirow{5}{*} &TimeGAN-D  &.496$\pm$.008  & .411$\pm$.040 & .479$\pm$.010   & .463$\pm$.025      \\ 
& &RCGAN-D  & .500$\pm$.000 & .500$\pm$.000  & .500$\pm$.000    & .500$\pm$.000     \\ 
& &C-RNN-GAN-D &  .500$\pm$.000 & .500$\pm$.000  & .500$\pm$.000    & .500$\pm$.000     \\ 
& &T-Forcing-D   & .408$\pm$.087 & .409$\pm$.051  & .347$\pm$.046   &  .494$\pm$.004    \\ 
& &P-Forcing-D   & .500$\pm$.000  &  .480$\pm$.060   &   .491$\pm$.020  & .500$\pm$.000      \\ \hline \hline
\multirow{11}{*}{\rotatebox[origin=c]{90}{Predictive Score}}
& \multicolumn{2}{c}{GT-GAN}        &  \textbf{.099$\pm$.004}   &   \textbf{.021$\pm$.003}    & \textbf{.066$\pm$.001} & \textbf{.048$\pm$.001}     \\ \cline{2-7} 
&\multirow{5}{*} & TimeGAN-$\bigtriangleup t$   & .145$\pm$.025 &  .087$\pm$.001 & .375$\pm$.011 & .118$\pm$.032\\ 
& & RCGAN-$\bigtriangleup t$   & .144$\pm$.028 & .181$\pm$.014 & .351$\pm$.056 & .433$\pm$.021 \\ 
& & C-RNN-GAN-$\bigtriangleup t$  & .754$\pm$.000  & .091$\pm$.007 & .500$\pm$.000 & .447$\pm$.000   \\
& & T-Forcing-$\bigtriangleup t$  & .116$\pm$.002 & .070$\pm$.013 & .251$\pm$.000 &  .056$\pm$.001 \\ 
& & P-Forcing-$\bigtriangleup t$    &.102$\pm$.002 & .083$\pm$.018 & .255$\pm$.001& .089$\pm$.011   \\ \cline{2-7}
&\multirow{5}{*} & TimeGAN-D   & .192$\pm$.082 & .105$\pm$.053   & .248$\pm$.024 & .098$\pm$.006\\ 
& & RCGAN-D    & .388$\pm$.113  & .523$\pm$.020& .409$\pm$.020   & .361$\pm$.073    \\ 
& & C-RNN-GAN-D &  .664$\pm$.001 & .345$\pm$.002& .440$\pm$.000  &  .457$\pm$.001    \\
& & T-Forcing-D  & .100$\pm$.002  & .027$\pm$.002  &  .090$\pm$.001  & .100$\pm$.001 \\ 
& & P-Forcing-D   & .154$\pm$.004 & .079$\pm$.008  & .147$\pm$.001 &  .173$\pm$.002    \\ \cline{2-7} 
& \multicolumn{2}{c}{Original}  & .071$\pm$.004  & .011$\pm$.002  & .045$\pm$.001   &  .041$\pm$.002   \\ \hline
\end{tabular}
\end{minipage}
\end{table*}

\begin{figure*}[t]
    \centering
    {{\includegraphics[width=0.24\textwidth]{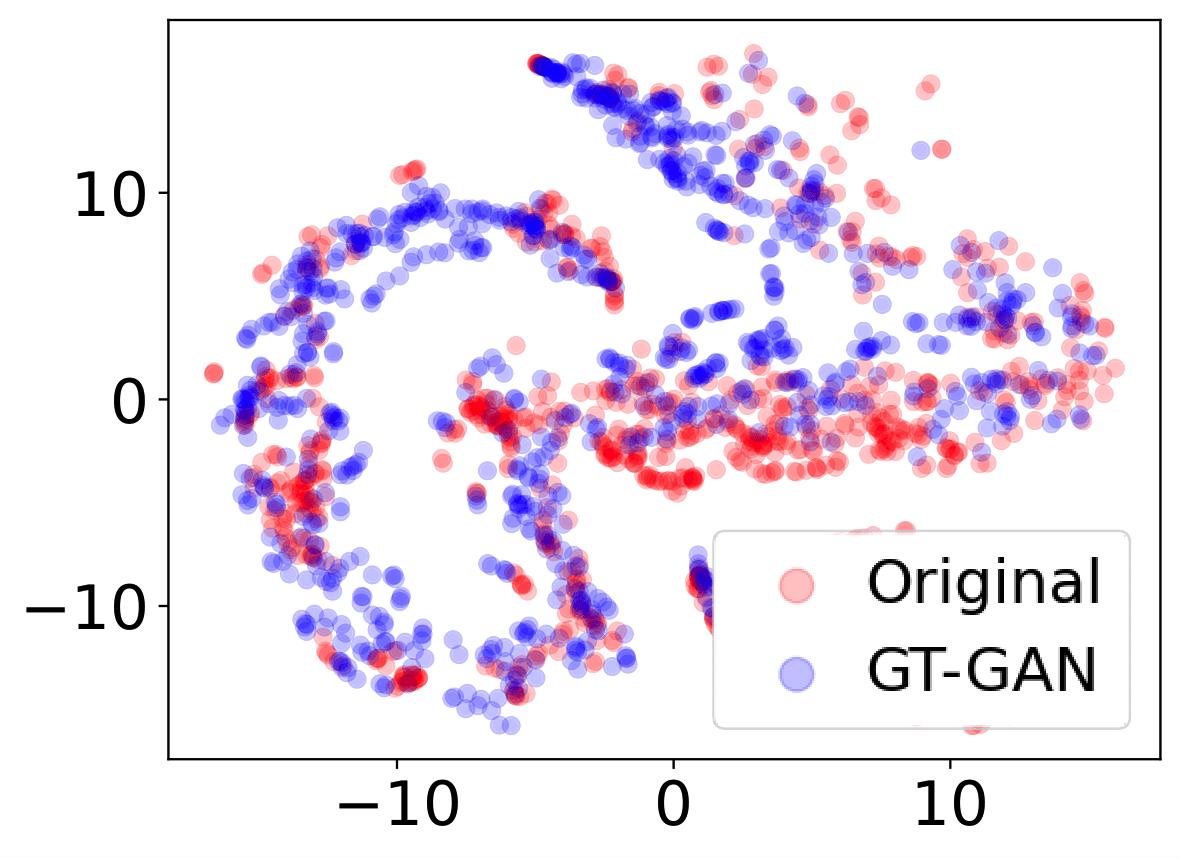}}{\includegraphics[width=0.24\textwidth]{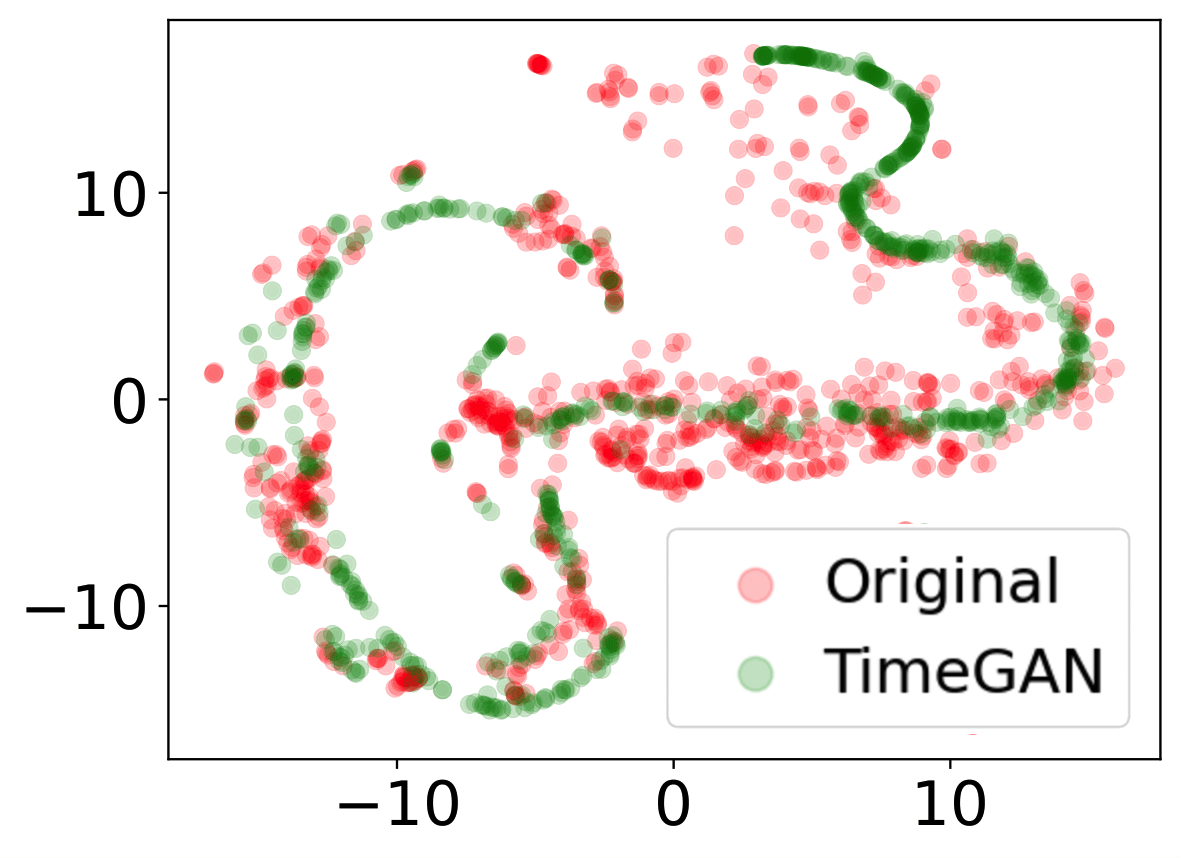}}\hfill\hspace*{-2mm}
    {\includegraphics[width=0.24\textwidth]{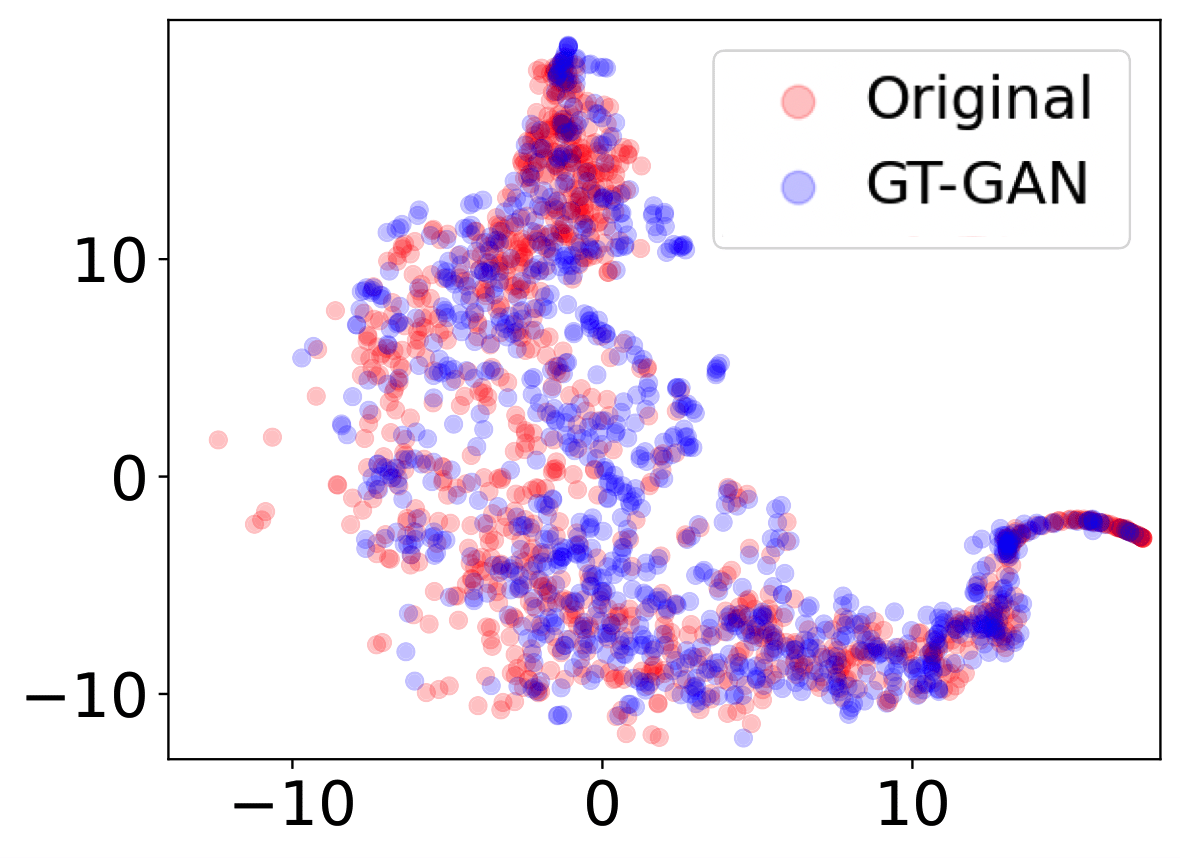}}\hspace*{2mm}{\includegraphics[width=0.24\textwidth]{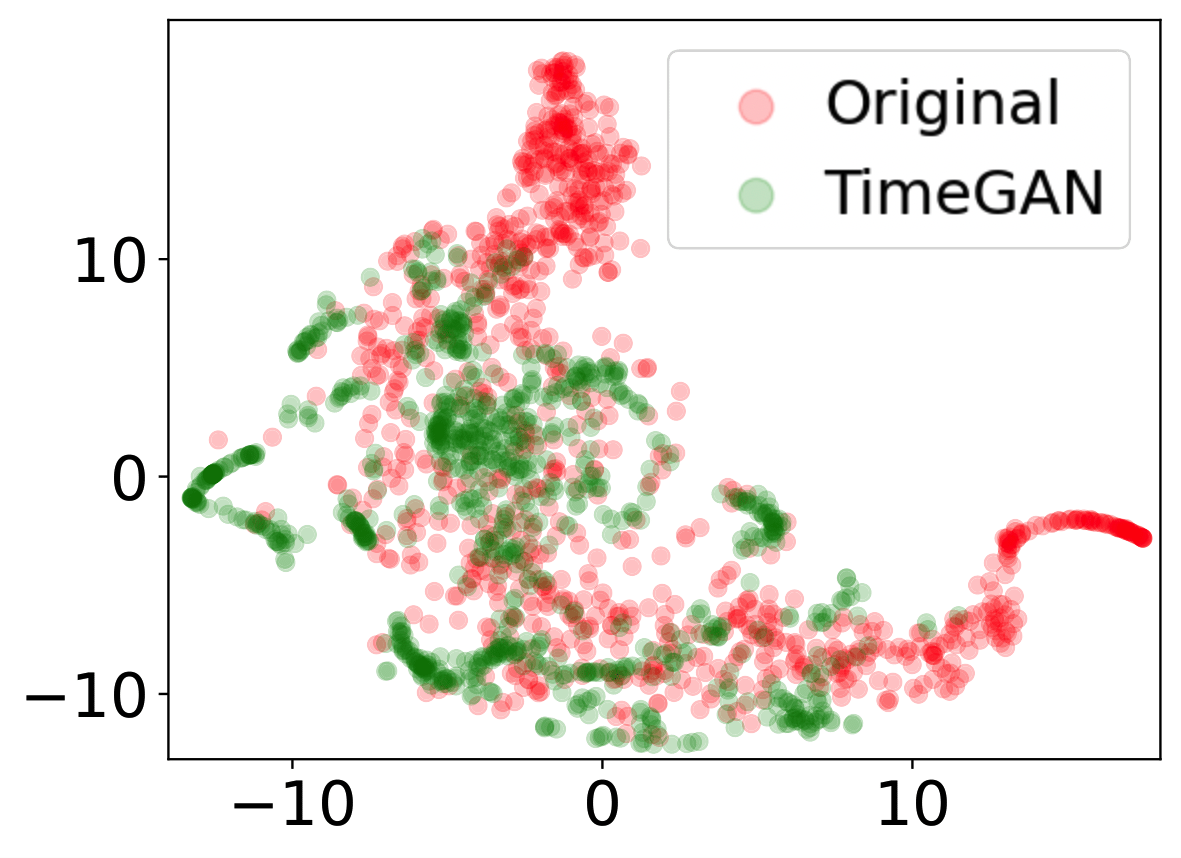}}}\\
    \subfigure[Stocks]{{\includegraphics[width=0.24\textwidth]{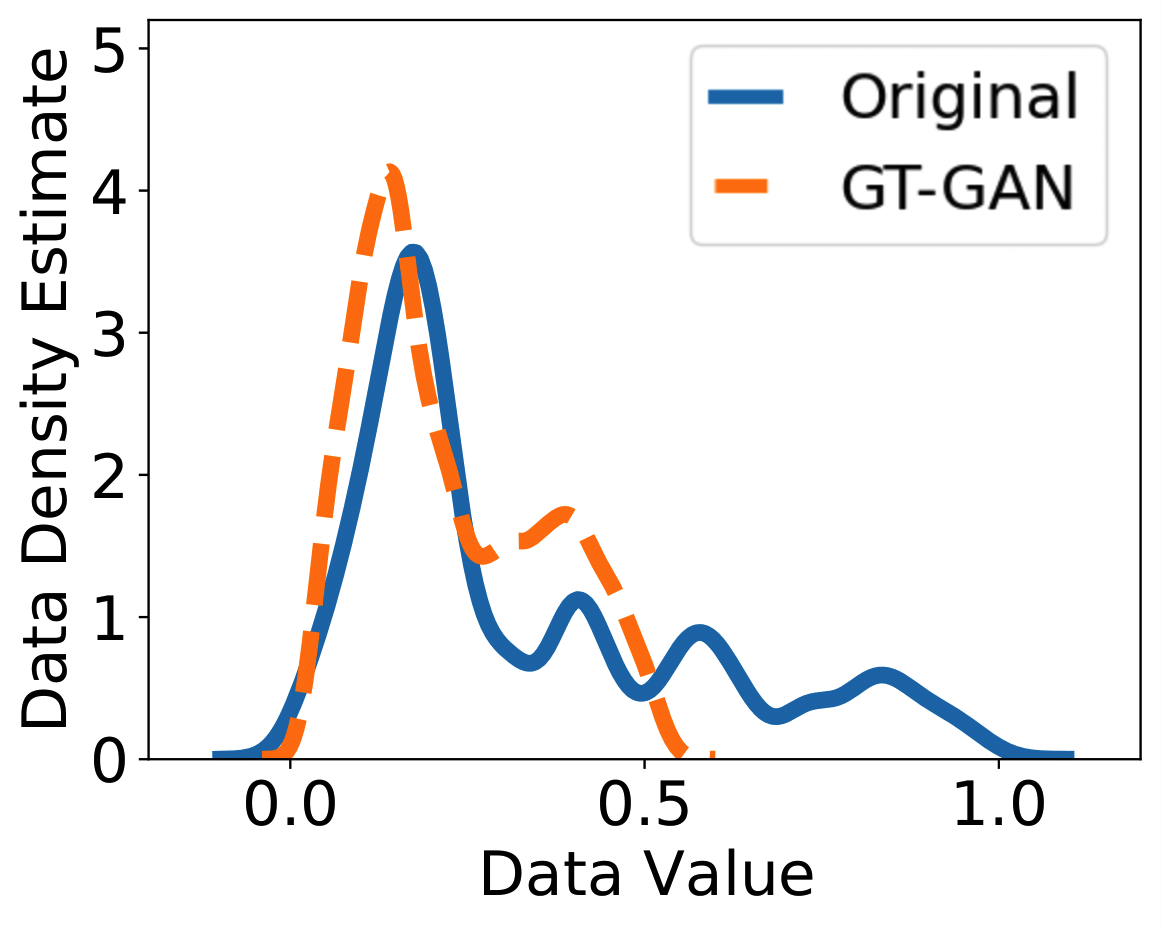}}{\includegraphics[width=0.24\textwidth]{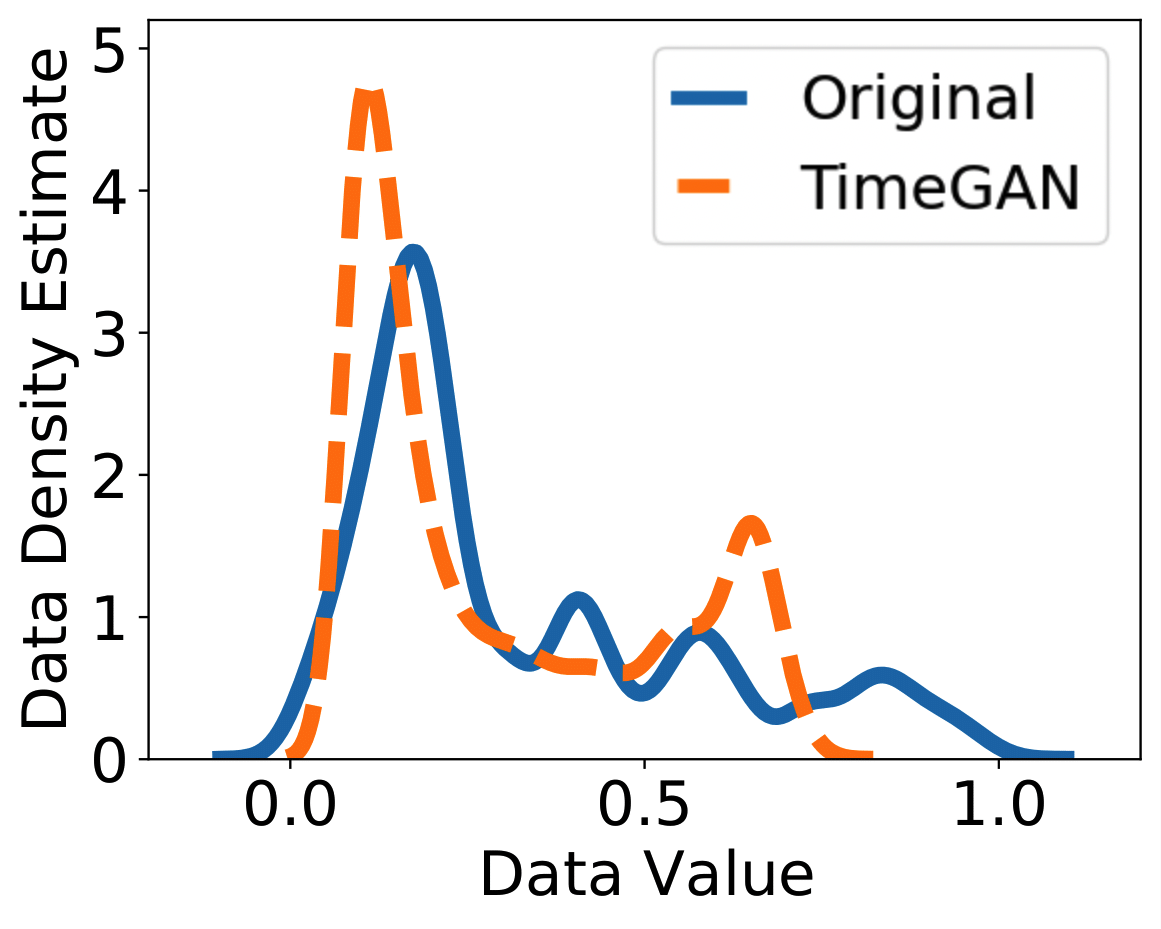}}}\hfill
    \subfigure[Energy]{{\includegraphics[width=0.25\textwidth]{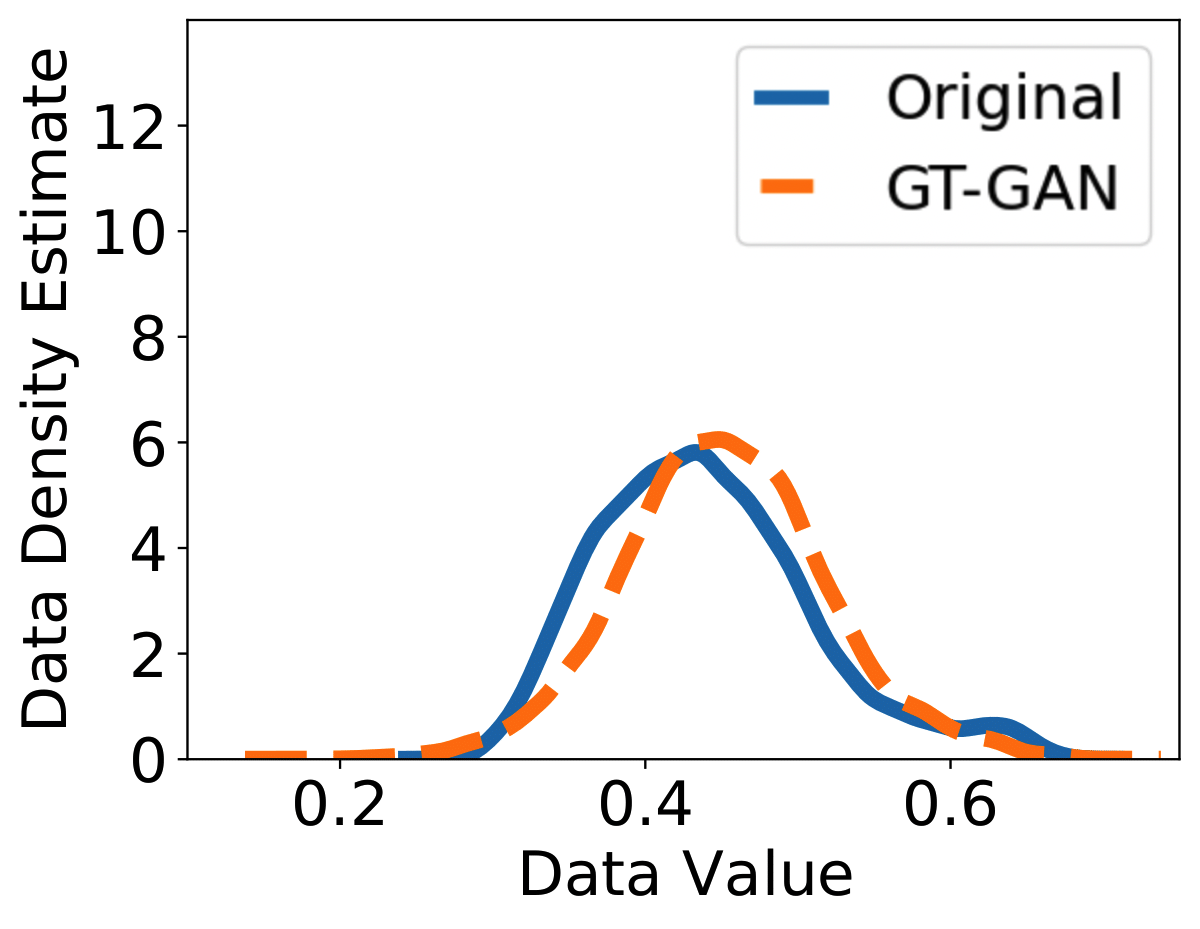}}{\includegraphics[width=0.25\textwidth]{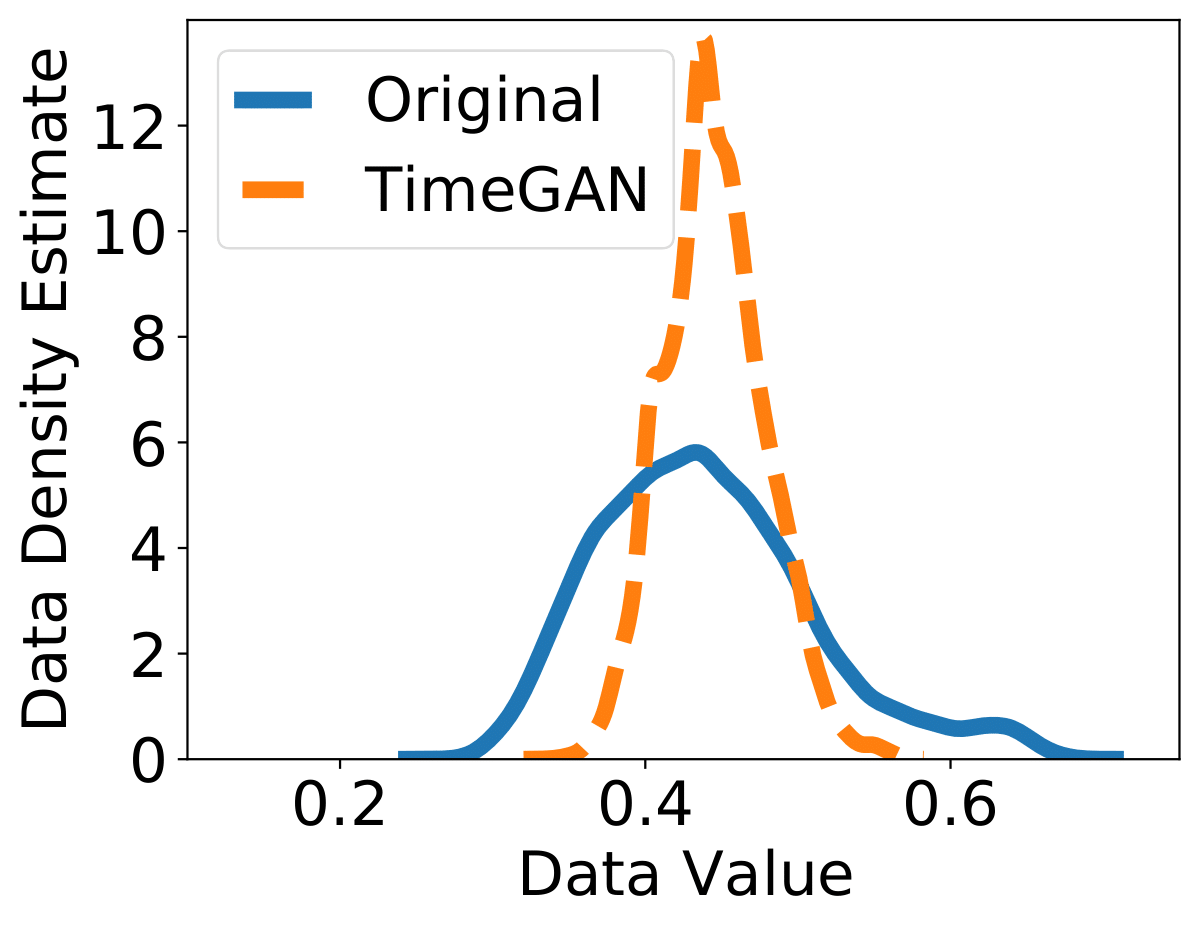}}}\hfill
    \caption{Visualizations and distributions of the regular time series synthesized by GT-GAN and TimeGAN}
    \label{fig:main_visualization_regular}
    \vspace{-1em}
\end{figure*}

\paragraph{Regular time series synthesis}
In Table~\ref{tab:regular}, we list the results of the regular time series synthesis. GT-GAN shows better performance on most cases than TimeGAN, the previous state-of-the-art model. As shown in the 1$^{\text{st}}$ row in Fig.~\ref{fig:main_visualization_regular}, GT-GAN covers original data areas better than TimeGAN. In addition, the 2$^{\text{nd}}$ row in Fig.~\ref{fig:main_visualization_regular} is the distributions of the fake data generated by GT-GAN and TimeGAN. The synthesized data's distributions from GT-GAN are more similar to those of the original data than TimeGAN, which shows the efficacy of the \emph{explicit} likelihood training of GT-GAN against the \emph{implicit} likelihood training of TimeGAN.

\begin{table*}[t]
\begin{minipage}{.5\linewidth}
\centering
\scriptsize
\setlength{\tabcolsep}{1pt}
\captionof{table}{\label{tab:irregular50}Irregular time series (50\% dropped)}
\begin{tabular}{ccccccc}
\hline
 &\multicolumn{2}{c}{Method}   & Sines & Stocks & Energy & MuJoCo \\ \hline
\multirow{11}{*}{\rotatebox[origin=c]{90}{Discriminative Score}}  &\multicolumn{2}{c}{GT-GAN}   &  \textbf{ .372$\pm$.128}   & \textbf{.265$\pm$.073}  &  \textbf{.317$\pm$.010}  &  \textbf{.270$\pm$.016}    \\   \cline{2-7} 
&\multirow{5}{*}  & TimeGAN-$\bigtriangleup t$   & .496$\pm$.008 &.487$\pm$.019 & .479$\pm$.020 &  .483$\pm$.023\\ 
&& RCGAN-$\bigtriangleup t$   & .406$\pm$.165 & .478$\pm$.049 & .500$\pm$.000  & .500$\pm$.000     \\ 
&& C-RNN-GAN-$\bigtriangleup t$  & .500$\pm$.000& .500$\pm$.000 & .500$\pm$.000 & .500$\pm$.000 \\  
&& T-Forcing -$\bigtriangleup t$   & .408$\pm$.137 & .308$\pm$.010 & .478$\pm$.011&  .486$\pm$.005\\ 

&& P-Forcing-$\bigtriangleup t$    & .428$\pm$.044 & .388$\pm$.026 & .498$\pm$.005& .491$\pm$.012\\ \cline{2-7} 
&\multirow{5}{*} & TimeGAN-D   &  .500$\pm$.000  &  .477$\pm$.021  &  .473$\pm$.015  &  .500$\pm$.000    \\ 
&& RCGAN-D   & .500$\pm$.000 &  .500$\pm$.000 &  .500$\pm$.000   &  .500$\pm$.000    \\ 
&& C-RNN-GAN-D & .500$\pm$.000 &  .500$\pm$.000 &  .500$\pm$.000   &  .500$\pm$.000       \\  
&& T-Forcing-D  & .430$\pm$.101 & .407$\pm$.034  &.376$\pm$.046    & .498$\pm$.001     \\ 
&& P-Forcing-D  & .499$\pm$.000  & .500$\pm$.000 &  .500$\pm$.000  &  .500$\pm$.000 \\   \hline \hline
\multirow{11}{*}{\rotatebox[origin=c]{90}{Predictive Score}}
& \multicolumn{2}{c}{GT-GAN}     &    \textbf{.101$\pm$.010}   &    \textbf{.018$\pm$.002}  & \textbf{.064$\pm$.001} & \textbf{.056$\pm$.003}\\  \cline{2-7} 
&\multirow{5}{*} & TimeGAN-$\bigtriangleup t$   & .123$\pm$.040 & .058$\pm$.003 & .501$\pm$.008 & .402 $\pm$.021\\ 
&& RCGAN-$\bigtriangleup t$   & .142$\pm$.005 & .094$\pm$.013 & .391$\pm$.014 & .277$\pm$.061 \\ 
&& C-RNN-GAN-$\bigtriangleup t$ & .741$\pm$.026&.089$\pm$.001 & .500$\pm$.000 & .448$\pm$.001 \\
&& T-Forcing-$\bigtriangleup t$    & .379$\pm$.029 & .075$\pm$.032 & .251$\pm$.000  &  .069$\pm$.002  \\ 
&& P-Forcing-$\bigtriangleup t$  &.120$\pm$.005 & .067$\pm$.014& .263$\pm$.003& .189$\pm$.026    \\ \cline{2-7} 
&\multirow{5}{*} & TimeGAN-D   & .169$\pm$.074 & .254$\pm$.047  & .339$\pm$.029 &  .375$\pm$.011 \\ 
&& RCGAN-D     & .519$\pm$.046  & .333$\pm$.044 &  .250$\pm$.010  &    .314$\pm$.023  \\ 
&& C-RNN-GAN-D & .754$\pm$.000  & .273$\pm$.000  & .438$\pm$.000  & .479$\pm$.000   \\
&& T-Forcing-D   & .104$\pm$.001& .038$\pm$.003  &.090$\pm$.000    & .113$\pm$.001     \\ 
&& P-Forcing-D   & .190$\pm$.002& .089$\pm$.010  &  .198$\pm$.005  & .207$\pm$.008     \\ \cline{2-7} 
&\multicolumn{2}{c}{Original}  & .071$\pm$.004  & .011$\pm$.002  & .045$\pm$.001   &  .041$\pm$.002   \\ \hline
\end{tabular}
\end{minipage}
\begin{minipage}{.5\linewidth}
\centering
\scriptsize
\setlength{\tabcolsep}{1pt}
\captionof{table}{\label{tab:irregular70}Irregular time series (70\% dropped)}
\begin{tabular}{ccccccc}
\hline
 &\multicolumn{2}{c}{Method}    & Sines & Stocks & Energy & MuJoCo \\ \hline
\multirow{11}{*}{\rotatebox[origin=c]{90}{Discriminative Score}} &\multicolumn{2}{c}{GT-GAN}      &   \textbf{.278$\pm$.022}   & \textbf{.230$\pm$.053}   & \textbf{.325$\pm$.047}   &   \textbf{.275$\pm$.023}   \\   \cline{2-7} 
& \multirow{5}{*} & TimeGAN-$\bigtriangleup t$    & .500$\pm$.000 &  .488$\pm$.009 & .496$\pm$.008 & .494$\pm$.009  \\
& & RCGAN-$\bigtriangleup t$  & .433$\pm$.142 & .381$\pm$.086  &   .500$\pm$.000  & .500$\pm$.000    \\ 
& & C-RNN-GAN-$\bigtriangleup t$ & .500$\pm$.000 & .500$\pm$.000 & .500$\pm$.000 & .500$\pm$.000  \\  
& & T-Forcing-$\bigtriangleup t$   & .374$\pm$.087 & .365$\pm$.027 & .468$\pm$.008 &  .428$\pm$.022  \\ 
& & P-Forcing-$\bigtriangleup t$   & .288$\pm$.047 & .317$\pm$.019&.500$\pm$.000 & .498$\pm$.003  \\ \cline{2-7} 
&\multirow{5}{*} & TimeGAN-D & .498$\pm$.006  & .485$\pm$.022  &  .500$\pm$.000  & .492$\pm$.009     \\
& & RCGAN-D    & .500$\pm$.000 & .500$\pm$.000  & .500$\pm$.000    &  .500$\pm$.000    \\ 
& & C-RNN-GAN-D & .500$\pm$.000 &  .500$\pm$.000 &  .500$\pm$.000   &  .500$\pm$.000    \\  
& & T-Forcing-D & .436$\pm$.067 & .404$\pm$.068  &  .336$\pm$.032   & .493$\pm$.005     \\ 
& & P-Forcing-D   & .500$\pm$.000 & .449$\pm$.150   & .494$\pm$.011   & .499$\pm$.000   \\ \hline \hline
\multirow{11}{*}{\rotatebox[origin=c]{90}{Predictive Score}} & \multicolumn{2}{c}{GT-GAN}        &    \textbf{.088$\pm$.005}   &  \textbf{.020$\pm$.005}  & \textbf{.076$\pm$.001}   & \textbf{.051$\pm$.001}     \\  \cline{2-7} 
&\multirow{5}{*} & TimeGAN-$\bigtriangleup t$    & .734$\pm$.000 &  .072$\pm$.000&  .496$\pm$.000  & .442$\pm$.000 \\ 
&& RCGAN-$\bigtriangleup t$     & .218$\pm$.072  &  .155$\pm$.009 & .498$\pm$.000  &  .222$\pm$.041 \\ 
&& C-RNN-GAN-$\bigtriangleup t$ & .751$\pm$.014 & .084$\pm$.002 & .500$\pm$.000 & .448$\pm$.001\\
&& T-Forcing-$\bigtriangleup t$    & .113$\pm$.001 & .070$\pm$.022 & .251$\pm$.000 & .053$\pm$.002  \\ 
&& P-Forcing-$\bigtriangleup t$   &.123$\pm$.004 &.050$\pm$.002& .285$\pm$.006&.117$\pm$.034 \\ \cline{2-7} 
&\multirow{5}{*}& TimeGAN-D   &  .752$\pm$.001 & .228$\pm$.000 &  .443$\pm$.000 & .372$\pm$.089\\ 
&& RCGAN-D     & .404$\pm$.034  & .441$\pm$.045 & .349$\pm$.027   & .420$\pm$.056 \\ 
&& C-RNN-GAN-D &  .632$\pm$.001& .281$\pm$.019   &  .436$\pm$.000 &  .479$\pm$.001 \\
&& T-Forcing-D  &  .102$\pm$.001 & .031$\pm$.002  &  .091$\pm$.000  & .114$\pm$.003     \\ 
&& P-Forcing-D  & .278$\pm$.045 & .107$\pm$.009  &  .193$\pm$.006  & .191$\pm$.005  \\ \cline{2-7} 
& \multicolumn{2}{c}{Original}  & .071$\pm$.004  & .011$\pm$.002  & .045$\pm$.001   &  .041$\pm$.002    \\ \hline
\end{tabular}
\end{minipage}
\vspace{-0.7em}
\end{table*}

\begin{figure*}[!ht]
    \centering
    {{\includegraphics[width=0.24\textwidth]{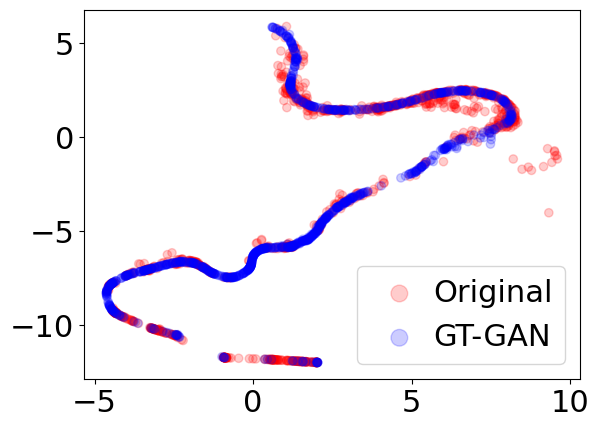}}{\includegraphics[width=0.24\textwidth]{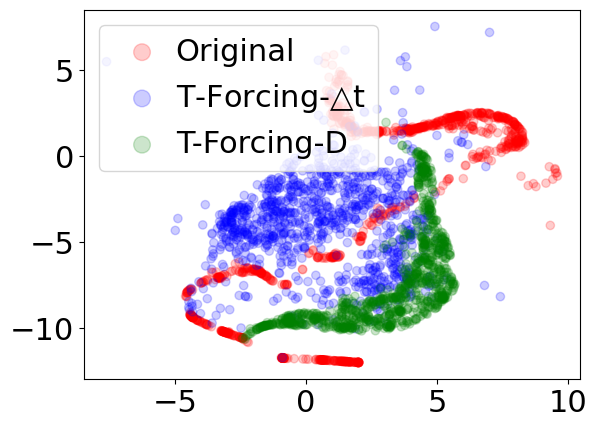}}\hfill\hspace*{-3mm}
    {\includegraphics[width=0.24\textwidth]{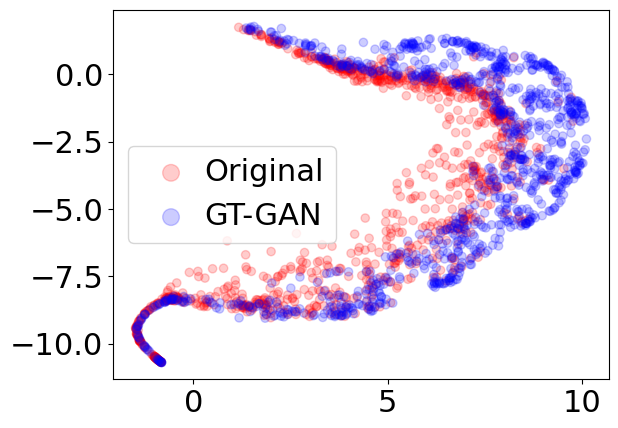}}{\includegraphics[width=0.24\textwidth]{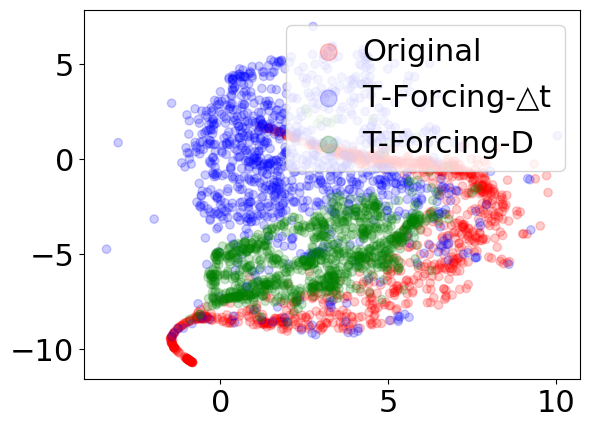}}}\\
    \subfigure[Stocks]{{\includegraphics[width=0.24\textwidth]{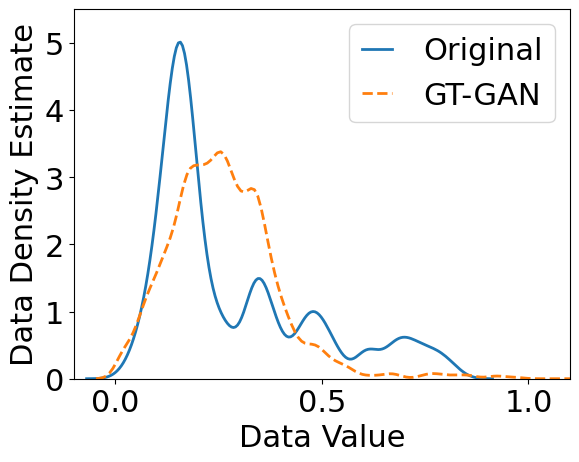}}{\includegraphics[width=0.24\textwidth]{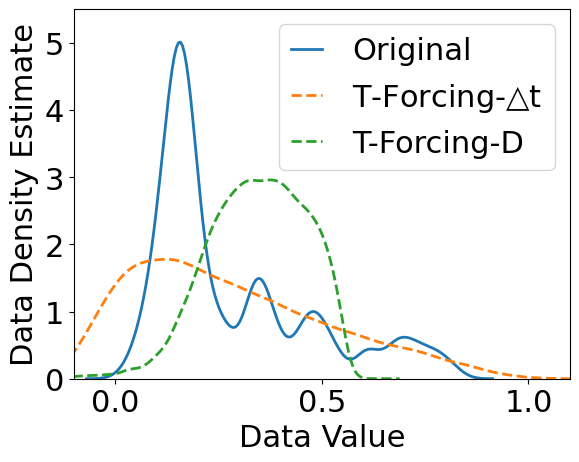}}}\hfill
    \subfigure[Energy]{{\includegraphics[width=0.24\textwidth]{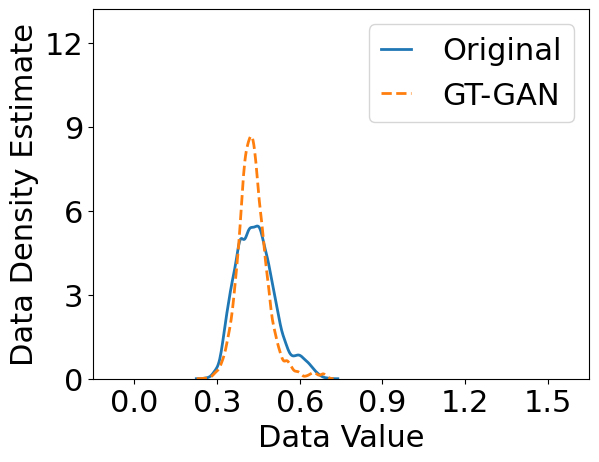}}{\includegraphics[width=0.24\textwidth]{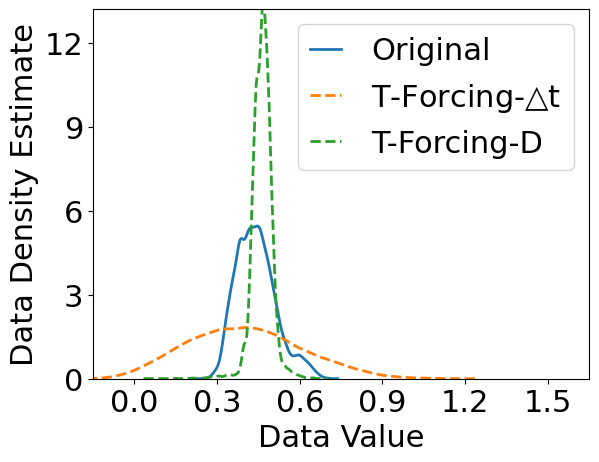}}}\hfill
    \caption{Visualizations and distributions of the irregular time series (70\% dropped) by GT-GAN, T-Forcing-$\bigtriangleup$t and T-Forcing-D}
    \label{fig:main_drop_70}
    \vspace{-1em}
\end{figure*}

\paragraph{Irregular time series synthesis}
In Tables~\ref{tab:irregular30},~\ref{tab:irregular50}, and ~\ref{tab:irregular70}, we list the results of the irregular time series synthesis. GT-GAN shows better discriminative and predictive scores than other baselines in all cases. In Table~\ref{tab:irregular30}, where we drop random 30\% of observations from each time series sample, GT-GAN shows the best outcomes, outperforming TimeGAN by large margins. Baselines modified with GRU-$\bigtriangleup t$ and those with GRU-Decay show comparable results and it is hard to say one is better than the other in this table.

In Table~\ref{tab:irregular50} (50\% dropped), many baselines do not show reasonable synthesis quality, e.g., TimeGAN-D, TimeGAN-$\bigtriangleup t$, RCGAN-D, C-RNN-GAN-D, and C-RNN-GAN-$\bigtriangleup t$ have a discriminative score of 0.5. Surprisingly, T-Forcing-D, T-Forcing-$\bigtriangleup t$, P-Forcing-D, and P-Forcing-$\bigtriangleup t$ work well in this case. However, our model clearly shows the best performance in all datasets. Baselines modified with GRU-$\bigtriangleup t$ show slightly better than them modified with GRU-Decay in this case.

\begin{wraptable}{r}{8.5cm}
\vspace{-1.2em}
\scriptsize
\centering
\caption{\label{tab:ablation}Ablation study for training options. Refer to Appendix~\ref{a:abl} for other ablation studies with irregular time series. }
\setlength{\tabcolsep}{5pt}
\begin{tabular}{cccccc}
\hline
 & Method (Regular)    &  Sines & Stocks & Energy & MuJoCo \\ \hline
\multirow{3}{*}{\rotatebox[origin=c]{90}{Disc.}} 
& GT-GAN      &   \textbf{.012}   & \textbf{.077}  &  \textbf{.221}  &  \textbf{.245}    \\  
& w/o Eq.~\eqref{eq:for}   &  .023  &  .159  &  .356  &  .278    \\ 
& w/o pre-training     & .046 &  .175 &  .312   &  .290 \\ \hline
\multirow{3}{*}{\rotatebox[origin=c]{90}{Pred.}} 
& GT-GAN     &   .097   & .040  &  .312  &  .055    \\  
& w/o Eq.~\eqref{eq:for}   &  .097  & .043  & .315  &  .057    \\ 
& w/o pre-training     & \textbf{.096} &  \textbf{.038} &  \textbf{.299}  &  \textbf{.052}\\ \hline
\end{tabular}
\vspace{-1em}
\end{wraptable}

Finally, Table~\ref{tab:irregular70} (70\% dropped) shows the results of the most challenging experiments in  our paper. All baselines do not work well because of the high dropping rate. T-Forcing-D, T-Forcing-$\bigtriangleup t$, P-Forcing-D, and P-Forcing-$\bigtriangleup t$, which showed reasonable performance with a dropping rate no larger than 50\%, do not work well in this case. This shows that they are vulnerable to highly irregular time series data. Other GAN-based baselines are vulnerable as well. Our method greatly outperforms all existing methods, e.g., a discriminative score of 0.278 by GT-GAN vs. 0.436 by T-Forcing-D vs. 0.288 by P-Forcing-$\bigtriangleup t$ for Sines, and a predictive score of 0.051 by GT-GAN vs. 0.114 by T-Forcing-D vs. 0.053 by T-Forcing-$\bigtriangleup t$ for MuJoCo. Fig~\ref{fig:main_drop_70} visually compare our method and the best performing baseline for the dropping rate of 70\% --- figures for other dropping rates and data are in Appendix~\ref{a:visualization} and similar patterns are observed in them.


\paragraph{Ablation \& sensitivity analyses}
GT-GAN is characterized by the MLE training with the negative log-density in Eq.~\eqref{eq:for}, and the pre-training step of the encoder and decoder. Table~\ref{tab:ablation} shows the results of various GT-GAN modifications with some training mechanism removed. The model using the negative log-density training shows better performance than the model not using it. That is, the MLE training makes the synthetic data more like the real data. When the pre-trained autoencoder is not used, the predictive score is better than GT-GAN. However, the discriminative score is the worst.

\begin{table}[t]
\vspace{-1em}
\caption{\label{tab:ablation_model4}Ablation study for model architecture in MuJoCo.}
\setlength{\tabcolsep}{5pt}
\scriptsize
\centering
\begin{tabular}{ccccccccc}
\hline
Energy  & \multicolumn{2}{c}{GT-GAN (w.o. AE)} & \multicolumn{2}{c}{GT-GAN (Flow only)} & \multicolumn{2}{c}{GT-GAN (AE only)} & \multicolumn{2}{c}{GT-GAN (Full model)} \\ \hline
Metric & \multicolumn{1}{c}{Disc.} & Pred. & \multicolumn{1}{c}{Disc.} & Pred. & \multicolumn{1}{c}{Disc.} & Pred. & \multicolumn{1}{c}{\;\;\;Disc.\;\;\;} & Pred. \\ \hline
30\% dropped & \multicolumn{1}{c}{.500} & .054 & \multicolumn{1}{c}{.467} & .156 & \multicolumn{1}{c}{.495} &.162 & \multicolumn{1}{c}{\textbf{.249}} & \textbf{.048} \\
50\% dropped & \multicolumn{1}{c}{.500} & .064& \multicolumn{1}{c}{.457} & .111& \multicolumn{1}{c}{.495} & .162 & \multicolumn{1}{c}{\textbf{.270}} & \textbf{.056} \\
70\% dropped & \multicolumn{1}{c}{.500} & .066 & \multicolumn{1}{c}{.455} & .107& \multicolumn{1}{c}{.496} & .146& \multicolumn{1}{c}{\textbf{.275}} & \textbf{.051} \\ \hline
\vspace{-3em}
\end{tabular}
\end{table}

\begin{wrapfigure}{r}{6.8cm}
\vspace{-2.0em}
\small
\vspace{1em}
\centering
\begin{minipage}{0.48\textwidth}
\centering
{{\includegraphics[width=0.4\textwidth]{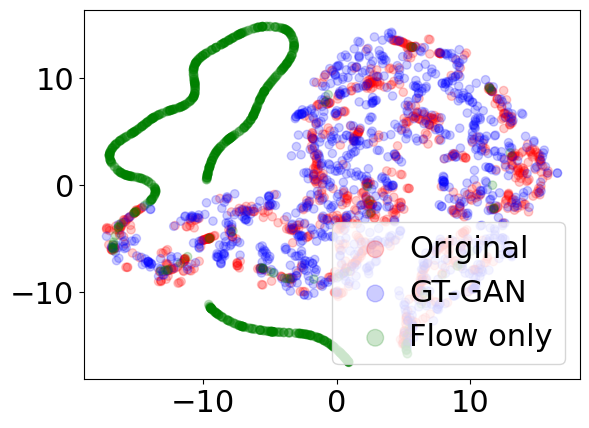}}\hspace{3em}{\includegraphics[width=0.4\textwidth]{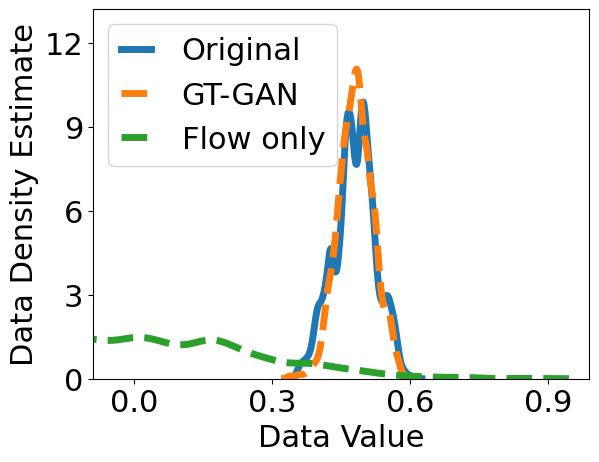}}}
\caption{Visualization and distribution of MuJoCo (70\% dropped) by GT-GAN and Flow only}
\label{fig:ablation_flow}
\end{minipage}\hfill
\vspace{-0.7em}
\end{wrapfigure}

In Table~\ref{tab:ablation_model4}, 
we alter the architecture for our model. We modify our proposed GT-GAN model by removing its sub-parts to create simpler ablation models: i) In the first ablation model, we remove the autoencoder and perform the adversarial training only with our generator and discriminator, denoted ``GT-GAN (w.o. AE)''. In other words, our generation directly outputs raw observations (instead of hidden vectors), which will be fed into our GRU-ODE-based discriminator. ii) The second ablation model, denoted ``GT-GAN (Flow only)'', has only our CTFP-based generator and we train it with the maximum likelihood training --- we note that this construction is the same as training flow-based models. This model is equivalent to the original CTFP model~\citep{DBLP:journals/corr/abs-2002-10516}. iii) The third ablation model has only the autoencoder, denoted ``GT-GAN (AE only)''. However, we convert it to a variational autoencoder (VAE) model. In the full GT-GAN model, the encoder produces a set of hidden vectors $\{(t_i,\mathbf{h}^{real}_i)\}_{i=0}^N$. In this ablation model, however, this is changed to$\{(t_i,\mathcal{N}(\mathbf{h}^{real}_i, \mathbf{1}))\}_{i=0}^N$, where $\mathcal{N}(\mathbf{h}^{real}_i, \mathbf{1})$ means the unit Gaussian centered at $\mathbf{h}^{real}_i$. The decoder is the same as its full model. We use the variational training for this model. Among the ablation models, GT-GAN (Flow only) outperforms the discriminator score in most cases. However, our full model is clearly the best in all cases. Our study shows that the ablation models of GT-GAN do not perform as well as its full model if any parts are missing, as shown in Fig~\ref{fig:ablation_flow}. Refer to Appendix~\ref{a:abl} for the ablation studies with other datasets.

The hyperparameters that significantly affect model performance are the absolute tolerance (atol), the relative tolerance (rtol), and the period of the MLE training ($P_{MLE}$) for the generator. The atol and rtol determine the error control performed by the ODE solvers in CTFPs. We test with various options of the hyperparameters in Appendix~\ref{a:sen}. We found that there is an appropriate error tolerance (atol, rtol) depending on the data input size. For example, the datasets with small input sizes (i.e., Sines, Stocks) have good discriminator scores with (1e-2, 1e-3), and the datasets with large input sizes (i.e., Energy, MuJoCo) show good results with (1e-3, 1e-2).

\section{Conclusions}
\label{discussions}
Time series synthesis is an important research topic in deep learning and had been separately studied for regular or irregular time series synthesis. However, there are still no existing generative models that can handle both regular and irregular time series without model changes. Our proposed method, GT-GAN, is based on various advanced deep learning technologies, ranging from GANs to NODEs, and NCDEs, and is able to process all possible types of time series without any changes in its model architecture and parameters. Our experiments, which incorporate various synthetic and real-world datasets, prove the efficacy of the proposed method. In our ablation studies, only our full method without any missing parts shows reasonable synthesis capabilities. The limitations and societal impacts of our proposed model are in Appendix~\ref{a:societal impacts}.

\begin{ack}
Noseong Park is the corresponding author. This work was partly supported by the Yonsei University Research Fund of 2022 (10\%), the Institute of Information \& Communications Technology Planning \& Evaluation (IITP) grant funded by the Korean government (MSIT) (No. 2020-0-01361, Artificial Intelligence Graduate School Program at Yonsei University, 10\%, and No. 2022-0-00113, Developing a Sustainable Collaborative Multi-modal Lifelong Learning Framework, 70\%), and the LG Display research fund (C2022000673, 10\%).



\end{ack}

\bibliographystyle{plainnat}
\bibliography{example_paper}

\begin{thebibliography}{39}
\providecommand{\natexlab}[1]{#1}
\providecommand{\url}[1]{\texttt{#1}}
\expandafter\ifx\csname urlstyle\endcsname\relax
  \providecommand{\doi}[1]{doi: #1}\else
  \providecommand{\doi}{doi: \begingroup \urlstyle{rm}\Url}\fi

\bibitem[Alaa et~al.(2021)Alaa, Chan, and van~der Schaar]{alaa2021generative}
Ahmed Alaa, Alex~James Chan, and Mihaela van~der Schaar.
\newblock Generative time-series modeling with fourier flows.
\newblock In \emph{ICLR}, 2021.

\bibitem[Bai et~al.(2019)Bai, Yao, Kanhere, Wang, and Sheng]{bai2019STG2Seq}
Lei Bai, Lina Yao, Salil~S. Kanhere, Xianzhi Wang, and Quan~Z. Sheng.
\newblock Stg2seq: Spatial-temporal graph to sequence model for multi-step
  passenger demand forecasting.
\newblock In \emph{Proceedings of the Twenty-Eighth International Joint
  Conference on Artificial Intelligence, {IJCAI-19}}, pages 1981--1987, 7 2019.
\newblock \doi{10.24963/ijcai.2019/274}.

\bibitem[Brouwer et~al.(2019)Brouwer, Simm, Arany, and
  Moreau]{debrouwer2019gruodebayes}
Edward~De Brouwer, Jaak Simm, Adam Arany, and Yves Moreau.
\newblock Gru-ode-bayes: Continuous modeling of sporadically-observed time
  series.
\newblock In \emph{NeurIPS}, 2019.

\bibitem[Che et~al.(2018)Che, Purushotham, Cho, Sontag, and
  Liu]{che2018recurrent}
Zhengping Che, Sanjay Purushotham, Kyunghyun Cho, David Sontag, and Yan Liu.
\newblock Recurrent neural networks for multivariate time series with missing
  values.
\newblock \emph{Scientific reports}, 8\penalty0 (1):\penalty0 1--12, 2018.

\bibitem[Chen et~al.(2018)Chen, Rubanova, Bettencourt, and
  Duvenaud]{chen2018neural}
Ricky~TQ Chen, Yulia Rubanova, Jesse Bettencourt, and David~K Duvenaud.
\newblock Neural ordinary differential equations.
\newblock \emph{Advances in neural information processing systems}, 31, 2018.

\bibitem[Deng et~al.(2020)Deng, Chang, Brubaker, Mori, and
  Lehrmann]{DBLP:journals/corr/abs-2002-10516}
Ruizhi Deng, Bo~Chang, Marcus~A. Brubaker, Greg Mori, and Andreas~M. Lehrmann.
\newblock Modeling continuous stochastic processes with dynamic normalizing
  flows.
\newblock In \emph{NeurIPS}, 2020.

\bibitem[Deng et~al.(2021)Deng, Brubaker, Mori, and
  Lehrmann]{DBLP:journals/corr/abs-2106-15580}
Ruizhi Deng, Marcus~A. Brubaker, Greg Mori, and Andreas~M. Lehrmann.
\newblock Continuous latent process flows.
\newblock In \emph{NeurIPS}, 2021.

\bibitem[Donahue et~al.(2019)Donahue, McAuley, and
  Puckette]{donahue2019adversarial}
Chris Donahue, Julian McAuley, and Miller Puckette.
\newblock Adversarial audio synthesis, 2019.

\bibitem[Esteban et~al.(2017)Esteban, Hyland, and
  Rätsch]{esteban2017realvalued}
Cristóbal Esteban, L.~Stephanie Hyland, and Gunnar Rätsch.
\newblock Real-valued (medical) time series generation with recurrent
  conditional gans, 2017.

\bibitem[Fu(2011)]{fu2011review}
Tak-chung Fu.
\newblock A review on time series data mining.
\newblock \emph{Engineering Applications of Artificial Intelligence},
  24\penalty0 (1):\penalty0 164--181, 2011.

\bibitem[Goodfellow et~al.(2014)Goodfellow, Pouget-Abadie, Mirza, Xu,
  Warde-Farley, Ozair, Courville, and Bengio]{NIPS2014_5423}
Ian Goodfellow, Jean Pouget-Abadie, Mehdi Mirza, Bing Xu, David Warde-Farley,
  Sherjil Ozair, Aaron Courville, and Yoshua Bengio.
\newblock Generative adversarial nets.
\newblock In \emph{NeurIPS}, 2014.

\bibitem[Grathwohl et~al.(2019)Grathwohl, Chen, Bettencourt, Sutskever, and
  Duvenaud]{grathwohl2019ffjord}
Will Grathwohl, Ricky T.~Q. Chen, Jesse Bettencourt, Ilya Sutskever, and David
  Duvenaud.
\newblock Ffjord: Free-form continuous dynamics for scalable reversible
  generative models.
\newblock In \emph{ICLR}, 2019.

\bibitem[Graves(2014)]{graves2014generating}
Alex Graves.
\newblock Generating sequences with recurrent neural networks, 2014.

\bibitem[{Grover} et~al.(2018){Grover}, {Dhar}, and {Ermon}]{grover2018flow}
Aditya {Grover}, Manik {Dhar}, and Stefano {Ermon}.
\newblock Flow-gan: Combining maximum likelihood and adversarial learning in
  generative models.
\newblock In \emph{AAAI}, 2018.

\bibitem[Gulrajani et~al.(2017)Gulrajani, Ahmed, Arjovsky, Dumoulin, and
  Courville]{gulrajani2017improved}
Ishaan Gulrajani, Faruk Ahmed, Martin Arjovsky, Vincent Dumoulin, and Aaron
  Courville.
\newblock Improved training of wasserstein gans.
\newblock \emph{arXiv preprint arXiv:1704.00028}, 2017.

\bibitem[Guo et~al.(2019)Guo, Lin, Feng, Song, and Wan]{guo2019astgcn}
Shengnan Guo, Youfang Lin, Ning Feng, Chao Song, and Huaiyu Wan.
\newblock Attention based spatial-temporal graph convolutional networks for
  traffic flow forecasting.
\newblock \emph{Proceedings of the AAAI Conference on Artificial Intelligence},
  33\penalty0 (01):\penalty0 922--929, Jul. 2019.
\newblock \doi{10.1609/aaai.v33i01.3301922}.

\bibitem[Huang et~al.(2020{\natexlab{a}})Huang, Huang, Liu, Dai, and
  Kong]{huang2020lsgcn}
Rongzhou Huang, Chuyin Huang, Yubao Liu, Genan Dai, and Weiyang Kong.
\newblock Lsgcn: Long short-term traffic prediction with graph convolutional
  networks.
\newblock In \emph{IJCAI}, pages 2355--2361, 2020{\natexlab{a}}.

\bibitem[Huang et~al.(2020{\natexlab{b}})Huang, Sun, and
  Wang]{huang2020learning}
Zijie Huang, Yizhou Sun, and Wei Wang.
\newblock Learning continuous system dynamics from irregularly-sampled partial
  observations.
\newblock In \emph{NeurIPS}, 2020{\natexlab{b}}.

\bibitem[Jhin et~al.(2021)Jhin, Shin, Hong, Jo, Park, Park, Lee, Maeng, and
  Jeon]{jhin2021attentive}
Sheo~Yon Jhin, Heejoo Shin, Seoyoung Hong, Minju Jo, Solhee Park, Noseong Park,
  Seungbeom Lee, Hwiyoung Maeng, and Seungmin Jeon.
\newblock Attentive neural controlled differential equations for time-series
  classification and forecasting.
\newblock In \emph{ICDM}, 2021.

\bibitem[Jia and Benson(2019)]{NIPS2019_9177}
Junteng Jia and Austin~R Benson.
\newblock Neural jump stochastic differential equations.
\newblock In \emph{NeurIPS}, 2019.

\bibitem[Kidger et~al.(2019)Kidger, Bonnier, Arribas, Salvi, and
  Lyons]{DBLP:conf/nips/KidgerBASL19}
Patrick Kidger, Patric Bonnier, Imanol~Perez Arribas, Cristopher Salvi, and
  Terry~J. Lyons.
\newblock Deep signature transforms.
\newblock In \emph{NeurIPS}, 2019.

\bibitem[Kidger et~al.(2020)Kidger, Morrill, Foster, and
  Lyons]{DBLP:conf/nips/KidgerMFL20}
Patrick Kidger, James Morrill, James Foster, and Terry~J. Lyons.
\newblock Neural controlled differential equations for irregular time series.
\newblock In \emph{NeurIPS}, 2020.

\bibitem[Lamb et~al.(2016)Lamb, Goyal, Zhang, Zhang, Courville, and
  Bengio]{lamb2016professor}
Alex Lamb, Anirudh Goyal, Ying Zhang, Saizheng Zhang, Aaron Courville, and
  Yoshua Bengio.
\newblock Professor forcing: A new algorithm for training recurrent networks,
  2016.

\bibitem[Li et~al.(2018)Li, Yu, Shahabi, and Liu]{li2018dcrnn_traffic}
Yaguang Li, Rose Yu, Cyrus Shahabi, and Yan Liu.
\newblock Diffusion convolutional recurrent neural network: Data-driven traffic
  forecasting.
\newblock In \emph{International Conference on Learning Representations (ICLR
  '18)}, 2018.

\bibitem[Lyons et~al.(2007)Lyons, Caruana, and L{\'e}vy]{lyons2007differential}
Terry~J Lyons, Michael Caruana, and Thierry L{\'e}vy.
\newblock \emph{Differential equations driven by rough paths}.
\newblock Springer, 2007.

\bibitem[Mogren(2016)]{mogren2016crnngan}
Olof Mogren.
\newblock C-rnn-gan: Continuous recurrent neural networks with adversarial
  training, 2016.

\bibitem[Radford et~al.(2016)Radford, Metz, and
  Chintala]{radford2016unsupervised}
Alec Radford, Luke Metz, and Soumith Chintala.
\newblock Unsupervised representation learning with deep convolutional
  generative adversarial networks, 2016.

\bibitem[Reinsel(2003)]{reinsel2003elements}
Gregory~C Reinsel.
\newblock \emph{Elements of multivariate time series analysis}.
\newblock Springer Science \& Business Media, 2003.

\bibitem[Ren et~al.(2021)Ren, Li, Ren, Song, Xu, Deng, and Wang]{ren2021deep}
Xiaoli Ren, Xiaoyong Li, Kaijun Ren, Junqiang Song, Zichen Xu, Kefeng Deng, and
  Xiang Wang.
\newblock Deep learning-based weather prediction: A survey.
\newblock \emph{Big Data Research}, 23, 2021.

\bibitem[Song et~al.(2020)Song, Lin, Guo, and Wan]{song2020stsgcn}
Chao Song, Youfang Lin, Shengnan Guo, and Huaiyu Wan.
\newblock Spatial-temporal synchronous graph convolutional networks: A new
  framework for spatial-temporal network data forecasting.
\newblock \emph{Proceedings of the AAAI Conference on Artificial Intelligence},
  34\penalty0 (01):\penalty0 914--921, Apr. 2020.
\newblock \doi{10.1609/aaai.v34i01.5438}.

\bibitem[Tang et~al.(2020)Tang, Yao, Sun, Aggarwal, Mitra, and
  Wang]{tang2020joint}
Xianfeng Tang, Huaxiu Yao, Yiwei Sun, Charu Aggarwal, Prasenjit Mitra, and
  Suhang Wang.
\newblock Joint modeling of local and global temporal dynamics for multivariate
  time series forecasting with missing values.
\newblock In \emph{AAAI}, 2020.

\bibitem[Tekin et~al.(2021)Tekin, Karaahmetoglu, Ilhan, Balaban, and
  Kozat]{tekin2021spatio}
Selim~Furkan Tekin, Oguzhan Karaahmetoglu, Fatih Ilhan, Ismail Balaban, and
  Suleyman~Serdar Kozat.
\newblock Spatio-temporal weather forecasting and attention mechanism on
  convolutional lstms.
\newblock \emph{arXiv preprint}, 2021.

\bibitem[van~den Oord et~al.(2016)van~den Oord, Dieleman, Zen, Simonyan,
  Vinyals, Graves, Kalchbrenner, Senior, and Kavukcuoglu]{oord2016wavenet}
Aaron van~den Oord, Sander Dieleman, Heiga Zen, Karen Simonyan, Oriol Vinyals,
  Alex Graves, Nal Kalchbrenner, Andrew Senior, and Koray Kavukcuoglu.
\newblock Wavenet: A generative model for raw audio, 2016.

\bibitem[Van~der Maaten and Hinton(2008)]{van2008visualizing}
Laurens Van~der Maaten and Geoffrey Hinton.
\newblock Visualizing data using t-sne.
\newblock \emph{Journal of machine learning research}, 9\penalty0 (11), 2008.

\bibitem[Wu et~al.(2019)Wu, Pan, Long, Jiang, and Zhang]{wu2019graphwavenet}
Zonghan Wu, Shirui Pan, Guodong Long, Jing Jiang, and Chengqi Zhang.
\newblock Graph wavenet for deep spatial-temporal graph modeling.
\newblock In \emph{Proceedings of the Twenty-Eighth International Joint
  Conference on Artificial Intelligence, {IJCAI-19}}, pages 1907--1913, 7 2019.

\bibitem[Xu and Xie(2020)]{xu2020conformal}
Chen Xu and Yao Xie.
\newblock Conformal prediction for dynamic time-series.
\newblock In \emph{ICML}, 2020.

\bibitem[Yoon et~al.(2019)Yoon, Jarrett, and van~der
  Schaar]{NEURIPS2019_c9efe5f2}
Jinsung Yoon, Daniel Jarrett, and Mihaela van~der Schaar.
\newblock Time-series generative adversarial networks.
\newblock In \emph{NeurIPS}, 2019.

\bibitem[Yu et~al.(2018)Yu, Yin, and Zhu]{bing2018stgcn}
Bing Yu, Haoteng Yin, and Zhanxing Zhu.
\newblock Spatio-temporal graph convolutional networks: A deep learning
  framework for traffic forecasting.
\newblock In \emph{Proceedings of the Twenty-Seventh International Joint
  Conference on Artificial Intelligence, {IJCAI-18}}, pages 3634--3640, 7 2018.
\newblock \doi{10.24963/ijcai.2018/505}.

\bibitem[Zhang et~al.(2021)Zhang, Zeman, Tsiligkaridis, and
  Zitnik]{zhang2021graph}
Xiang Zhang, Marko Zeman, Theodoros Tsiligkaridis, and Marinka Zitnik.
\newblock Graph-guided network for irregularly sampled multivariate time
  series.
\newblock In \emph{ICLR}, 2021.

\end{thebibliography}

\section*{Checklist}


\begin{enumerate}

\item For all authors...
\begin{enumerate}
  \item Do the main claims made in the abstract and introduction accurately reflect the paper's contributions and scope?
    \answerYes{We did.}
  \item Did you describe the limitations of your work?
    \answerYes{See Section~\ref{discussions}}
  \item Did you discuss any potential negative societal impacts of your work?
    \answerYes{See Section~\ref{discussions}}
  \item Have you read the ethics review guidelines and ensured that your paper conforms to them?
    \answerYes{}
\end{enumerate}

\item If you are including theoretical results...
\begin{enumerate}
  \item Did you state the full set of assumptions of all theoretical results?
    \answerNA{}
        \item Did you include complete proofs of all theoretical results?
    \answerNA{}
\end{enumerate}

\item If you ran experiments...
\begin{enumerate}
  \item Did you include the code, data, and instructions needed to reproduce the main experimental results (either in the supplemental material or as a URL)?
     \answerYes{See our supplemental material. }
  \item Did you specify all the training details (e.g., data splits, hyperparameters, how they were chosen)?
    \answerYes{In Appendix~\ref{reproduce}, we specify all the training details.}
        \item Did you report error bars (e.g., with respect to the random seed after running experiments multiple times)?
    \answerYes{In Section~\ref{Experimental}, We report the results of 10 experiments by calculating the mean and variance.}
        \item Did you include the total amount of compute and the type of resources used (e.g., type of GPUs, internal cluster, or cloud provider)?
    \answerYes{See Section~\ref{Experimental}}
\end{enumerate}

\item If you are using existing assets (e.g., code, data, models) or curating/releasing new assets...
\begin{enumerate}
  \item If your work uses existing assets, did you cite the creators?
    \answerYes{See Appendix~\ref{baselines} }
  \item Did you mention the license of the assets?
    \answerYes{}
  \item Did you include any new assets either in the supplemental material or as a URL?
    \answerYes{We release our model. See our supplementary material.}
  \item Did you discuss whether and how consent was obtained from people whose data you're using/curating?
    \answerNA{}
  \item Did you discuss whether the data you are using/curating contains personally identifiable information or offensive content?
    \answerNA{}
\end{enumerate}

\item If you used crowdsourcing or conducted research with human subjects...
\begin{enumerate}
  \item Did you include the full text of instructions given to participants and screenshots, if applicable?
    \answerNA{}
  \item Did you describe any potential participant risks, with links to Institutional Review Board (IRB) approvals, if applicable?
    \answerNA{}
  \item Did you include the estimated hourly wage paid to participants and the total amount spent on participant compensation?
    \answerNA{}
\end{enumerate}

\end{enumerate}


\clearpage
\appendix
\label{appendix}

\section{Datasets}
We use 2 simulated (Sines, MuJoCo) and 2 real-world (Stocks, Energy) datasets. Table.~\ref{tab:data_statisics} shows the statistics of the datasets. All datasets are available online via the link. We note that in some of our datasets, the time series length $N$ can be varied from one time series sample to another. However, our framework has no problems in dealing with those varying lengths.

\begin{table}[ht]
\centering
\caption{\label{tab:data_statisics}Dataset Statistics}
\begin{tabular}{clccccc}
\hline
\multicolumn{2}{c}{Dataset} & \# of Samples & $\dim(\mathbf{x})$ & Average of $N$ & Link & License \\ \hline
\multicolumn{2}{c}{Sines}   & 10,000     & 5    & 24 time-points    &    -         & -         \\
\multicolumn{2}{c}{Stocks}  & 3,773     & 6    & 24 days    &     \href{https://finance.yahoo.com/quote/GOOG/history?p=GOOG}{Link}      & -          \\
\multicolumn{2}{c}{Energy}  & 19,711    & 28   & 24 hours  & \href{https://archive.ics.uci.edu/ml/datasets/Appliances+energy+prediction }{Link}  & CC BY 4.0          \\
\multicolumn{2}{c}{MuJoCo}  & 4,620      & 14   & 24 time-points     &   \href{https://github.com/deepmind/dm_control}{Link}   & Apache License 2.0
          \\ \hline
\end{tabular}
\end{table}

\section{ODE/CDE functions in GT-GAN}\label{a:model}

\subsection{Encoder}
Our encoder based on NCDEs has the following CDE function $f$.\footnote{CDE: \url{https://github.com/patrick-kidger/NeuralCDE} (Apache-2.0 license)}
\begin{table}[ht]
\centering
\setlength{\tabcolsep}{2pt}
\caption{The architecture of the network $f$ in the encoder}\label{tbl:encoder_f}
\begin{tabular}{cccc}\hline
Layer & Design & Input Size & Output Size \\ \hline
1 & ReLU(Linear) & $N \times \dim(\mathbf{x})$  & $N \times 4\dim(\mathbf{x})$ \\ 
2 & ReLU(Linear) & $N \times 4\dim(\mathbf{x})$ & $N \times 4\dim(\mathbf{x})$ \\ 
3 & ReLU(Linear) & $N \times 4\dim(\mathbf{x})$ & $N \times 4\dim(\mathbf{x})$ \\ 
4 & Tanh(Linear) & $N \times 4\dim(\mathbf{x})$ & $N \times \dim(\mathbf{x})$ \\ \hline
\end{tabular}
\end{table}

\subsection{Decoder, Discriminator}
Our decoder and discriminator based on GRU-ODEs have the following ODE functions.\footnote{ODE: \url{https://github.com/rtqichen/torchdiffeq} (MIT license)} They have the same architecture but their parameters are separated.

\begin{table}[ht]
\centering
\setlength{\tabcolsep}{2pt}
\caption{The architecture of the network $g$ in the decoder}\label{tbl:decoder_g}
\begin{tabular}{cccc}\hline
Layer & Design & Input Size & Output Size \\ \hline
\multirow{4}{*}{1} & $r_t = $Sigmoid(Linear)  & $N \times \dim(\mathbf{h})$  & $N \times \dim(\mathbf{h})$ \\
 &  $z_t = $Sigmoid(Linear) & $N \times \dim(\mathbf{h})$  & $N \times \dim(\mathbf{h})$ \\
 & $u_t = $Tanh(Linear) & $N \times \dim(\mathbf{h})$  & $N \times \dim(\mathbf{h})$ \\
 & $dh = (1-z_t) * (u_t-h_t)$ & $N \times \dim(\mathbf{h})$  & $N \times \dim(\mathbf{h})$ \\
\hline
\end{tabular}
\end{table}

\begin{table}[hbt!]
\centering
\setlength{\tabcolsep}{2pt}
\caption{The architecture of the network $q$ in the discriminator}\label{tbl:decoder_q}
\begin{tabular}{cccc}\hline
Layer & Design & Input Size & Output Size \\ \hline
\multirow{4}{*}{1} & $r_t = $Sigmoid(Linear)  & $N \times \dim(\mathbf{x}) $  & $N \times \dim(\mathbf{x}) $ \\
 &  $z_t = $Sigmoid(Linear) & $N \times \dim(\mathbf{x}) $  & $N \times \dim(\mathbf{x}) $ \\
 & $u_t = $Tanh(Linear) & $N \times \dim(\mathbf{x}) $  & $N \times \dim(\mathbf{x}) $ \\
 & $dh = (1-z_t) * (u_t-h_t)$ & $N \times \dim(\mathbf{x}) $  & $N \times \dim(\mathbf{x}) $ \\
\hline
\end{tabular}
\end{table}

\subsection{Generator}\label{a:generator}
Our generator has the following ODE function $f$ in Table~\ref{tbl:generator_f}. 

\begin{table}[ht]
\vspace{-5em}
\begin{minipage}{0.7\linewidth}
\caption{The architecture of the network $r$ in the generator}\label{tbl:generator_f}
\centering
\begin{tabular}{cccc}\hline
Layer & Design & Input Size & Output Size \\ \hline
1 & Softplus(Linear) & $N \times \dim(\mathbf{h}+1)$  & $N \times  \dim(\mathbf{h})$ \\ 
2 & Softplus(Linear) & $N \times  \dim(\mathbf{h}+1)  $ & $N \times  \dim(\mathbf{h}) $\\ 
3 & Softplus(Linear) & $N \times  \dim(\mathbf{h}+1) $ & $N \times  \dim(\mathbf{h})  $ \\  \hline
\end{tabular}
\end{minipage}\hfill
\begin{minipage}{0.3\linewidth}
\vspace{2em}
\begin{figure}[H]
\centering
\includegraphics[width=0.9\linewidth]{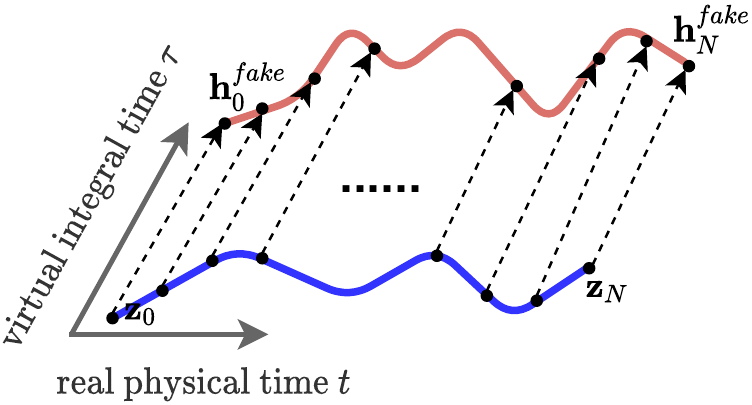}
\caption{An example of our generation}
\label{fig:ctfp}
\end{figure}
\vspace{0.5em}
\end{minipage}
\vspace{-4em}
\end{table}

\section{Baselines}
\label{baselines}
For the regular time series baseline models, i.e., TimeGAN, RCGAN, C-RNN-GAN, T-forcing, and P-forcing, we use the 3-layer GRU-based neural network architecture with a hidden size that is 4 times larger than the input size. We use or modify the following accessible source codes to run.
\begin{itemize}
\item TimeGAN : https://github.com/jsyoon0823/TimeGAN
\item RCGAN : https://github.com/3778/Ward2ICU
\item C-RNN-GAN : https://github.com/olofmogren/c-rnn-gan
\item T-forcing, P-forcing : https://github.com/mojesty/professor\_forcing
\item GRU-D : https://github.com/zhiyongc/GRU-D
\end{itemize}
Because ordinary GRUs can not be applied to irregular time series, we replace the first layer GRU to GRU-$\bigtriangleup t$ and GRU-D in all those baselines so that the redesigned baseline models, i.e., TimeGAN-$\bigtriangleup t$, RCGAN-$\bigtriangleup t$, C-RNN-GAN-$\bigtriangleup t$, T-forcing-$\bigtriangleup t$, P-forcing-$\bigtriangleup t$, TimeGAN-D, RCGAN-D, C-RNN-GAN-D, T-forcing-D and P-forcing-D, can process irregular time series data.

\section{Evaluation metrics}

\begin{figure}[hbt!]
    \hspace*{\fill}%
    \subfigure[How to calculate the predictive score for the regular time series synthesis in TimeGAN]{\includegraphics[width=0.35\columnwidth]{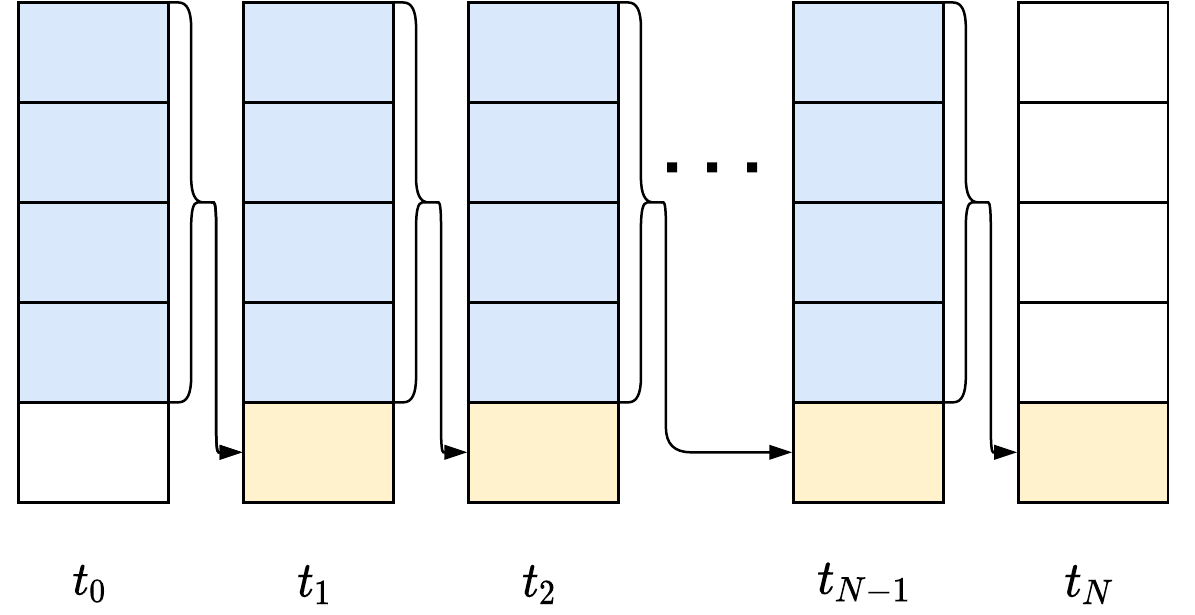}}\hfill
    \subfigure[How to calculate the predictive score for the irregular time series in this paper]{\includegraphics[width=0.35\columnwidth]{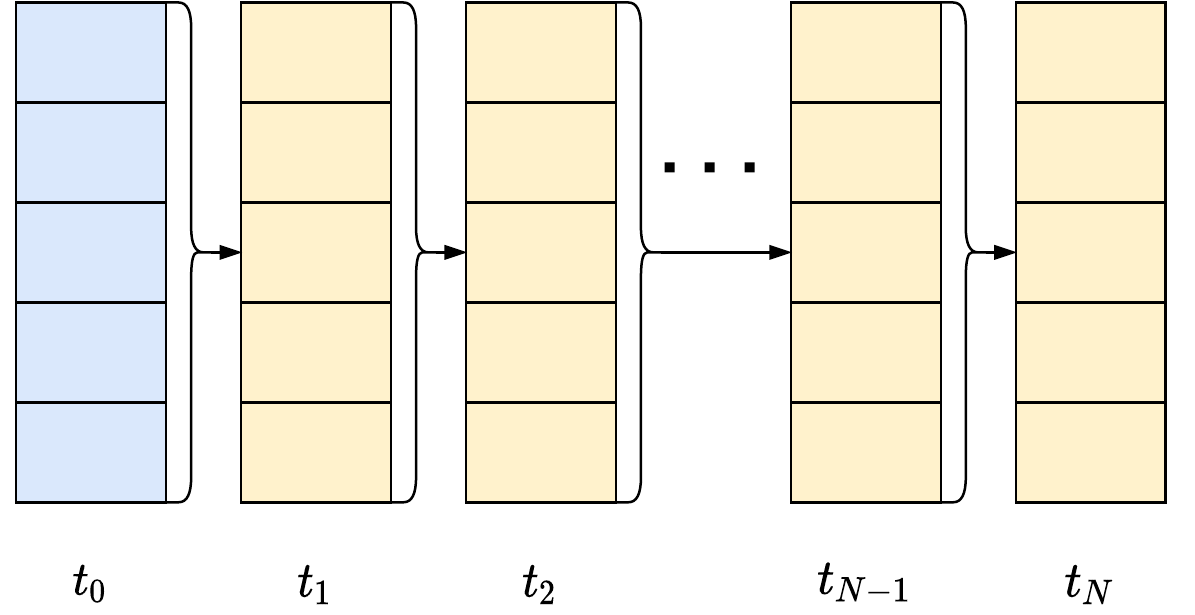}}
    \hspace*{\fill}%
    \caption{Predictive task according to the data type}
    \label{fig:predictive}
\end{figure}

For fair comparison, we reuse the experimental environments of TimeGAN for the discriminative score. However, we found that TimeGAN's predictive task is rather straightforward as shown in Fig.~\ref{fig:predictive} (a). It predicts only one element in yellow from other four past elements in blue. Since only one element is used for evaluation, we found that the original predictive score of TimeGAN can be biased. Instead, our predictive task predicts the entire vector, as shown in Fig.~\ref{fig:predictive} (b), and therefore, our predictive score is measured under much more challenging environments. We use this more challenging predictive score definition for our irregular time series synthesis. We stick to the TimeGAN's definition for the regular time series experiment for fair comparison but use our challenging predictive score metric for all other experiments.

\section{Additional ablation studies}\label{a:abl}

In Tables~\ref{tab:ablation_30} to~\ref{tab:ablation_model3}, we report the missing ablation study tables in the main paper.

\begin{table}[ht]
\vspace{-1em}
\centering
\caption{\label{tab:ablation_30}Ablation study for training options with the irregular time series (30\% dropped)}
\begin{tabular}{cccccc}
\hline
Metric & Method    & Sines & Stocks & Energy & MuJoCo \\ \hline
\multirow{3}{*}{\begin{tabular}[c]{@{}c@{}}Discriminative \\ Score\\ (Lower the Better)\end{tabular}} 
& GT-GAN      &   \textbf{.363} & \textbf{.251}   &  \textbf{.333}  &  .249  \\  
& w/o Eq.~\eqref{eq:for}   &  .498 & .266 &  .392  & .303    \\ 
& w/o pre-training & .499 & .305  &.345& \textbf{.241} \\ \hline
\multirow{3}{*}
{\begin{tabular}[c]{@{}c@{}}Predictive \\ Score\\ (Lower the Better)\end{tabular}}
& GT-GAN  & \textbf{.099}     &  .021     &.066   &  \textbf{.048}    \\ 
& w/o Eq.~\eqref{eq:for}   & .241  &  \textbf{.015} & .064 & .061\\ 
& w/o pre-training &.273 & .022 & \textbf{.061} &  .049 \\ \hline
\end{tabular}
\end{table}

\begin{table}[ht]
\vspace{-1em}
\centering
\caption{\label{tab:ablation_50}Ablation study for training options with the irregular time series (50\% dropped)}
\begin{tabular}{cccccc}
\hline
Metric & Method    & Sines & Stocks & Energy & MuJoCo \\ \hline
\multirow{3}{*}{\begin{tabular}[c]{@{}c@{}}Discriminative \\ Score\\ (Lower the Better)\end{tabular}} 
& GT-GAN      &   \textbf{.372} & .265   &  \textbf{.317}  &  \textbf{.270}  \\  
& w/o Eq.~\eqref{eq:for}   & .500 & .323 &  .381 & .274    \\ 
& w/o pre-training & .500& \textbf{.209}  & .325 &\textbf{.270} \\ \hline
\multirow{3}{*}
{\begin{tabular}[c]{@{}c@{}}Predictive \\ Score\\ (Lower the Better)\end{tabular}}                  
& GT-GAN  & \textbf{.101}   &  .018   &.064   &  .056   \\ 
& w/o Eq.~\eqref{eq:for}   &.277 &  .018 & \textbf{.063} & \textbf{.051}\\ 
& w/o pre-training & .103& \textbf{.017}& .071 & \textbf{.051}   \\ \hline
\end{tabular}
\end{table}

\begin{table}[ht]
\vspace{-1em}
\centering
\caption{\label{tab:ablation_70}Ablation study for training options with the irregular time series (70\% dropped)}
\begin{tabular}{cccccc}
\hline
Metric & Method    & Sines & Stocks & Energy & MuJoCo \\ \hline
\multirow{3}{*}{\begin{tabular}[c]{@{}c@{}}Discriminative \\ Score\\ (Lower the Better)\end{tabular}} 
& GT-GAN      &   \textbf{.278} & \textbf{.230}   &  \textbf{.325}  &  .275  \\  
& w/o Eq.~\eqref{eq:for}   & .319  & .274 & .382   & .290    \\ 
& w/o pre-training & .408&  .311 & .345 &\textbf{.249}  \\ \hline
\multirow{3}{*}
{\begin{tabular}[c]{@{}c@{}}Predictive \\ Score\\ (Lower the Better)\end{tabular}}
& GT-GAN  & .088    &  \textbf{.020}     &.076   &  .052    \\ 
& w/o Eq.~\eqref{eq:for}   & \textbf{.082} & .025 & \textbf{.066}& .051 \\ 
& w/o pre-training &.104& \textbf{.020} &.085 & \textbf{.049}   \\ \hline
\end{tabular}
\end{table}

\begin{table}
\vspace{-1em}
\small
\setlength{\tabcolsep}{3pt}
\centering
\caption{\label{tab:ablation_model1}Ablation study for model architecture in Sines}
\begin{tabular}{ccccccccc}
\hline
Sines & \multicolumn{2}{c}{GT-GAN (w.o. AE)} & \multicolumn{2}{c}{GT-GAN (Flow only)} & \multicolumn{2}{c}{GT-GAN (AE only)} & \multicolumn{2}{c}{GT-GAN (Full model)} \\ \hline
Metric & \multicolumn{1}{c}{\;\;\;Disc.\;\;\;} & Pred. & \multicolumn{1}{c}{\;\;\;Disc.\;\;\;} & Pred. & \multicolumn{1}{c}{\;\;\;Disc.\;\;\;} & Pred. & \multicolumn{1}{c}{\;\;\;Disc.\;\;\;} & Pred. \\ \hline
30\% dropped & \multicolumn{1}{c}{.472} & .262 & \multicolumn{1}{c}{.500} & .513 & \multicolumn{1}{c}{.477} & .270& \multicolumn{1}{c}{\textbf{.363}} & \textbf{.099} \\ 
50\% dropped & \multicolumn{1}{c}{.480} & .254 & \multicolumn{1}{c}{.500} & .610& \multicolumn{1}{c}{.475} & .253 & \multicolumn{1}{c}{\textbf{.372}} & \textbf{.101} \\
70\% dropped & \multicolumn{1}{c}{.481} &.248 & \multicolumn{1}{c}{.499} & .614 & \multicolumn{1}{c}{.477} & .266 & \multicolumn{1}{c}{\textbf{.278}} & \textbf{.088} \\ \hline
\end{tabular}
\end{table}

\begin{table}
\vspace{-1em}
\small
\setlength{\tabcolsep}{3pt}
\centering
\caption{\label{tab:ablation_model2}Ablation study for model architecture in Stocks}
\begin{tabular}{ccccccccc}\hline
Stocks & \multicolumn{2}{c}{GT-GAN (w.o. AE)} & \multicolumn{2}{c}{GT-GAN (Flow only)} & \multicolumn{2}{c}{GT-GAN (AE only)} & \multicolumn{2}{c}{GT-GAN (Full model)} \\ \hline
Metric & \multicolumn{1}{c}{Disc.} & \multicolumn{1}{c}{Pred.} & \multicolumn{1}{c}{Disc.} & \multicolumn{1}{c}{Pred.} & \multicolumn{1}{c}{Disc.} & \multicolumn{1}{c}{Pred.} & \multicolumn{1}{c}{\;\;\;Disc.\;\;\;} & \multicolumn{1}{c}{Pred.} \\ \hline
30\% dropped & \multicolumn{1}{c}{.500} & .088 & \multicolumn{1}{c}{.492} & .140 & \multicolumn{1}{c}{.486} & .088 & \multicolumn{1}{c}{\textbf{.251}} & \textbf{.021} \\
50\% dropped & \multicolumn{1}{c}{.500} & .088 & \multicolumn{1}{c}{.491} & .128 & \multicolumn{1}{c}{.492} & .125 & \multicolumn{1}{c}{\textbf{.265}} & \textbf{.018} \\
70\% dropped & \multicolumn{1}{c}{.500} & .088 & \multicolumn{1}{c}{.490} & .128 & \multicolumn{1}{c}{.492} & .122 & \multicolumn{1}{c}{\textbf{.230}} & \textbf{.020} \\ \hline
\end{tabular}
\end{table}

\begin{table}[ht]
\vspace{-1em}
\small
\setlength{\tabcolsep}{3pt}
\centering
\caption{\label{tab:ablation_model3}Ablation study for model architecture in Energy}
\begin{tabular}{ccccccccc}
\hline
Energy  & \multicolumn{2}{c}{GT-GAN (w.o. AE)} & \multicolumn{2}{c}{GT-GAN (Flow only)} & \multicolumn{2}{c}{GT-GAN (AE only)} & \multicolumn{2}{c}{GT-GAN (Full model)} \\ \hline
Metric & \multicolumn{1}{c}{Disc.} & Pred. & \multicolumn{1}{c}{Disc.} & Pred. & \multicolumn{1}{c}{Disc.} & Pred. & \multicolumn{1}{c}{\;\;\;Disc.\;\;\;} & Pred. \\ \hline
30\% dropped & \multicolumn{1}{c}{.500} &.305  & \multicolumn{1}{c}{.498} & .252 & \multicolumn{1}{c}{.495} & .162 & \multicolumn{1}{c}{\textbf{.333}} & \textbf{.066} \\
50\% dropped & \multicolumn{1}{c}{.500} & .365 & \multicolumn{1}{c}{.499} &.160 & \multicolumn{1}{c}{.499} & .135& \multicolumn{1}{c}{\textbf{.317}} & \textbf{.064} \\
70\% dropped & \multicolumn{1}{c}{.499} & .376 & \multicolumn{1}{c}{.499} & .184 & \multicolumn{1}{c}{.499} & .131 & \multicolumn{1}{c}{\textbf{.325}} & \textbf{.076} \\ \hline
\end{tabular}
\end{table}

\clearpage
\section{Sensitivity analyses}\label{a:sen}

We provide performance (discriminative score and predictive score) depending on hyperparameters (i.e. atol (absolute tolerance)$=\{1e-1, 1e-2, 1e-3\}$, rtol (relative tolerance)$=\{1e-1, 1e-2, 1e-3\}$ and $P_{MLE}=\{1, 2, 3\}$) for each different datasets.

\begin{figure}[hbt!]
    \centering
    {\includegraphics[width=0.24\columnwidth]{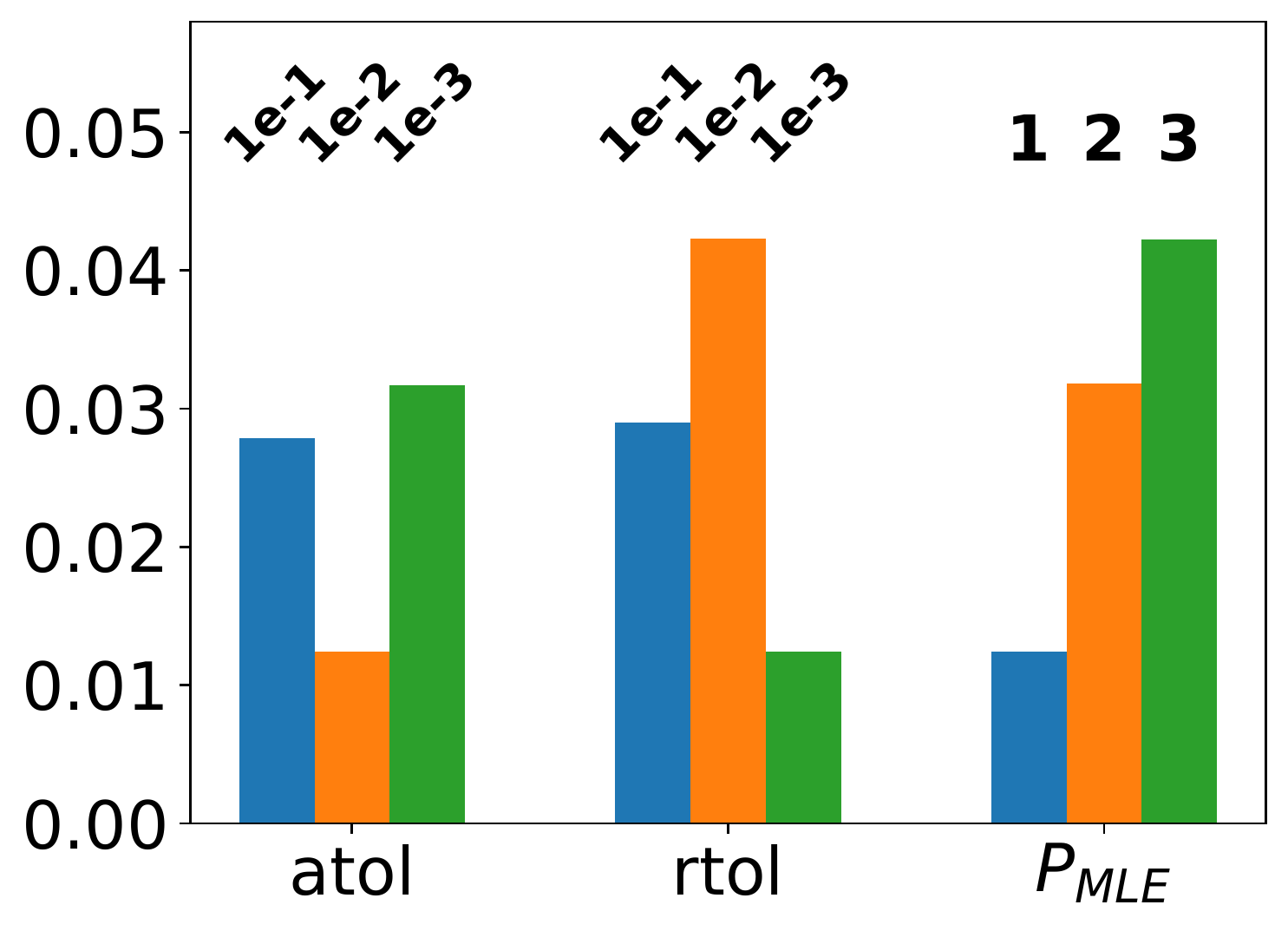}}\hfill
    {\includegraphics[width=0.24\columnwidth]{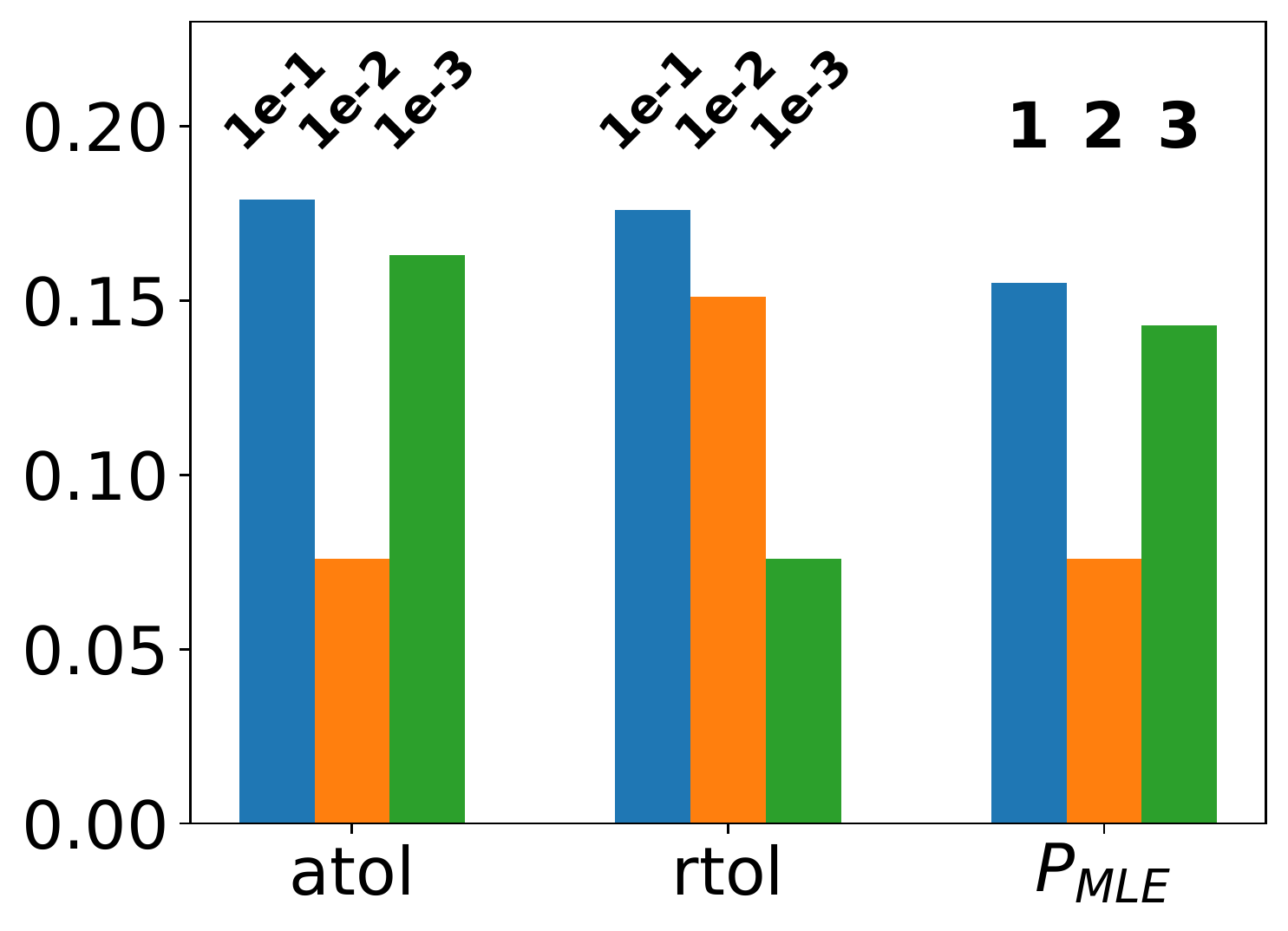}}\hfill
    {\includegraphics[width=0.24\columnwidth]{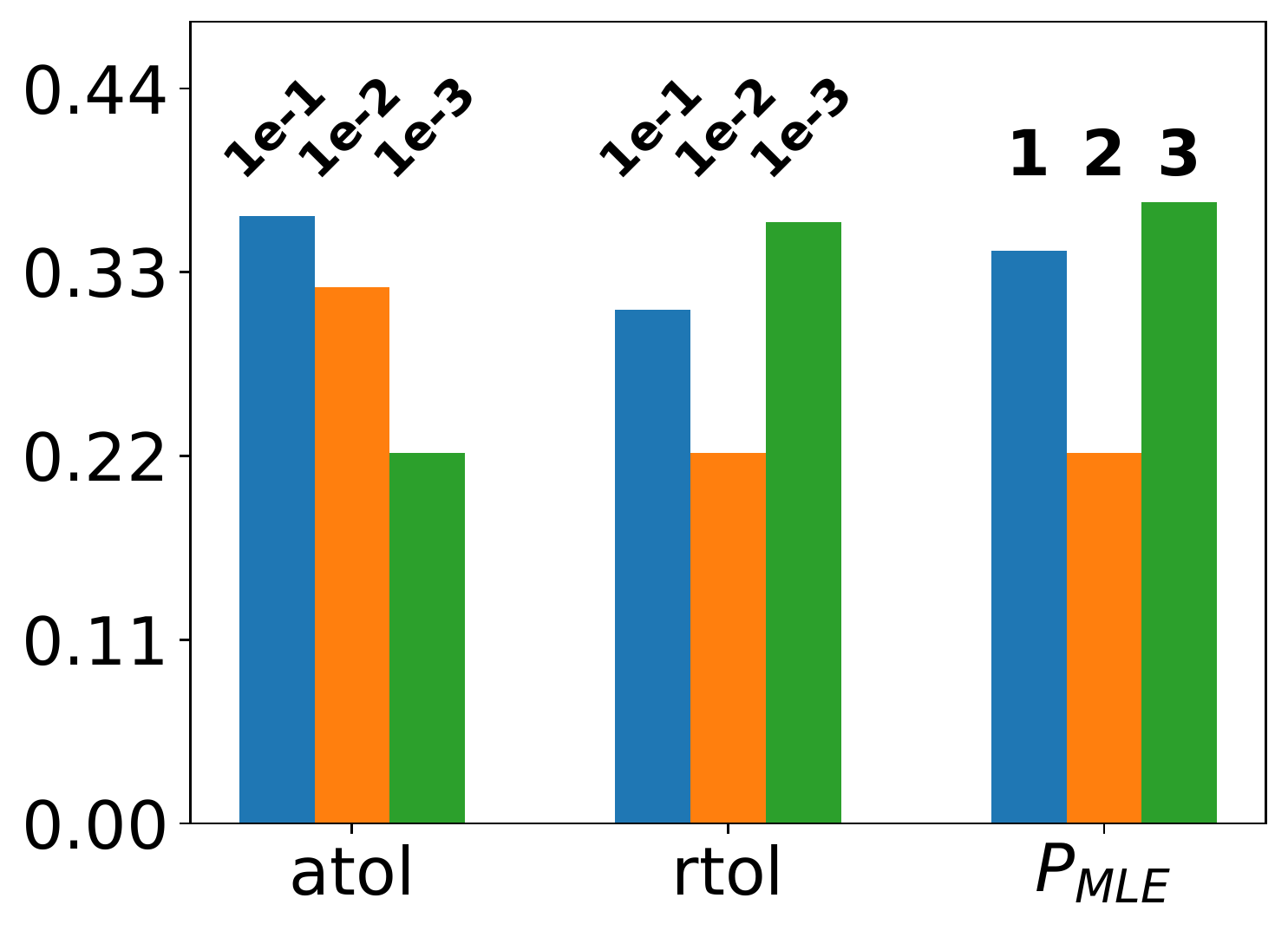}}\hfill
    {\includegraphics[width=0.24\columnwidth]{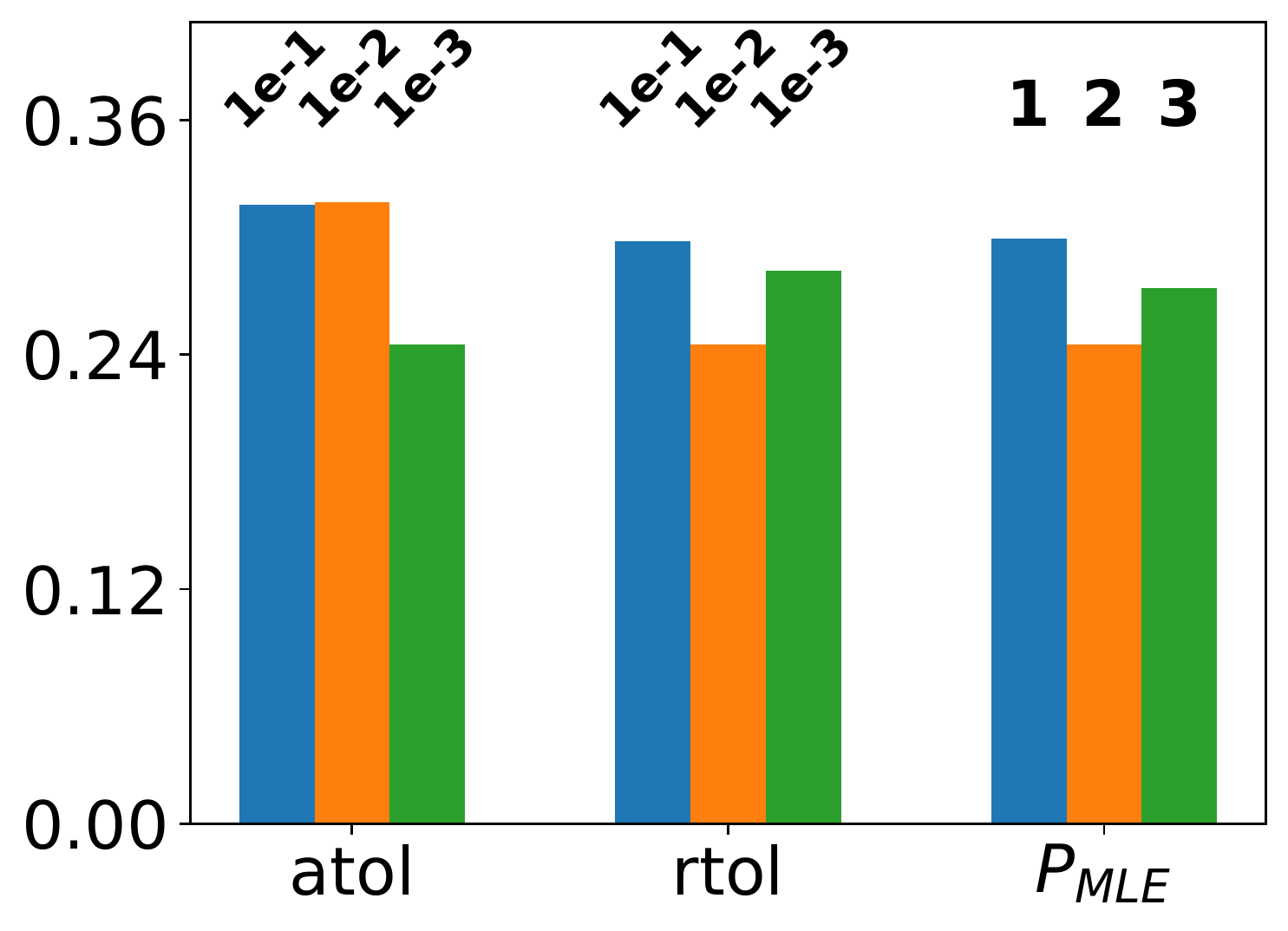}}\hfill
    \newline
    \centering
    \subfigure[Sines]{\includegraphics[width=0.24\columnwidth]{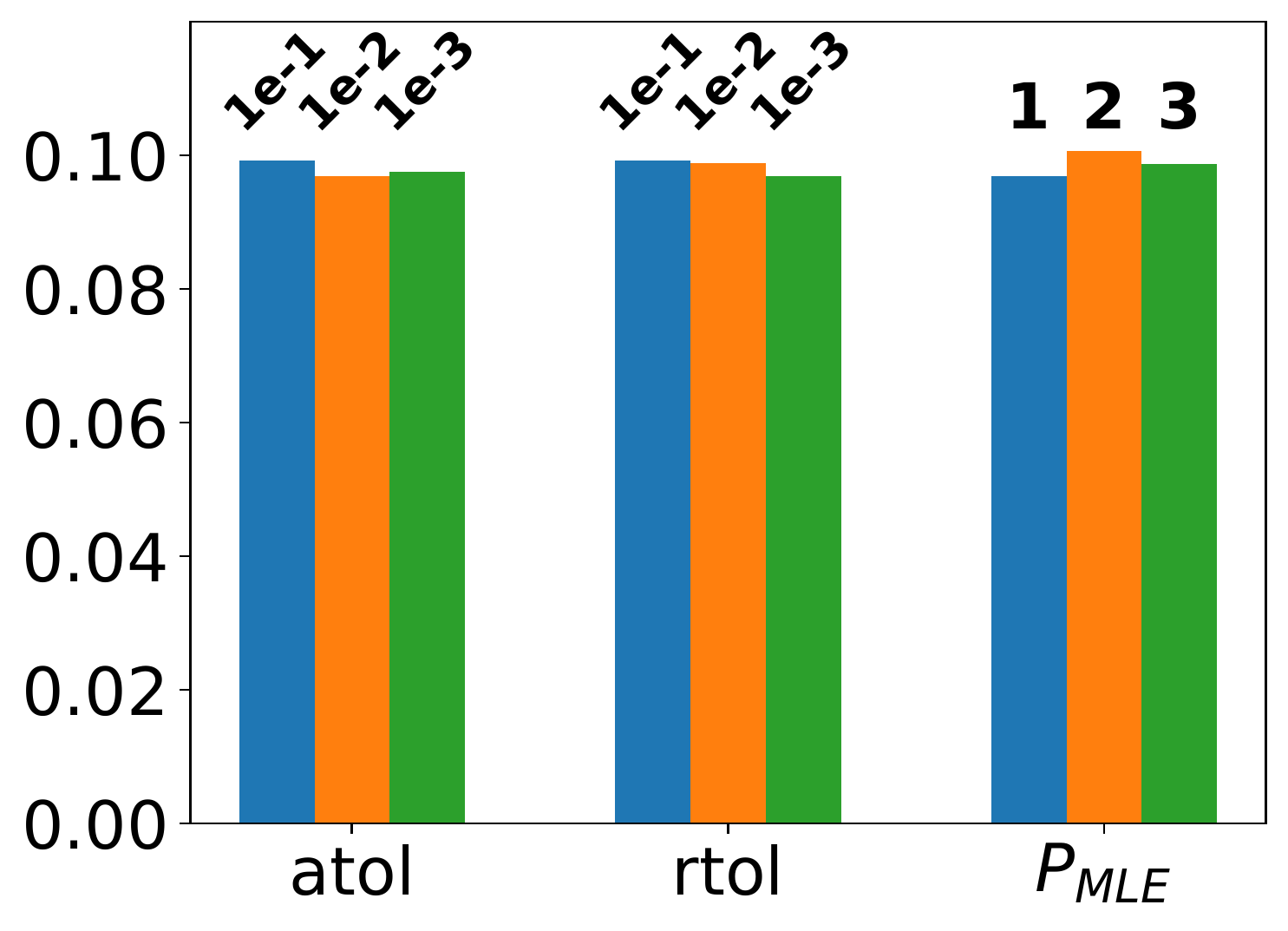}}\hfill
    \subfigure[Stocks]{\includegraphics[width=0.24\columnwidth]{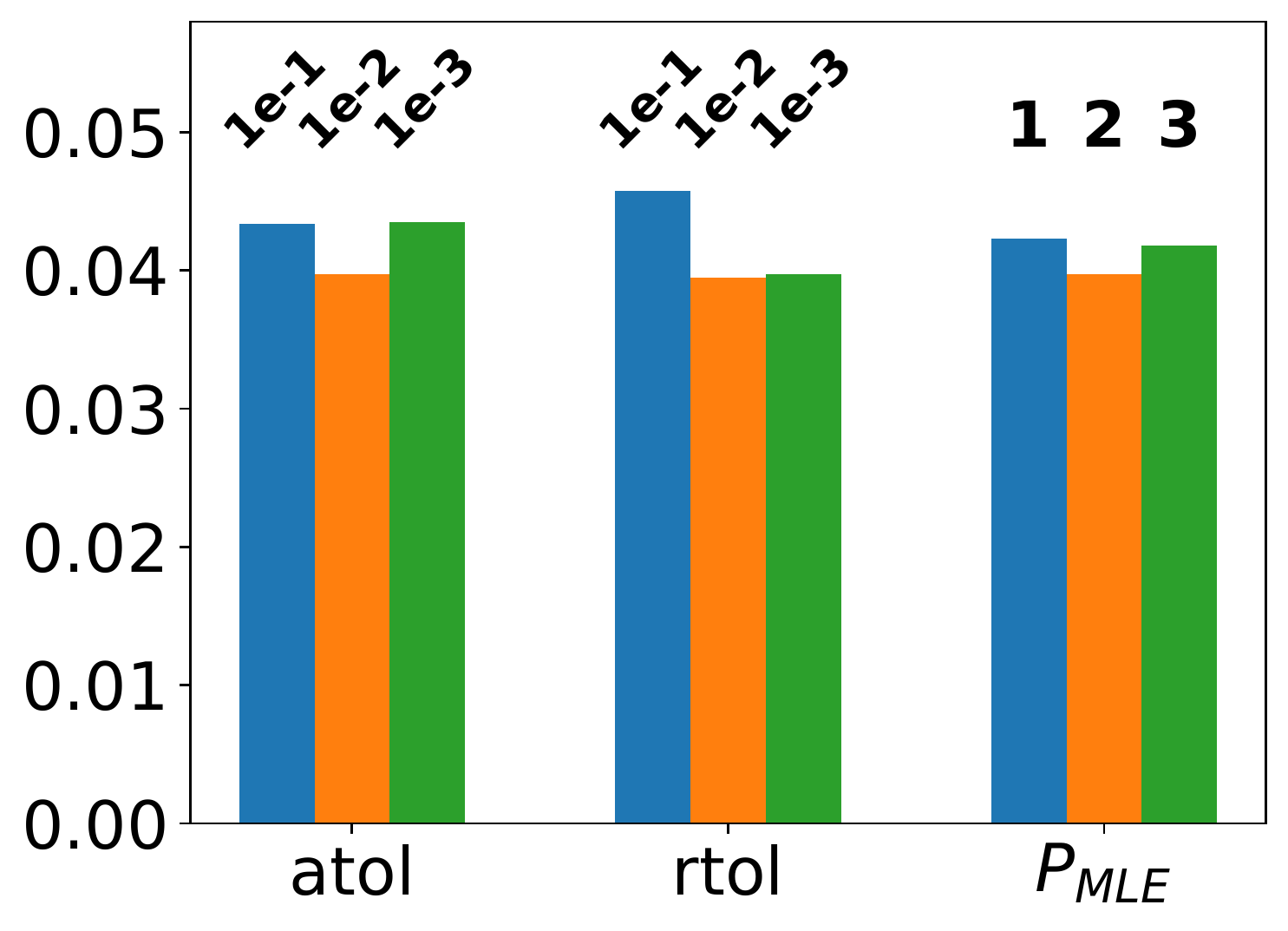}}\hfill
    \subfigure[Energy]{\includegraphics[width=0.24\columnwidth]{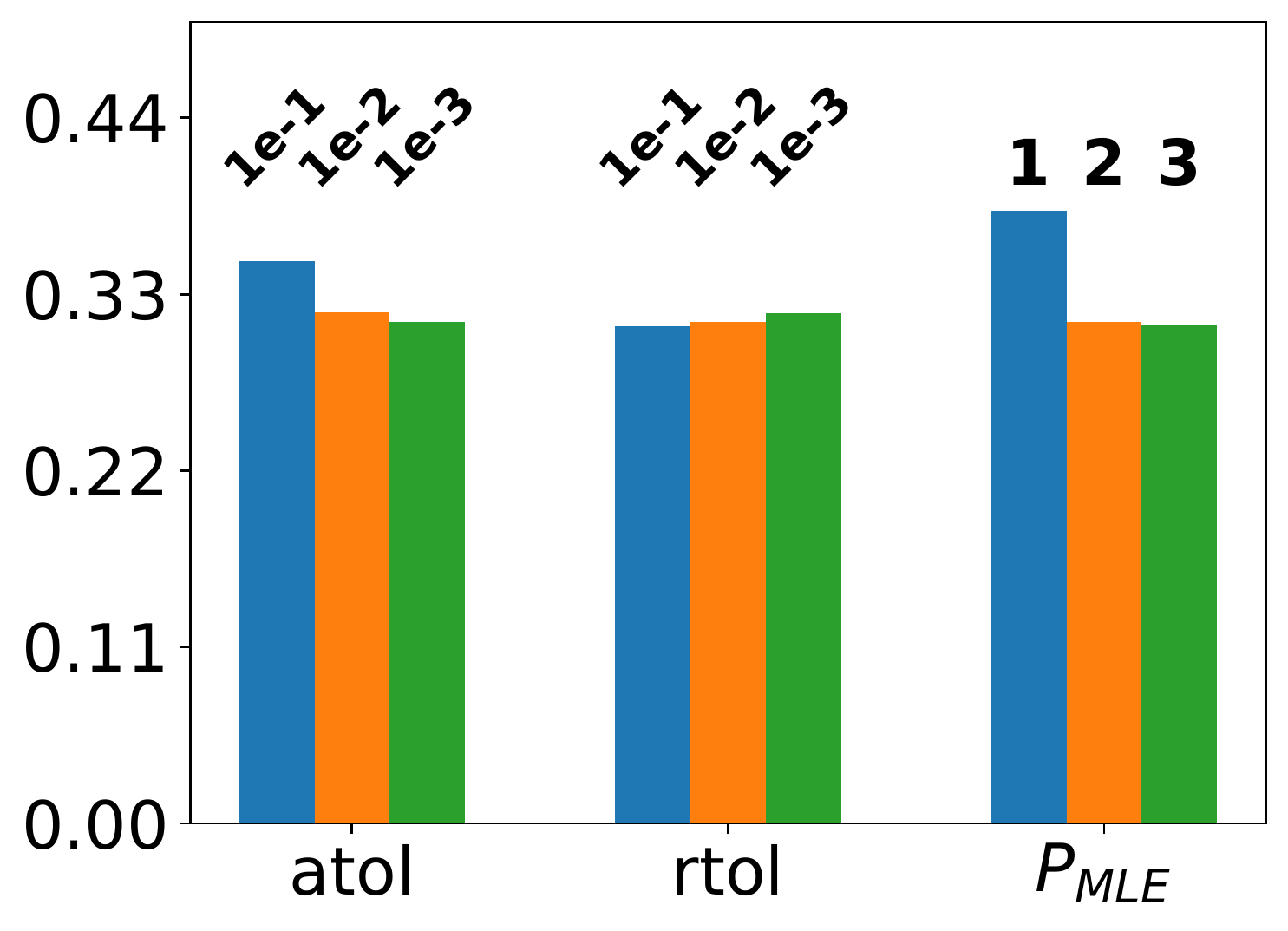}}\hfill
    \subfigure[MuJoCo]{\includegraphics[width=0.24\columnwidth]{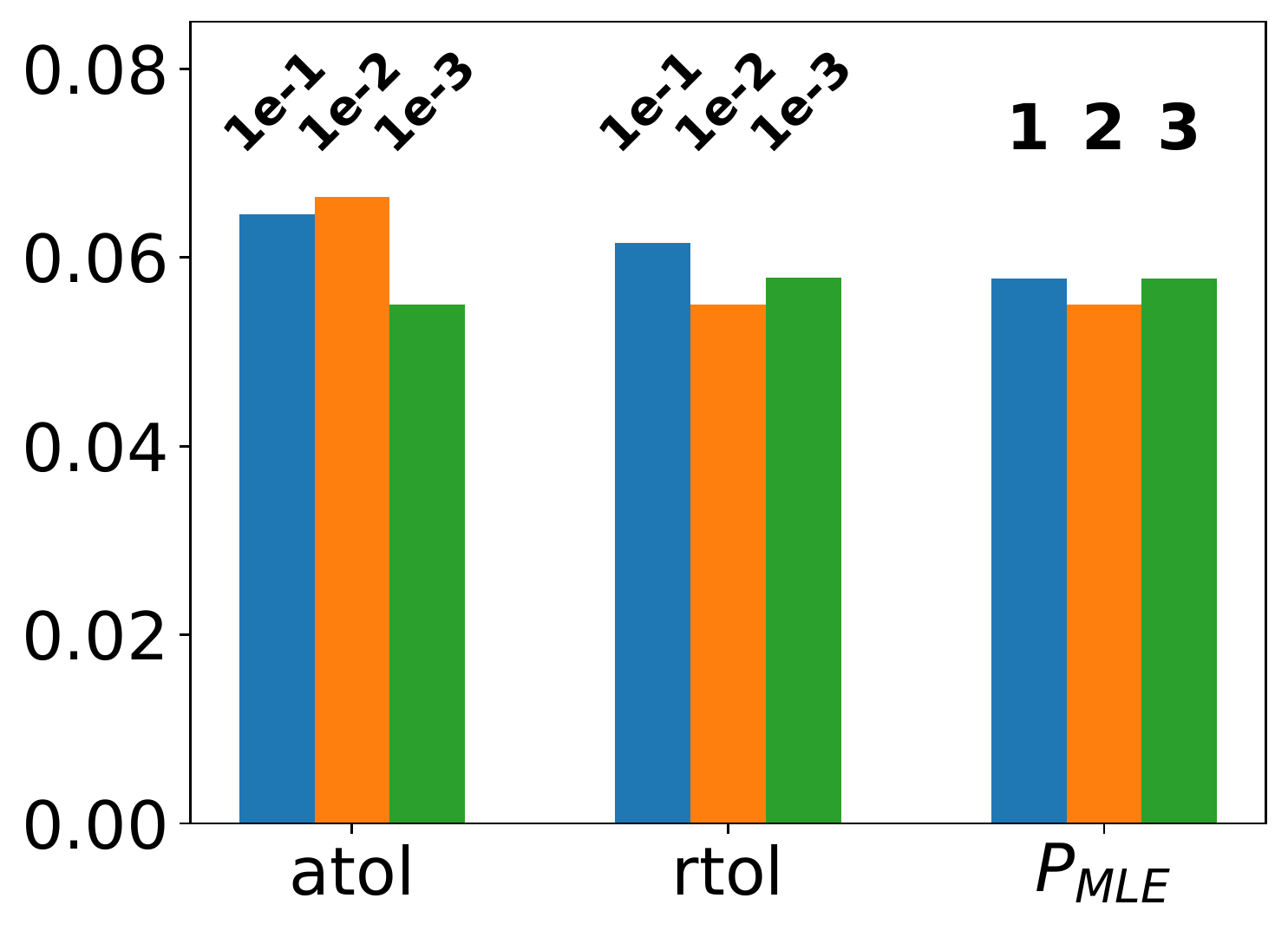}}\hfill
    \caption{The sensitivity of the discriminative score (the 1$^{\text{st}}$ row) and predictive score (the 2$^{\text{nd}}$ row) w.r.t. some key hyperparameters for regular data}
\end{figure}

\begin{figure}[hbt!]
\centering
{\includegraphics[width=0.24\columnwidth]{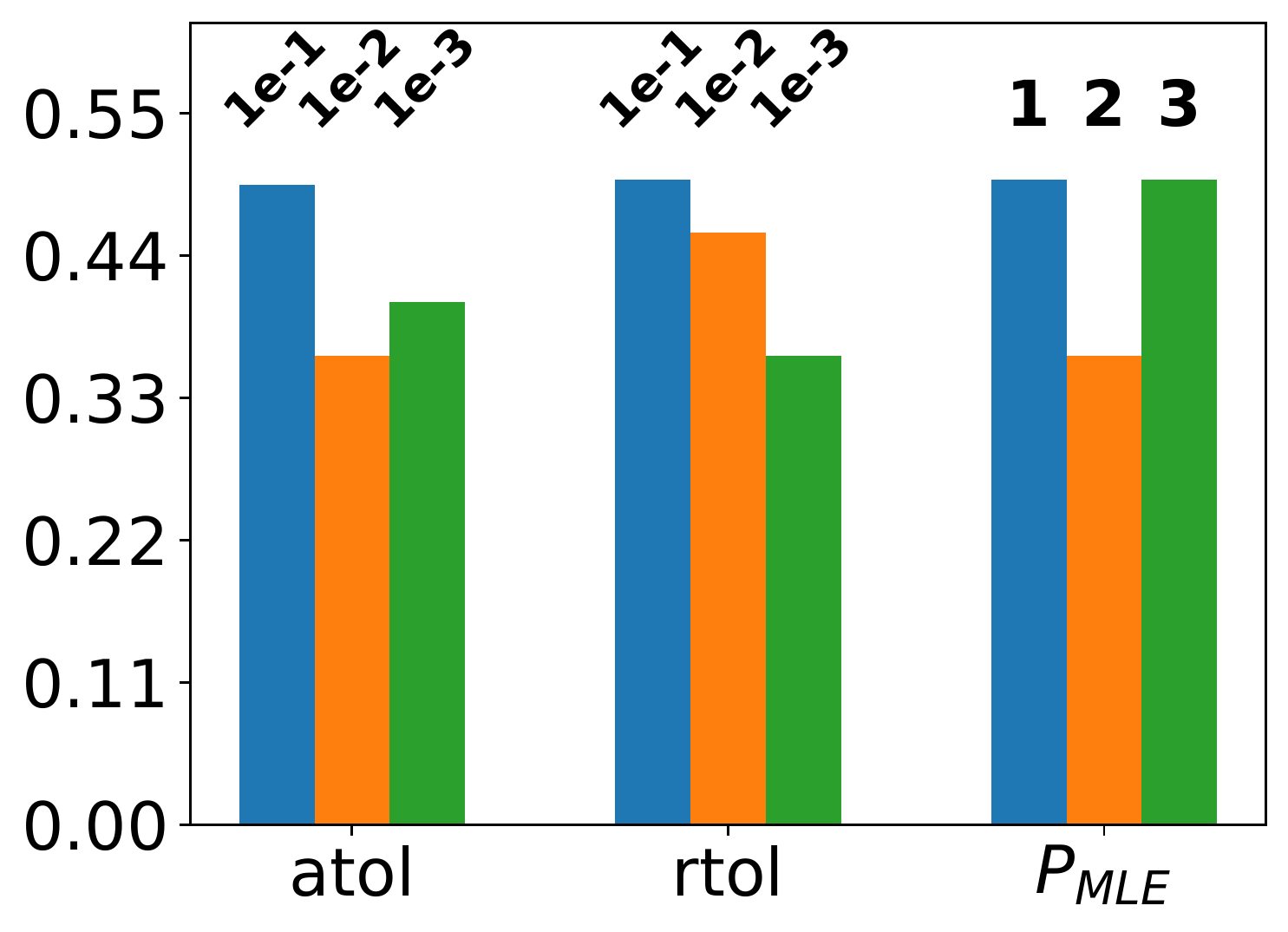}}\hfill
{\includegraphics[width=0.24\columnwidth]{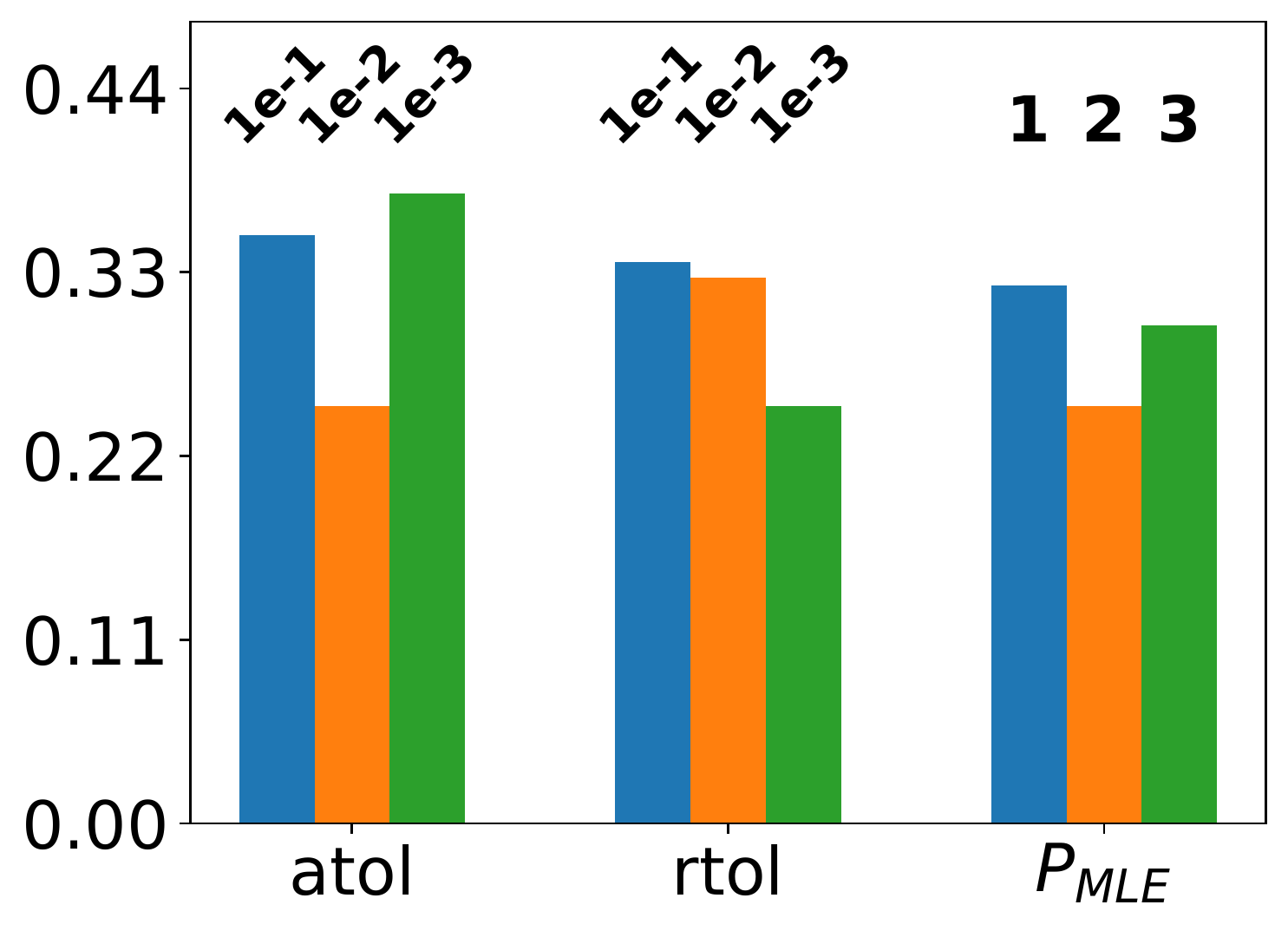}}\hfill
{\includegraphics[width=0.24\columnwidth]{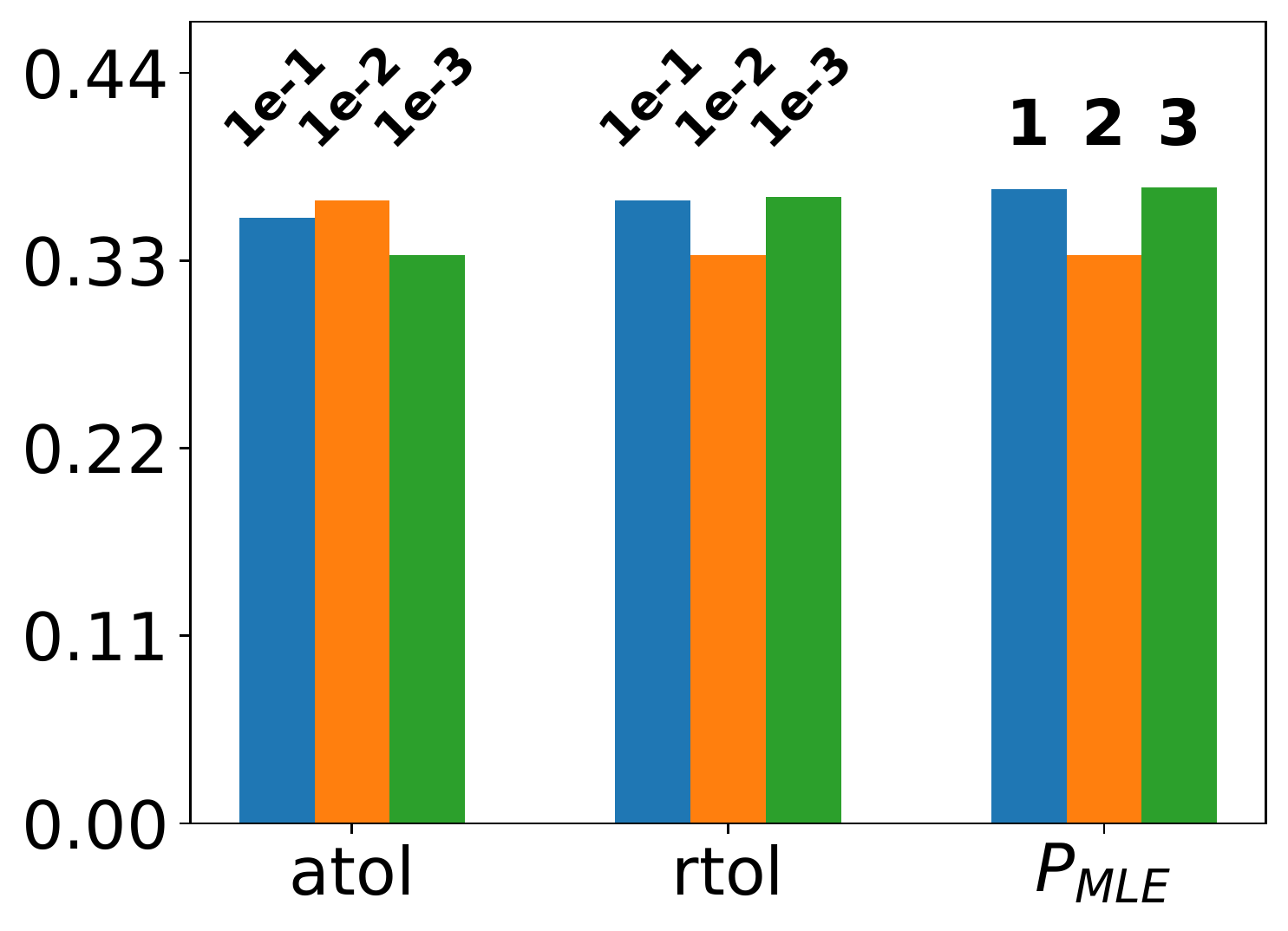}}\hfill
{\includegraphics[width=0.24\columnwidth]{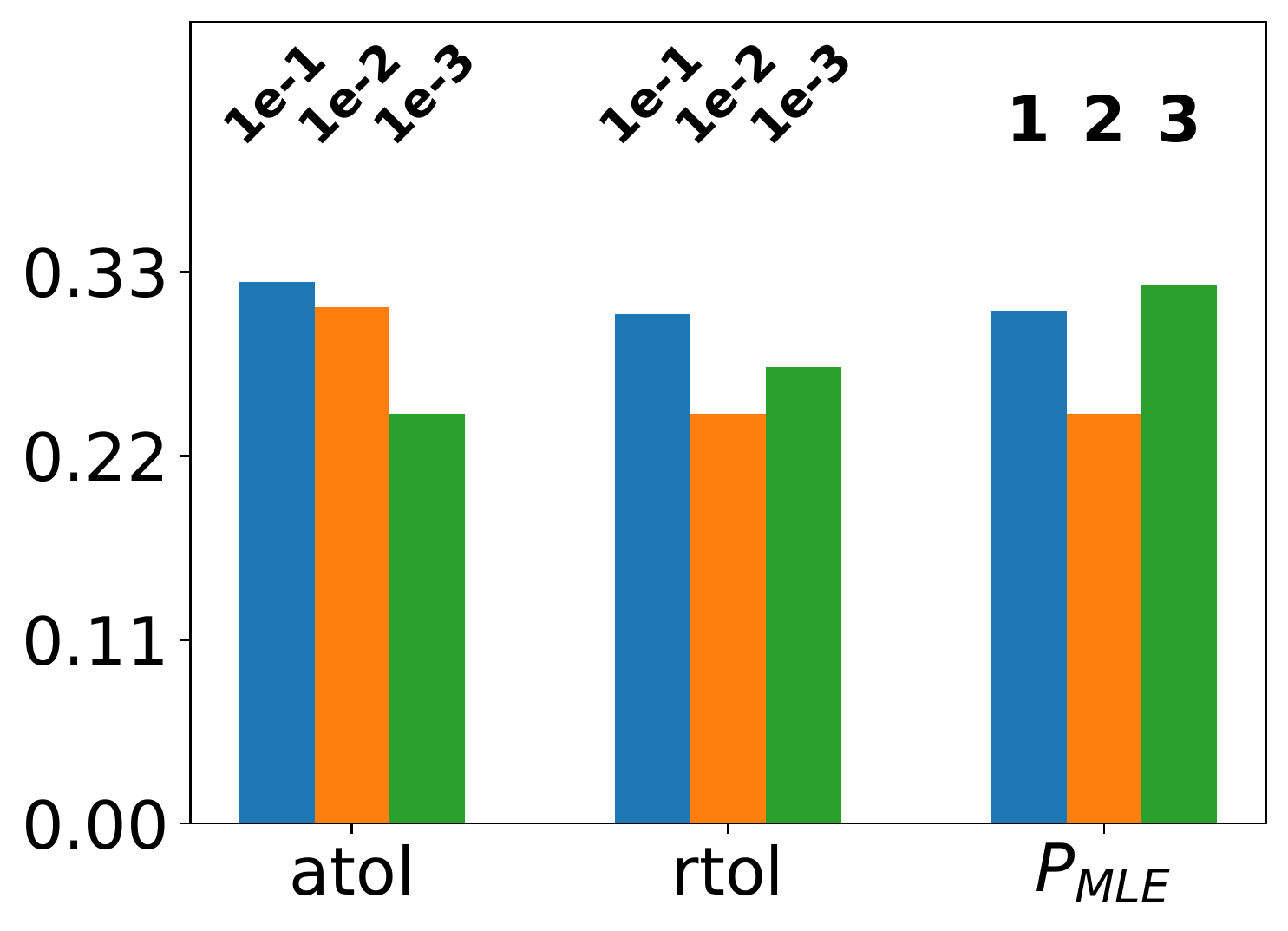}}\hfill
\newline
\centering
\subfigure[Sines]{\includegraphics[width=0.24\columnwidth]{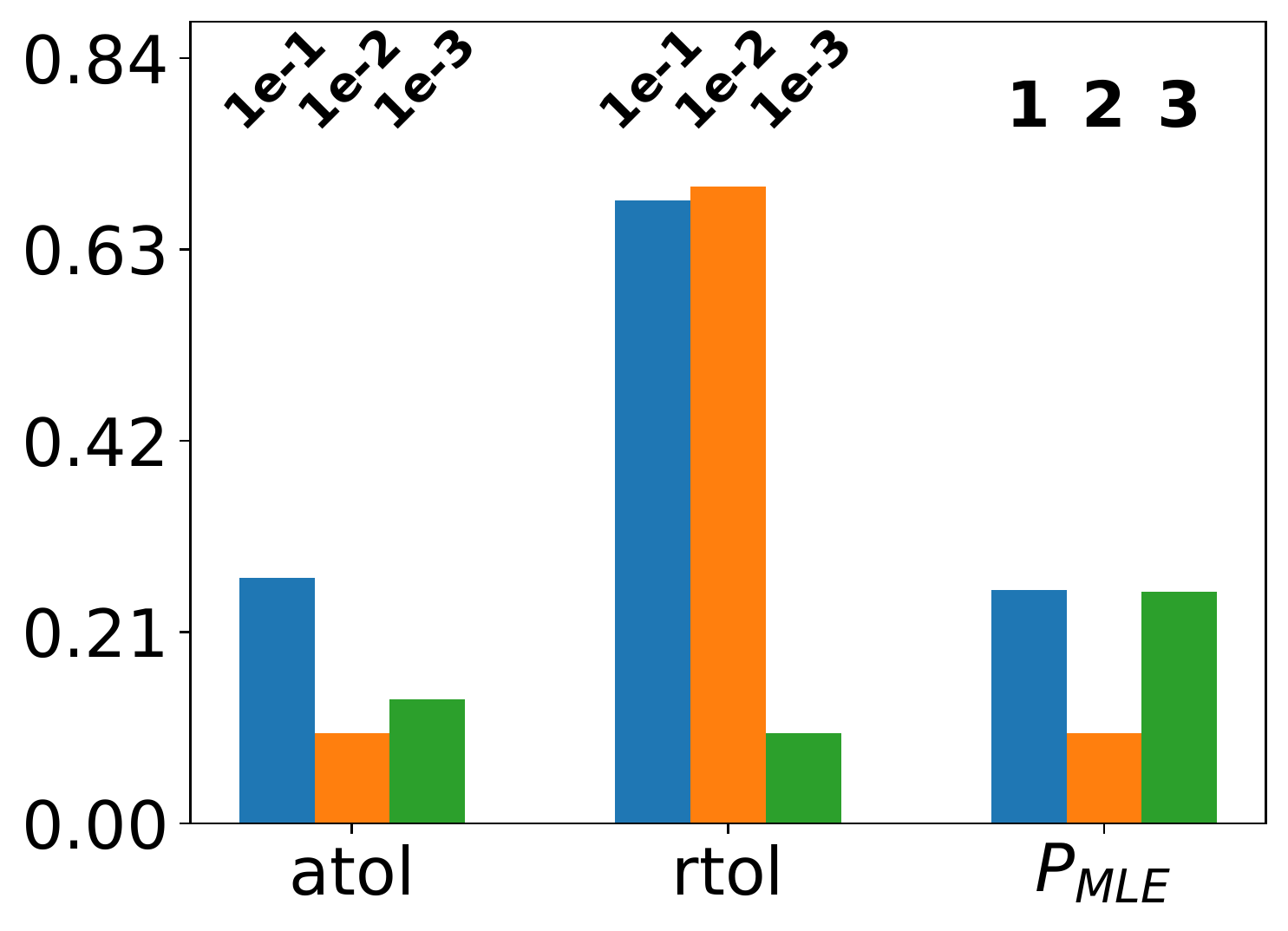}}\hfill
\subfigure[Stocks]{\includegraphics[width=0.24\columnwidth]{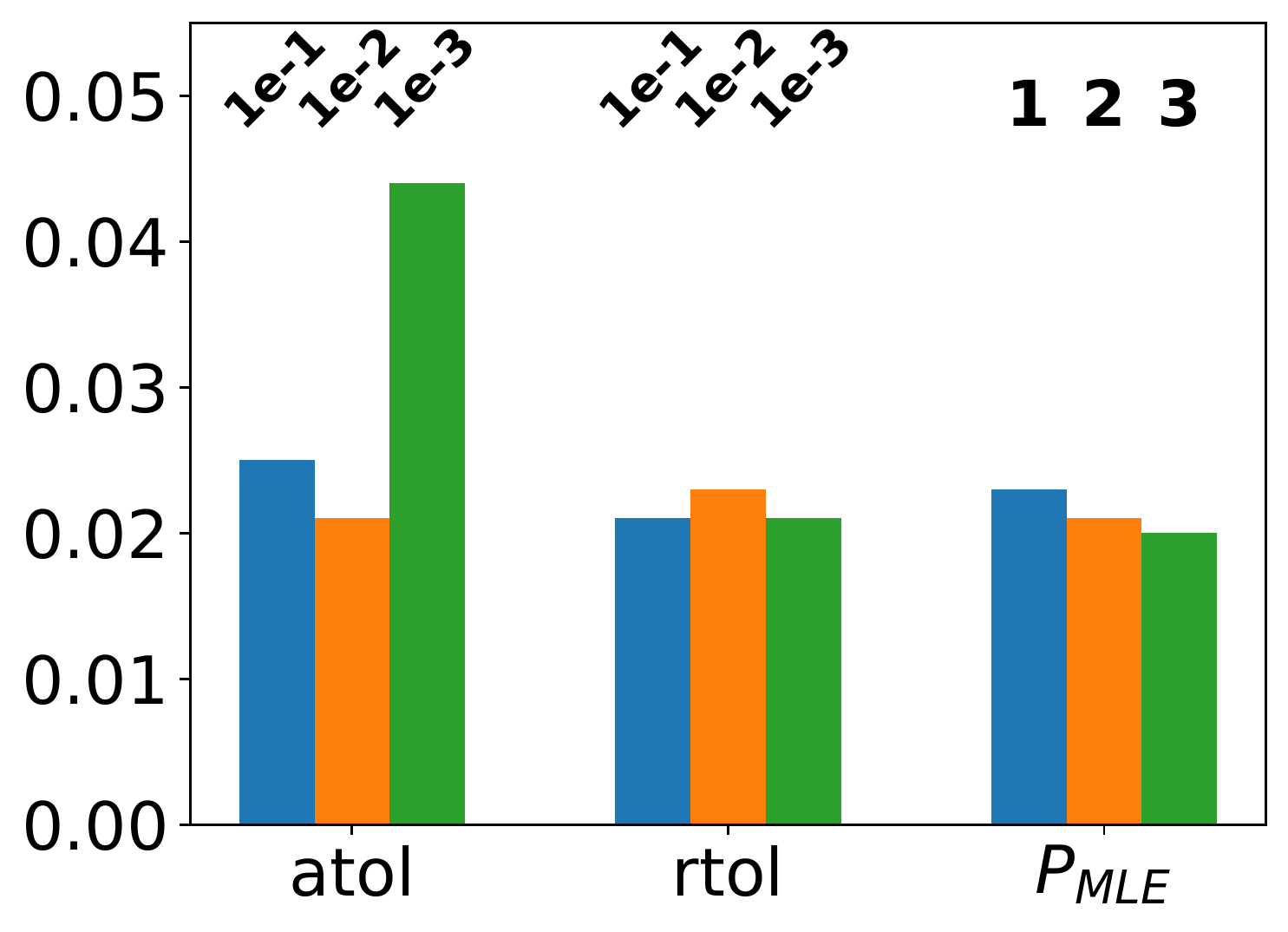}}\hfill
\subfigure[Energy]{\includegraphics[width=0.24\columnwidth]{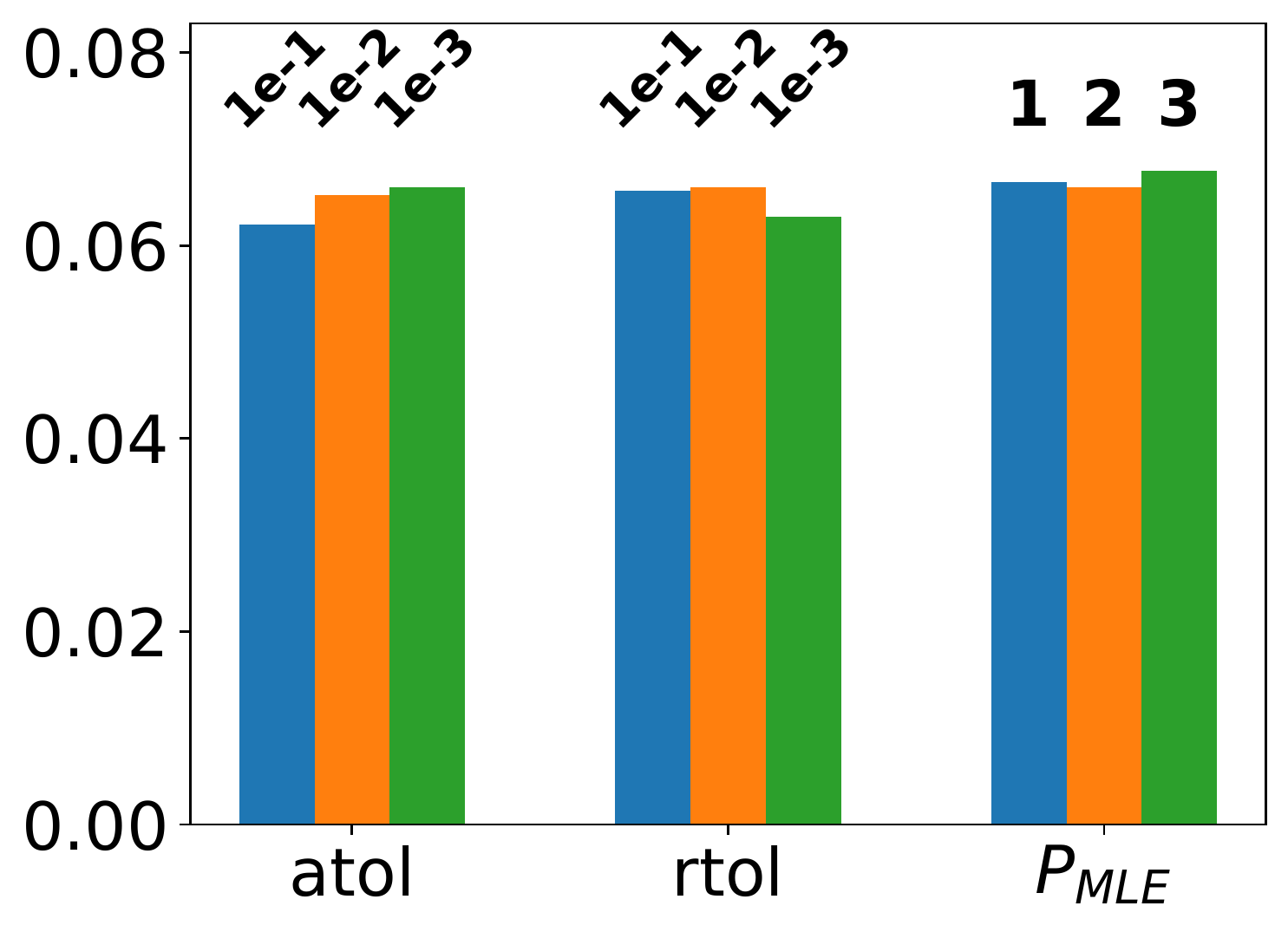}}\hfill
\subfigure[MuJoCo]{\includegraphics[width=0.24\columnwidth]{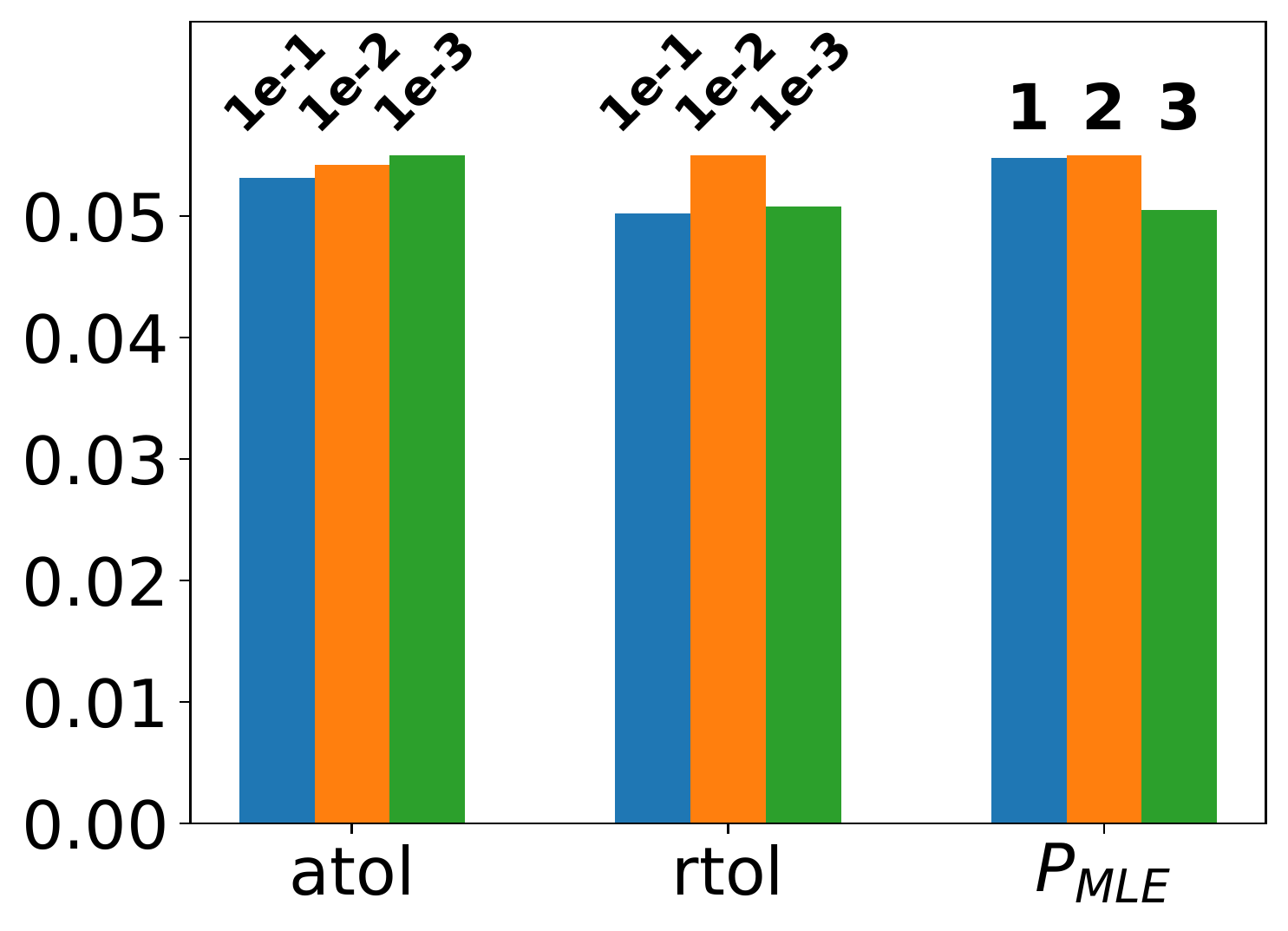}}\hfill
\caption{The sensitivity of the discriminative score (the 1$^{\text{st}}$ row) and predictive score (the 2$^{\text{nd}}$ row) w.r.t. some key hyperparameters for irregular data (dropped 30\%)}
\end{figure}

\begin{figure}[hbt!]
{\includegraphics[width=0.24\columnwidth]{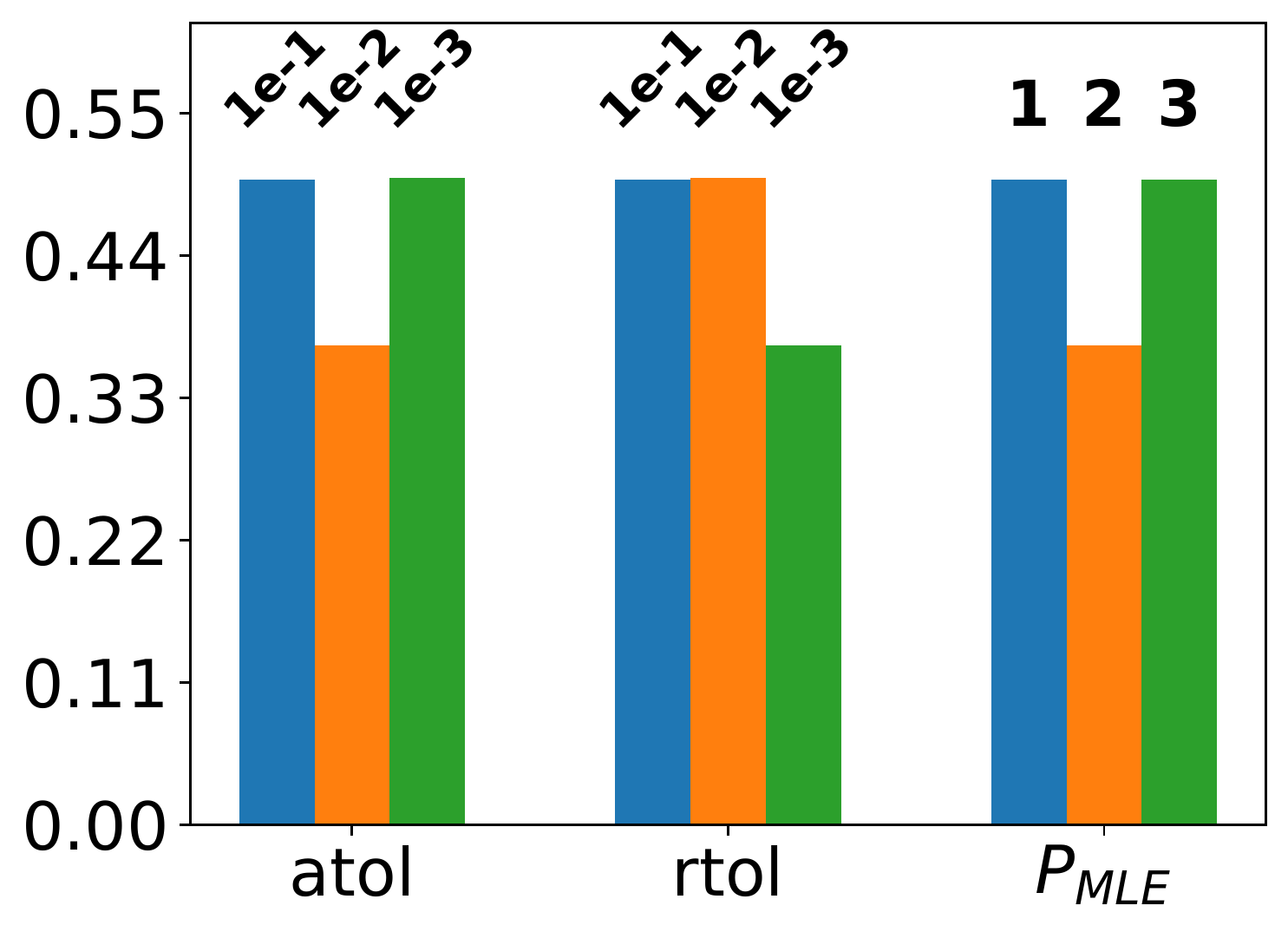}}\hfill
{\includegraphics[width=0.24\columnwidth]{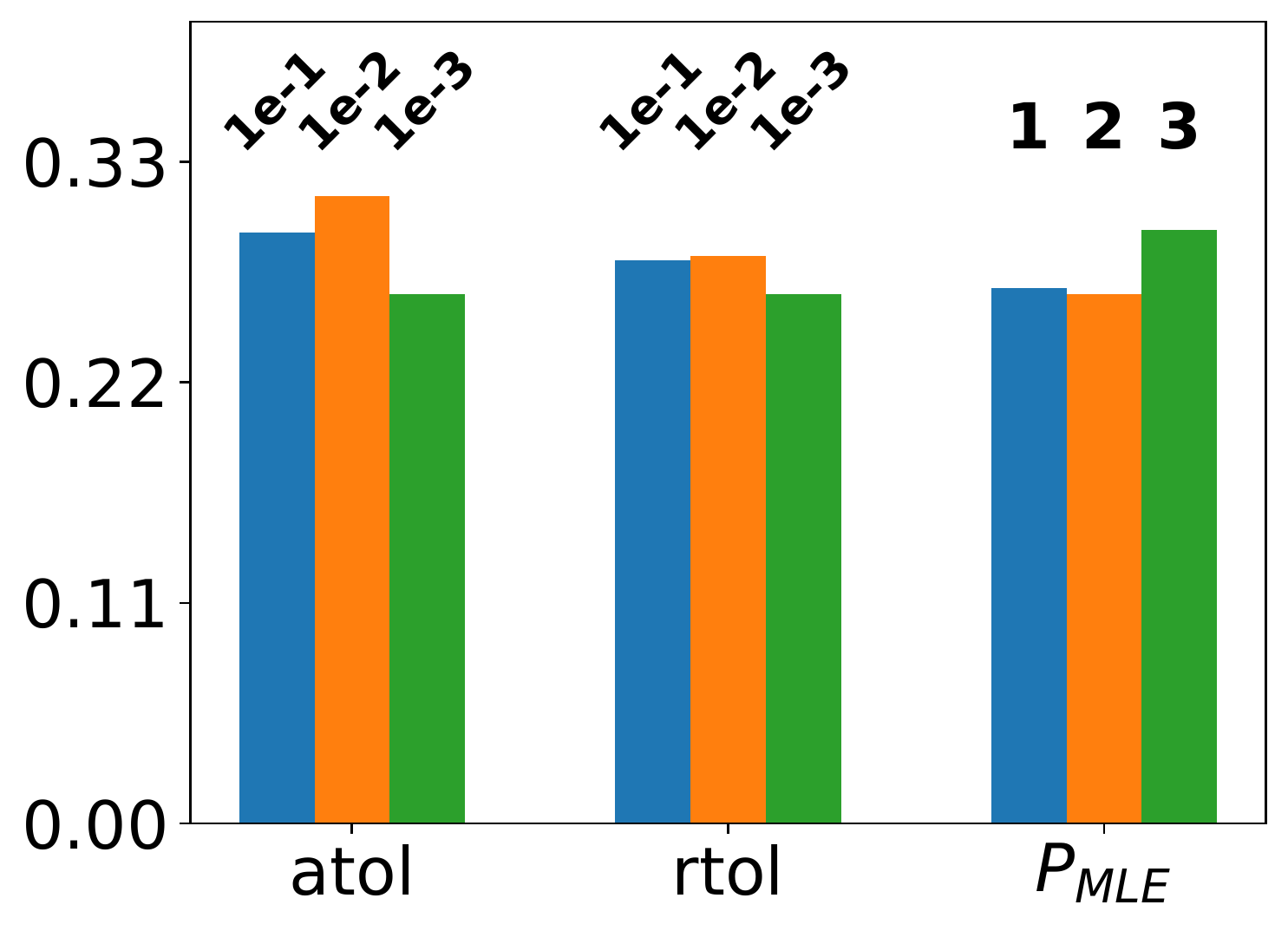}}\hfill
{\includegraphics[width=0.24\columnwidth]{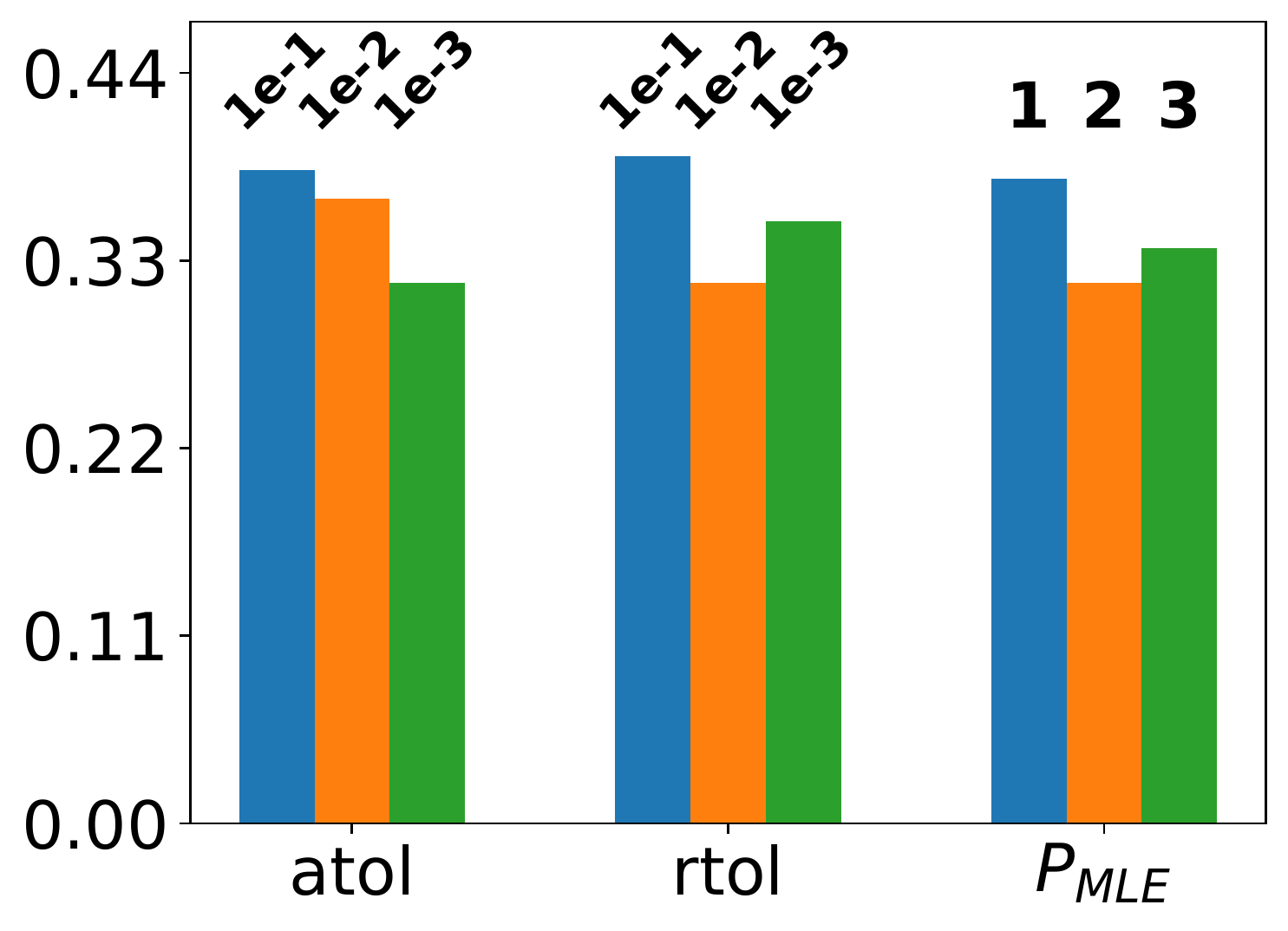}}\hfill
{\includegraphics[width=0.24\columnwidth]{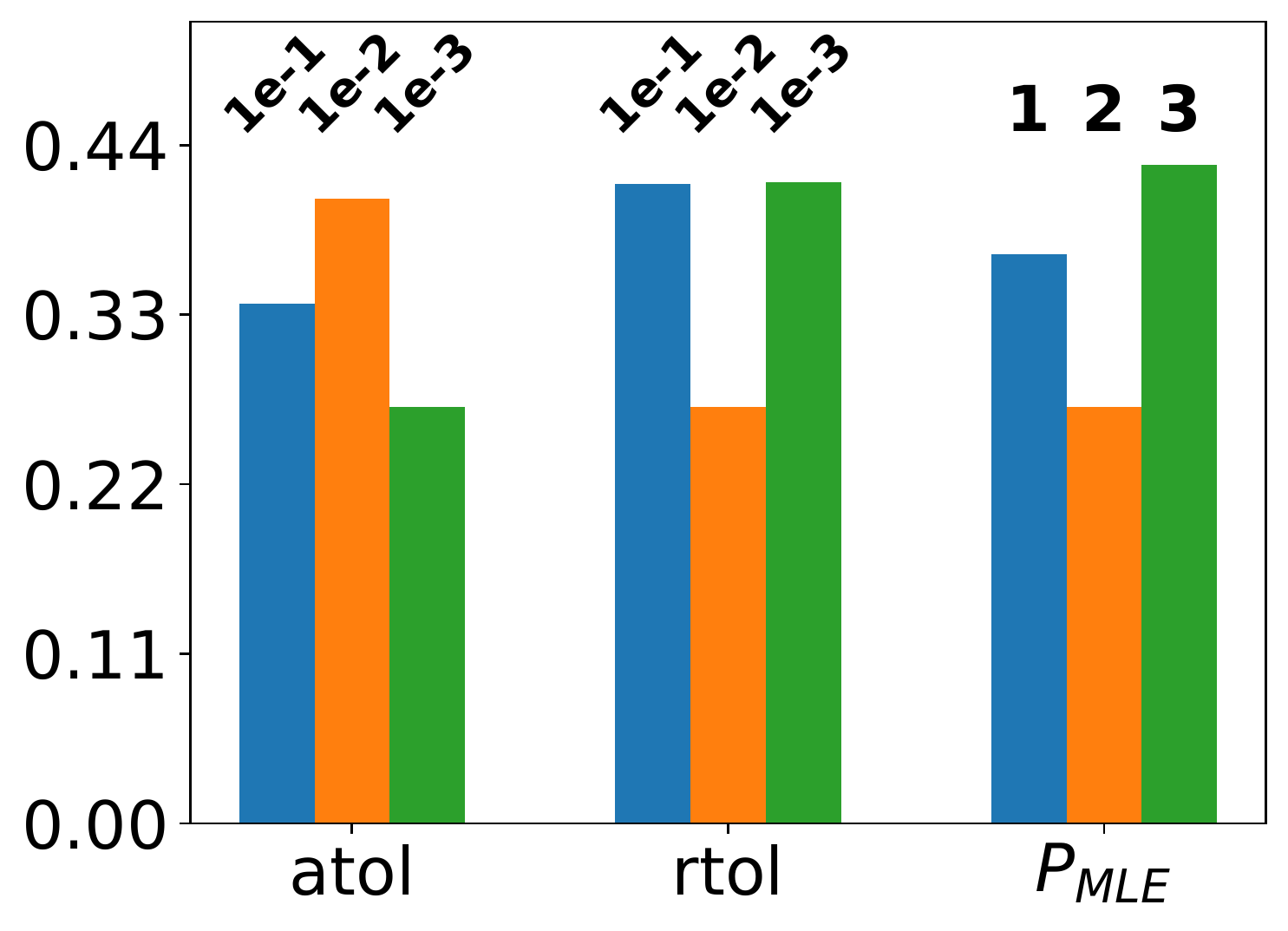}}\hfill
\newline
\centering
\subfigure[Sines]{\includegraphics[width=0.24\columnwidth]{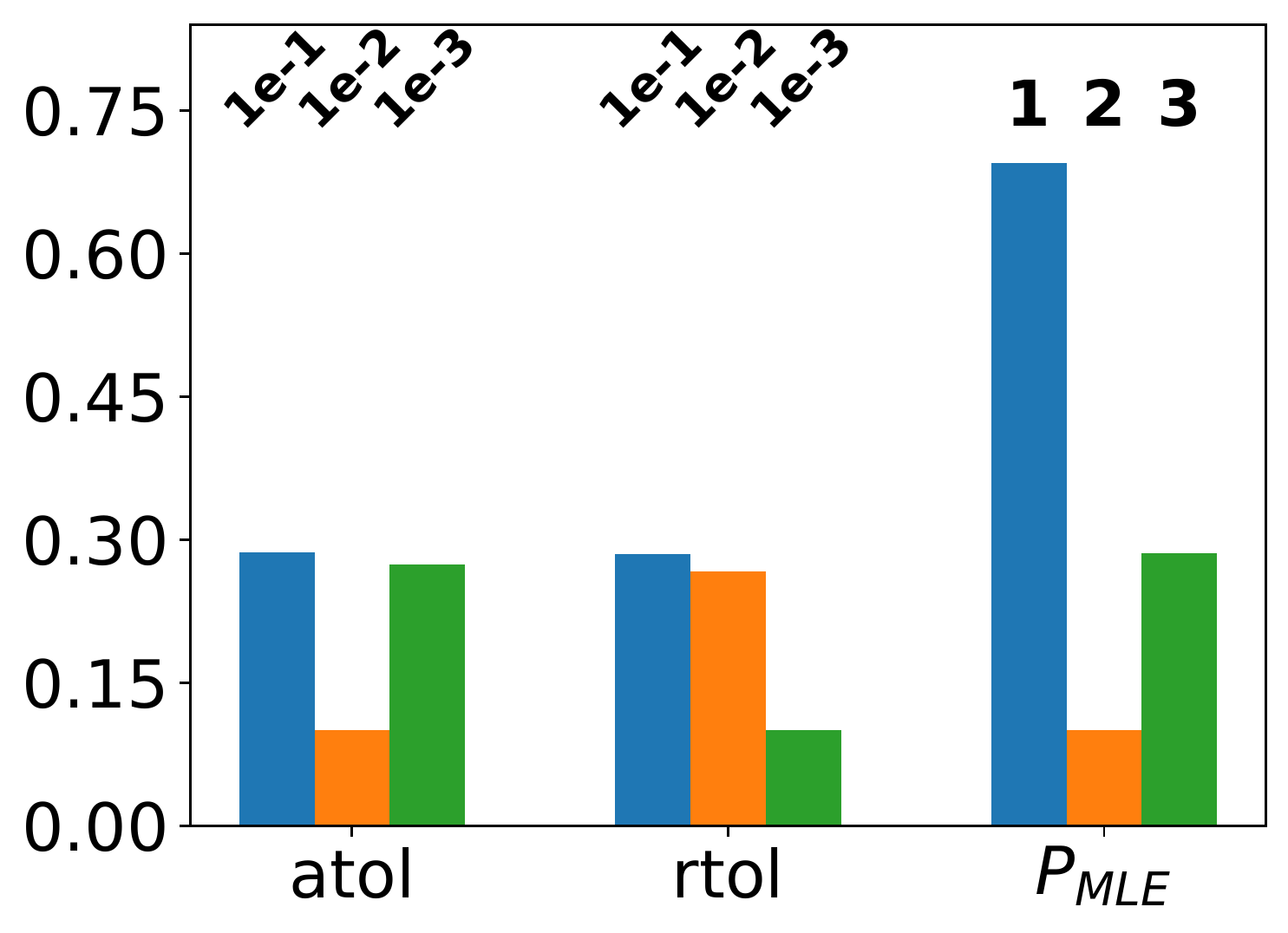}}\hfill
\subfigure[Stocks]{\includegraphics[width=0.24\columnwidth]{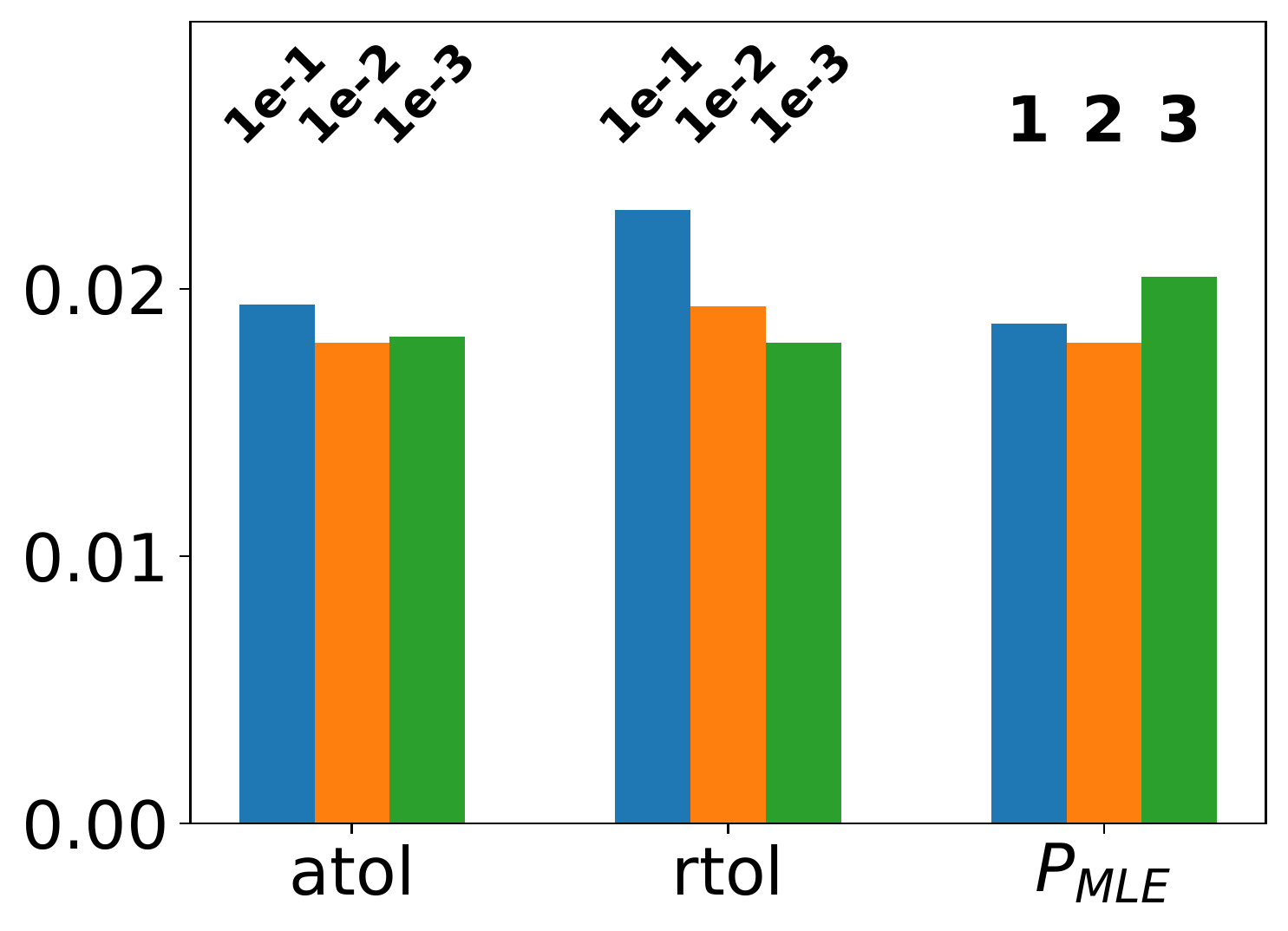}}\hfill
\subfigure[Energy]{\includegraphics[width=0.24\columnwidth]{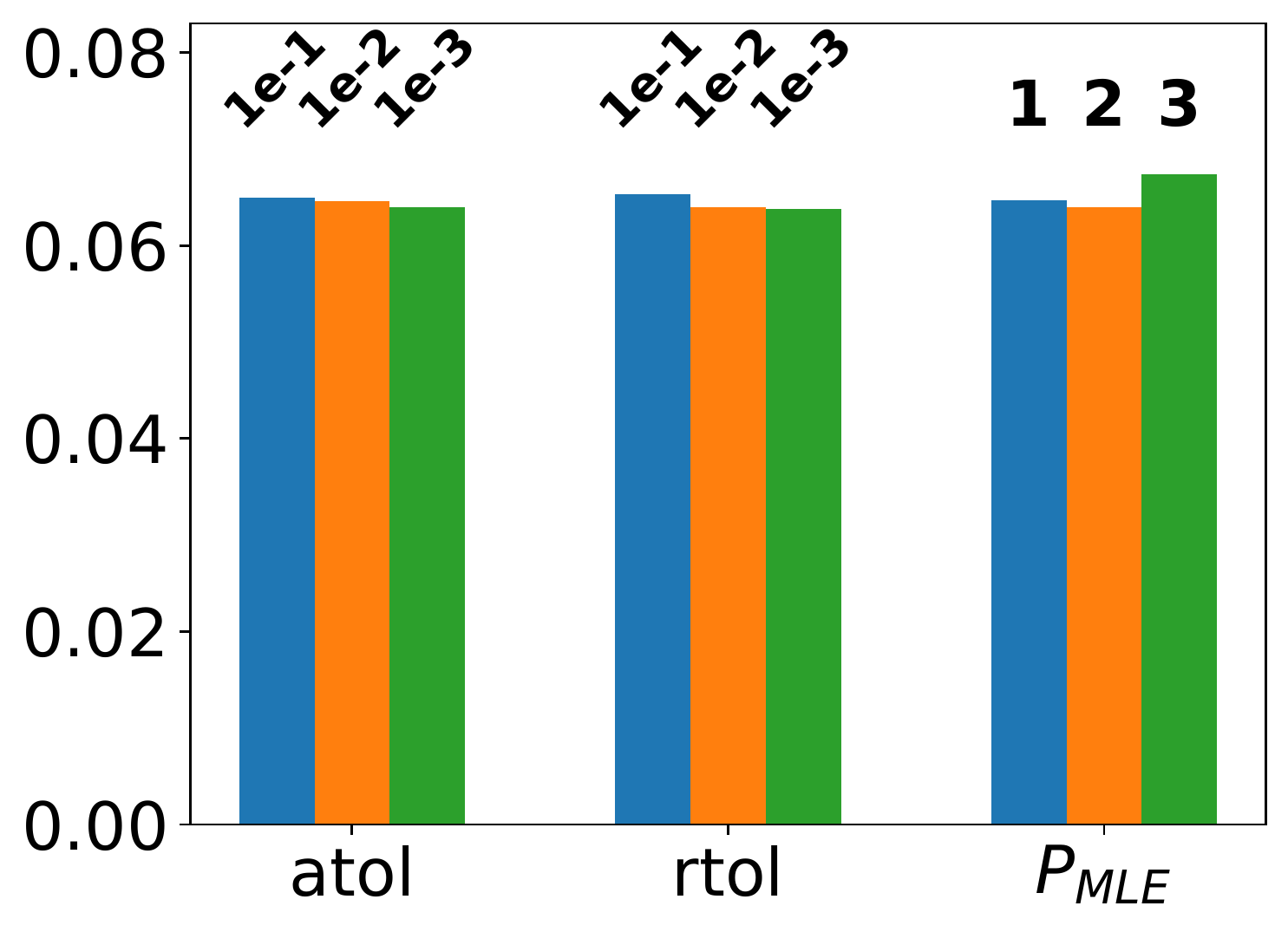}}\hfill
\subfigure[MuJoCo]{\includegraphics[width=0.24\columnwidth]{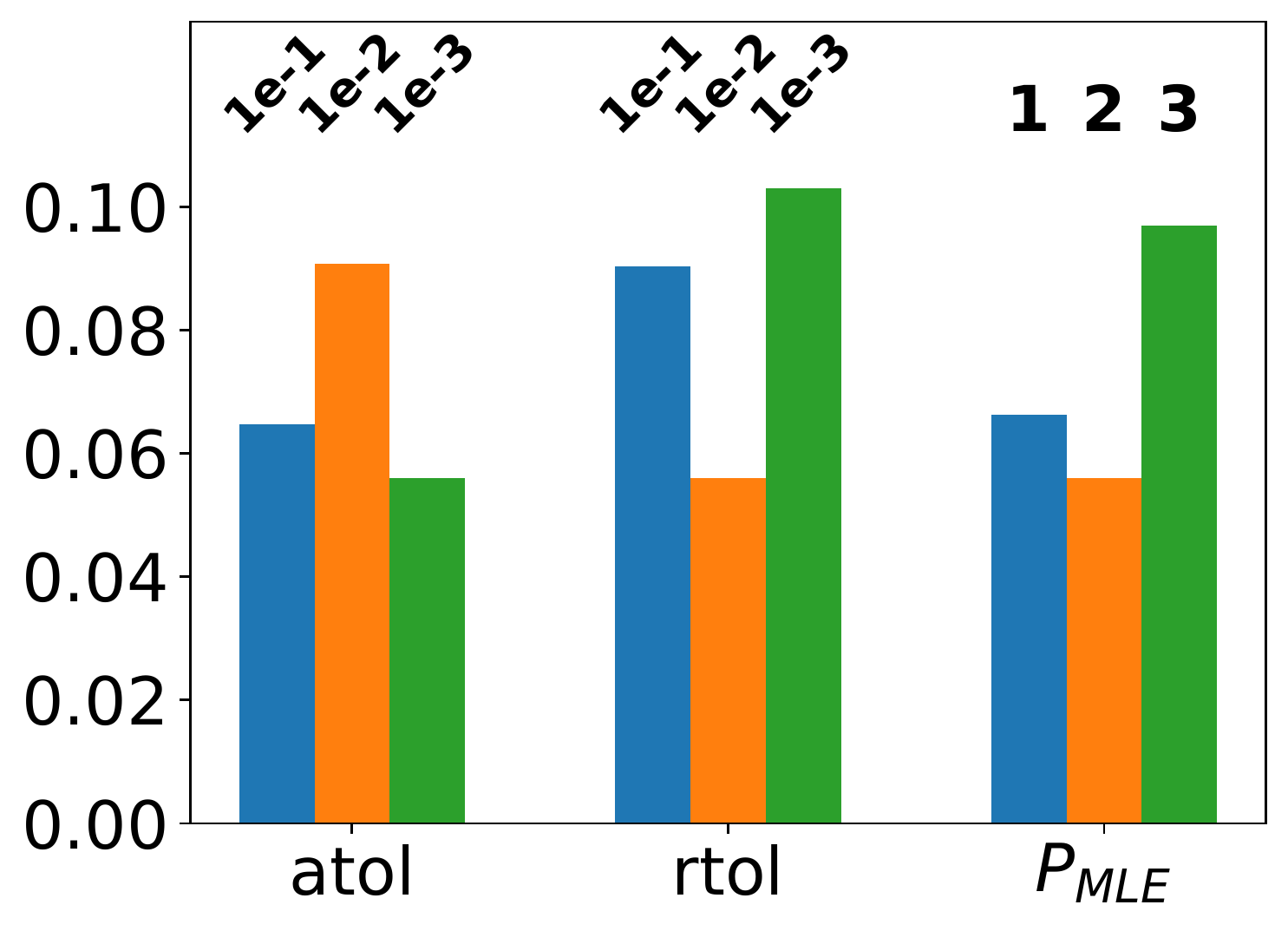}}\hfill
\caption{The sensitivity of the discriminative score (the 1$^{\text{st}}$ row) and predictive score (the 2$^{\text{nd}}$ row) w.r.t. some key hyperparameters for irregular data (dropped 50\%)}
\end{figure}

\begin{figure}[hbt!]
\centering
{\includegraphics[width=0.24\columnwidth]{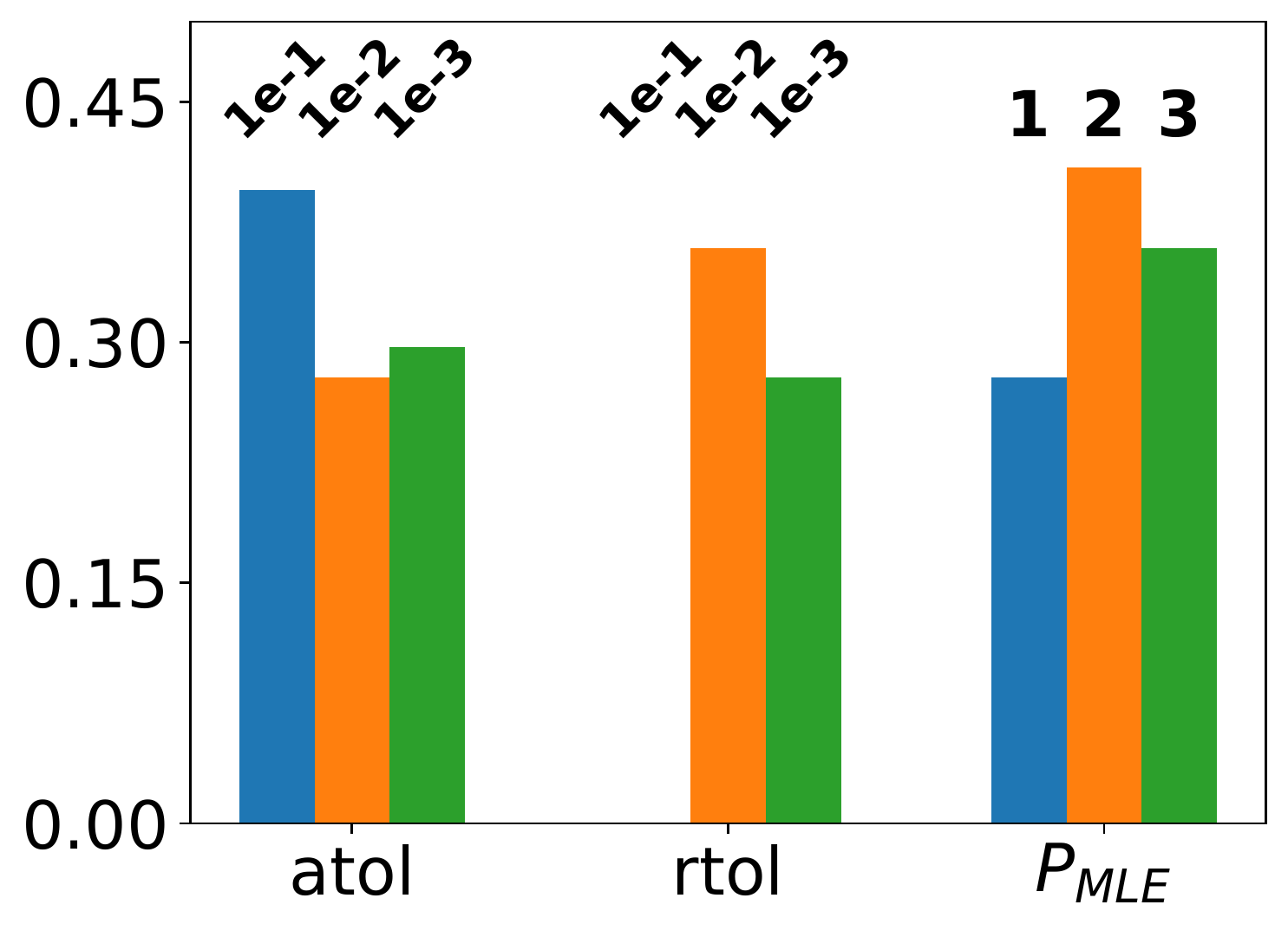}}\hfill
{\includegraphics[width=0.24\columnwidth]{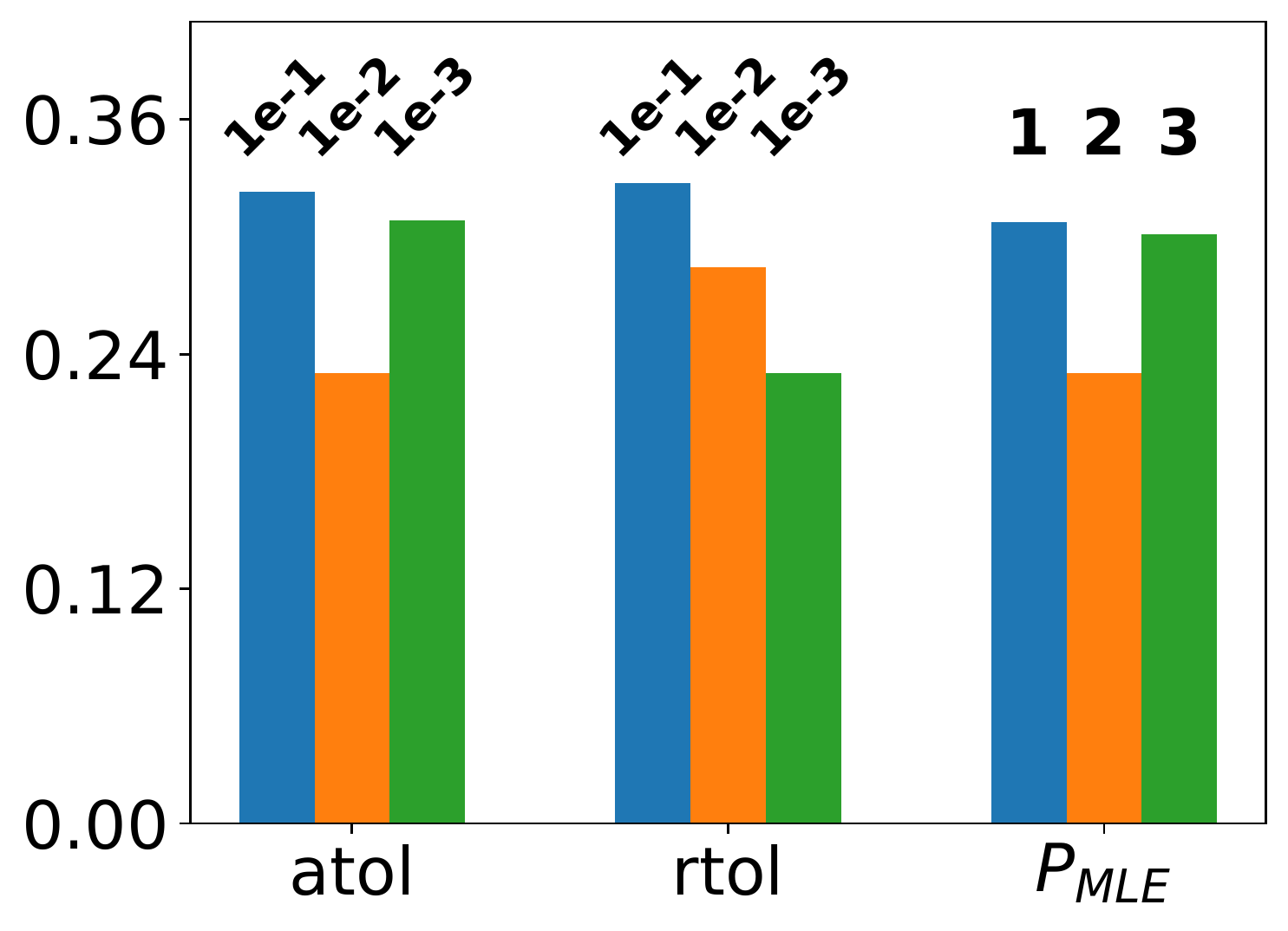}}\hfill
{\includegraphics[width=0.24\columnwidth]{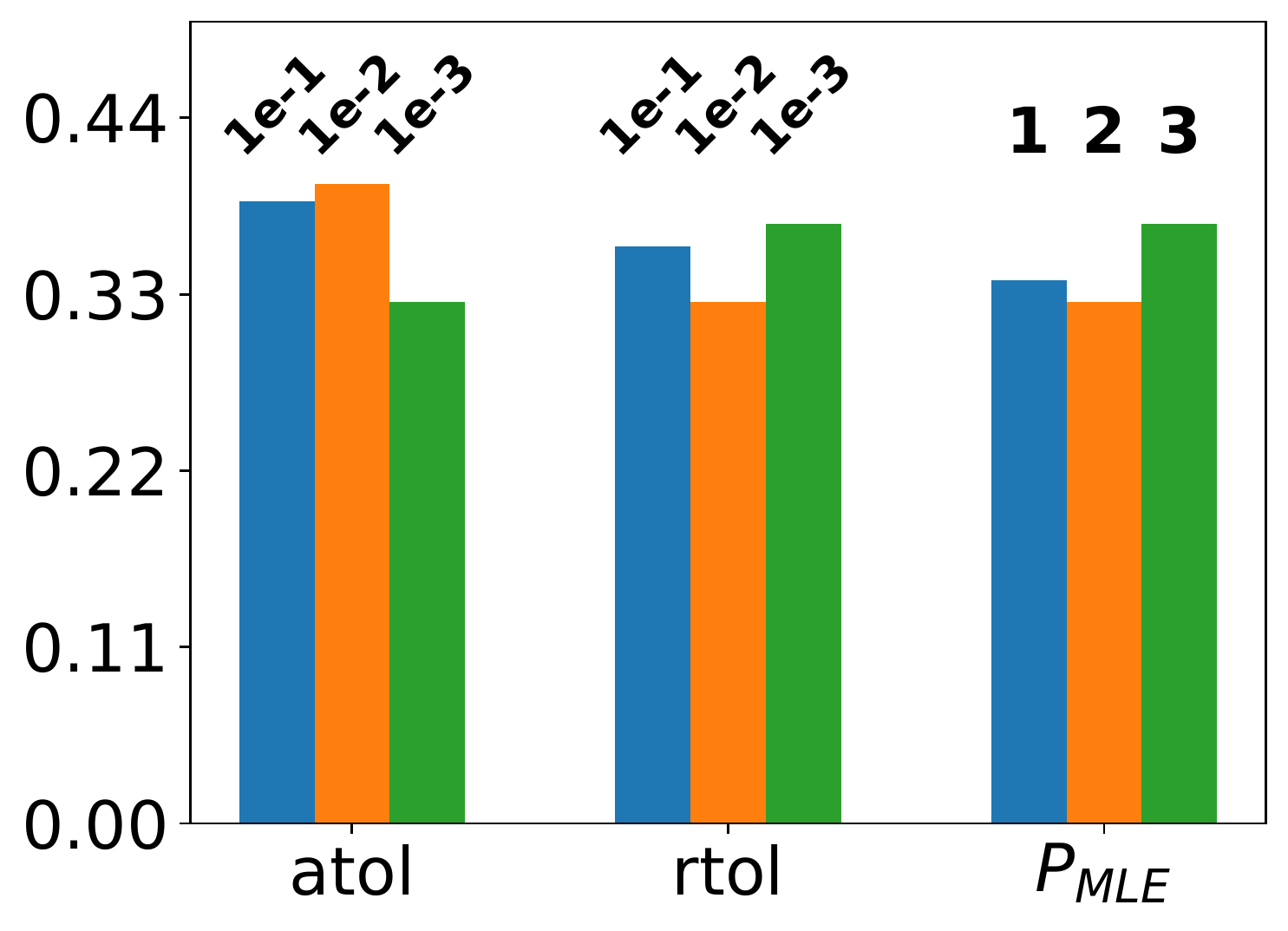}}\hfill
{\includegraphics[width=0.24\columnwidth]{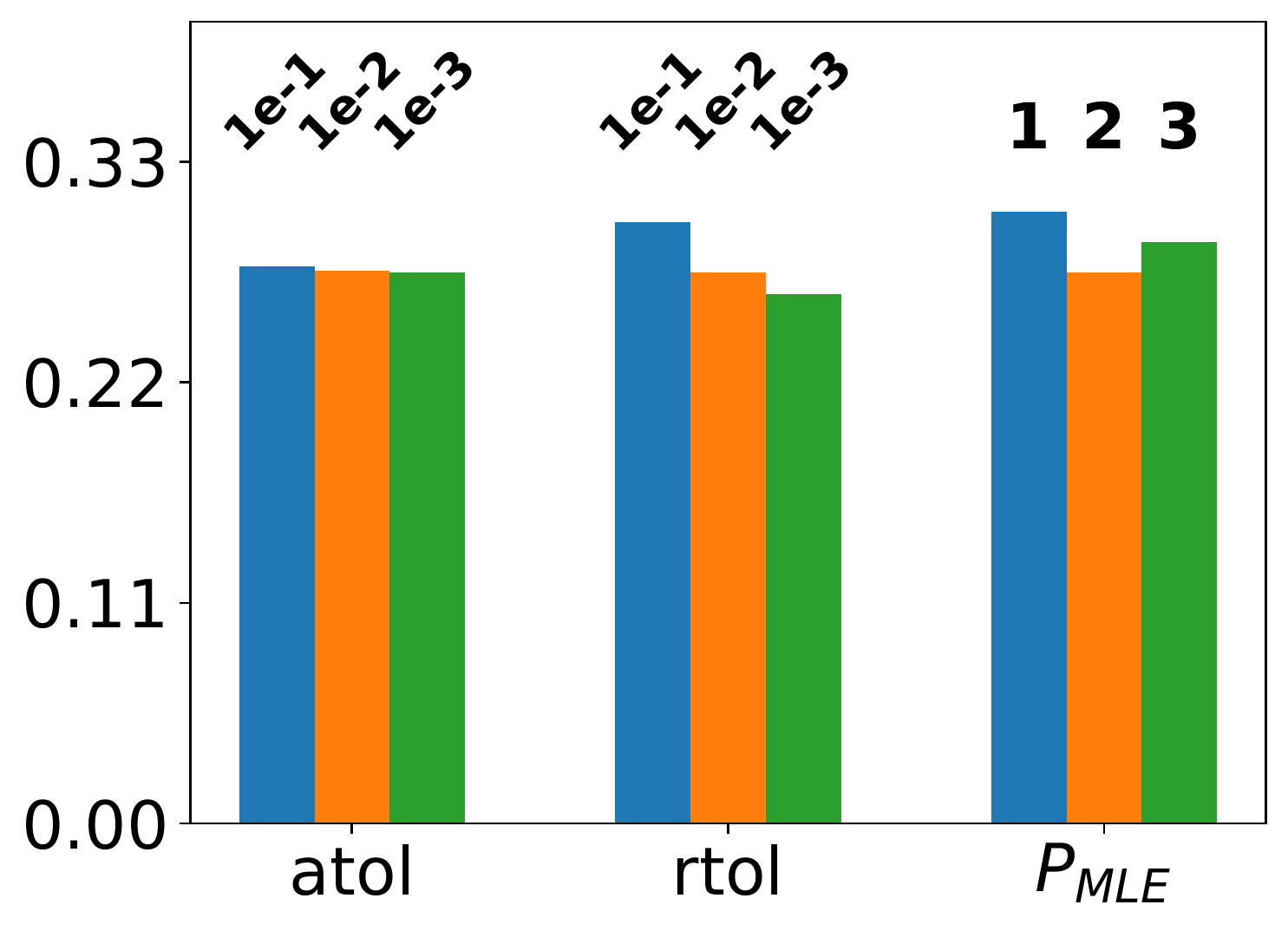}}\hfill
\newline
\centering
\subfigure[Sines]{\includegraphics[width=0.24\columnwidth]{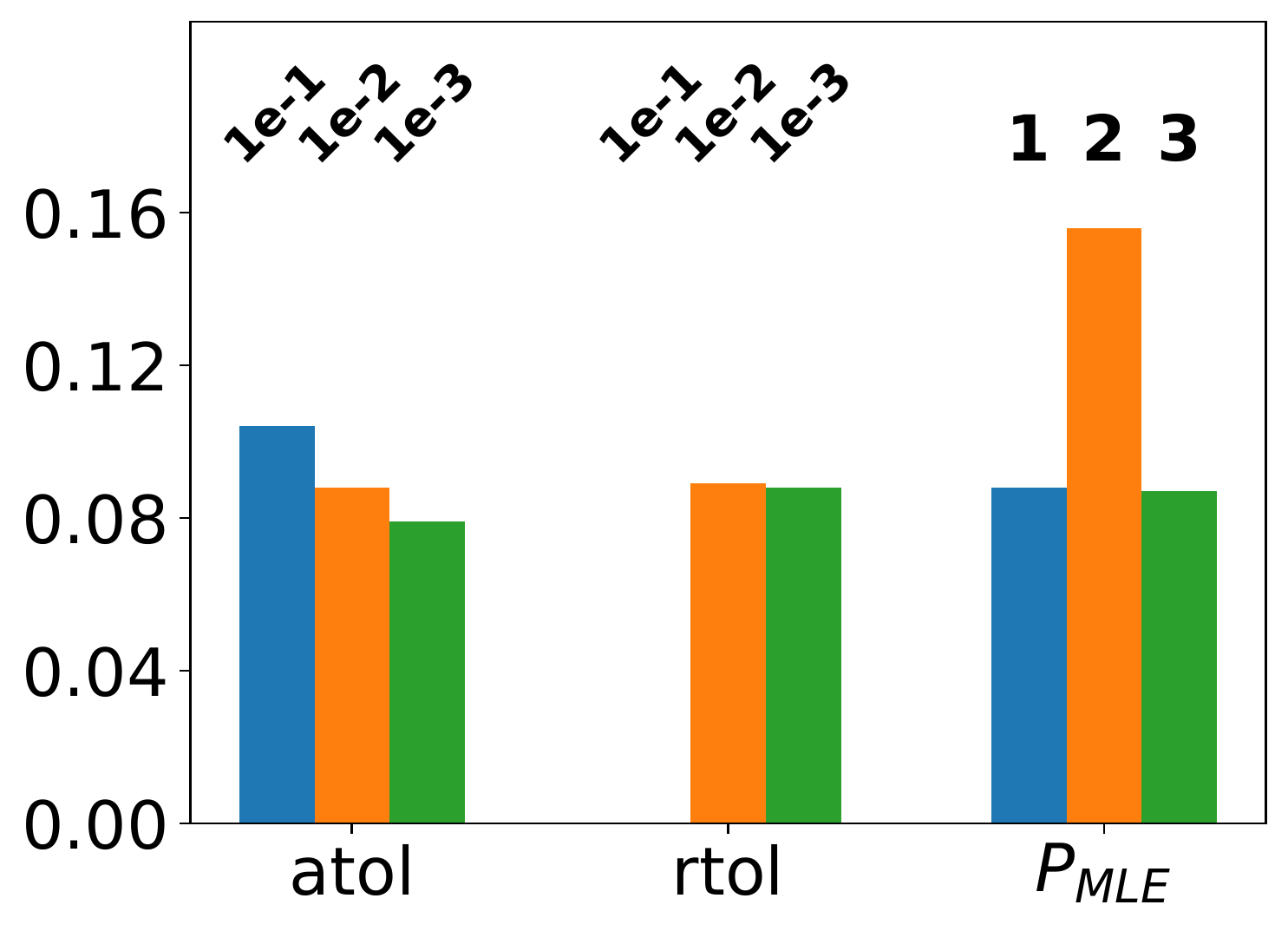}}\hfill
\subfigure[Stocks]{\includegraphics[width=0.24\columnwidth]{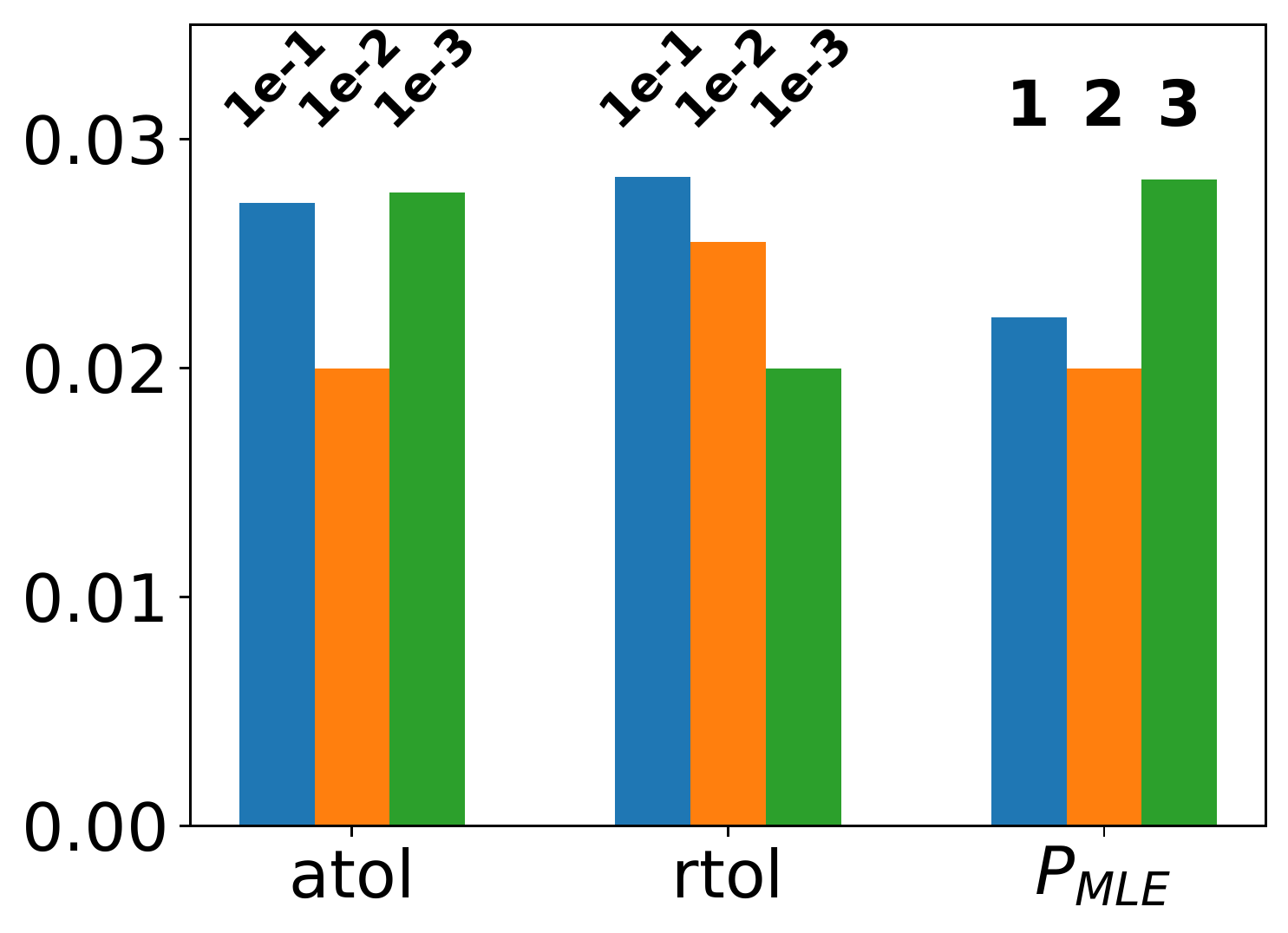}}\hfill
\subfigure[Energy]{\includegraphics[width=0.24\columnwidth]{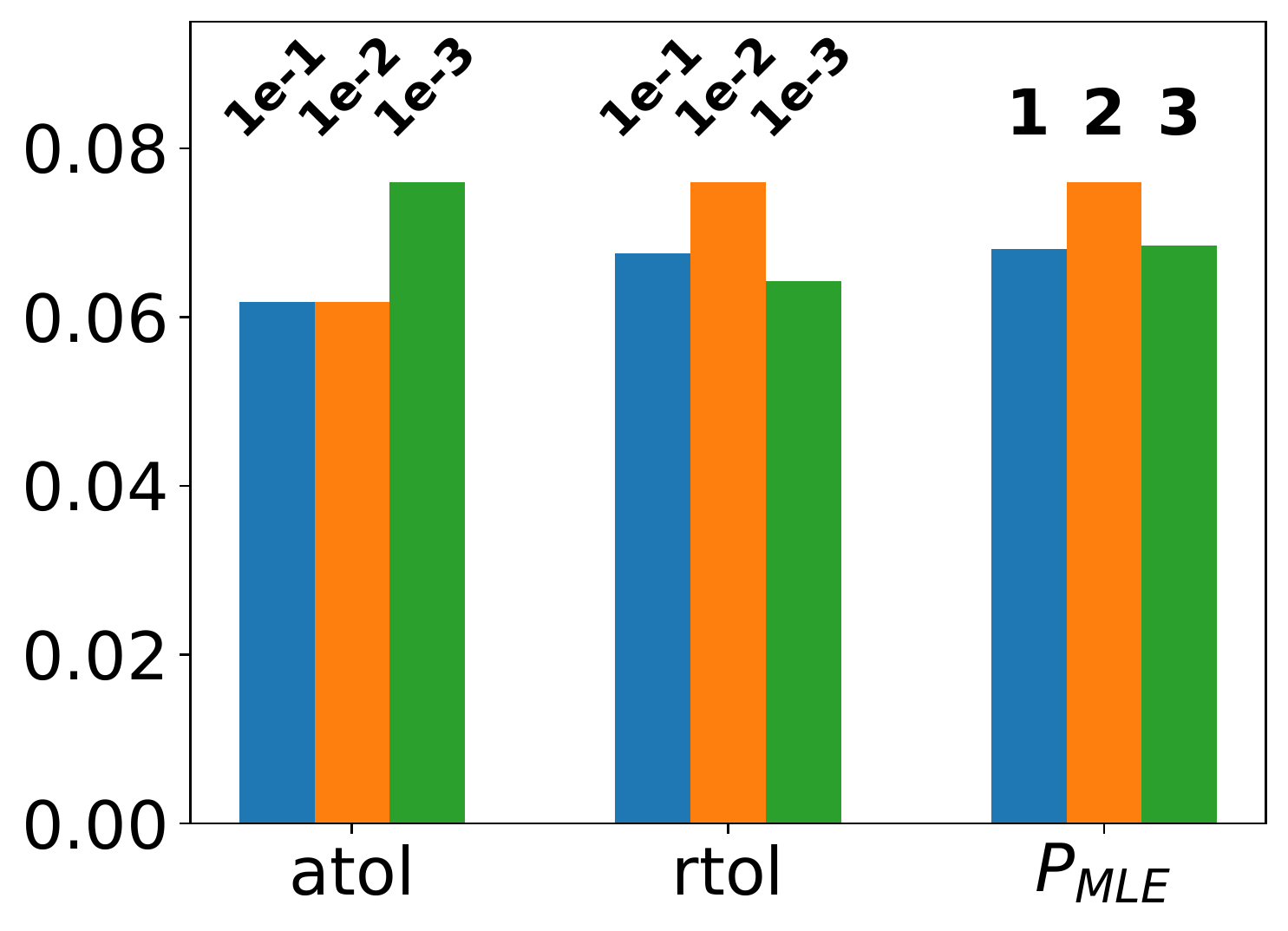}}\hfill
\subfigure[MuJoCo]{\includegraphics[width=0.24\columnwidth]{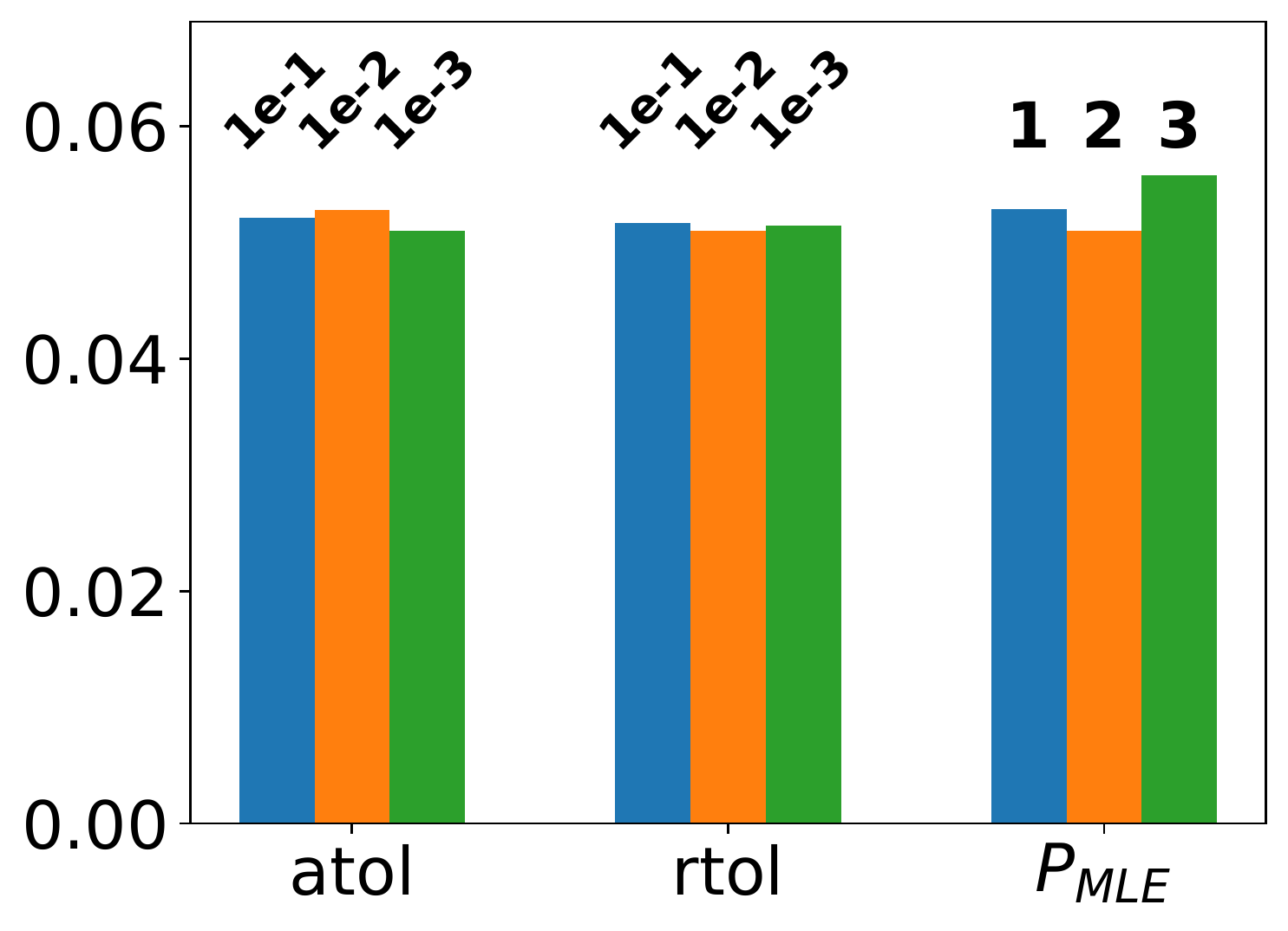}}\hfill
\caption{The sensitivity of the discriminative score (the 1$^{\text{st}}$ row) and predictive score (the 2$^{\text{nd}}$ row) w.r.t. some key hyperparameters for irregular data (dropped 70\%)}
\end{figure}

\clearpage

\section{The best hyperparamter set for GT-GAN}
\label{reproduce}

\begin{itemize}
    \item `atol' means absolute tolerance for the generator.
    \item `rtol' means relative tolerance for the generator.
    \item `$P_{MLE}$'means the period of the negative log-density training for the generator.
    \item `$K_{AE}$' means the autoencoder's pre-training iteration numbers.
    \item `d-layer' means the number of discriminator's GRU layers.
    \item `r-acti' means the last activation function of the decoder.
    \item `reg-recon' means the reconstruction regularization for the generator.
    \item `reg-kinetic' means the kinetic-energy regularization for the generator.
    \item `reg-jacobian' means the Jacobian-norm2 regularization for the generator.
    \item `reg-direct' means the directional-penalty regularization for the generator.
\end{itemize}

\begin{table}[hbt!]
\begin{minipage}{.99\linewidth}
\centering
\small
\setlength{\tabcolsep}{1pt}
\captionof{table}{\label{tab:hyperparameter}The best hyperparameters}
\begin{tabular}{ccccccccccccc}
\hline
method & Data & atol & rtol & $P_{MLE}$ &$K_{AE}$ & d-layer & r-acti &  reg-recon & reg-kinetic & reg-jacobian &  reg-direct \\
\hline
\multirow{4}{*}{\begin{tabular}[c]{@{}c@{}}GT-GAN \\ (Regular)\end{tabular}}
& Sines & 1e-2 & 1e-3& 1 &5000& 1& softplus & 0.01& 0.05& 0.1&0.1\\
& Stocks  & 1e-2 & 1e-3& 2& 10000 &1& softplue& 0.01 &0.01 &0.05  &0.01\\
&Energy & 1e-3 & 1e-2 & 2 & 5000 & 2 & sigmoid & 0.01 & 0.5 & 0.1 &0.01 \\
&MuJoCo & 1e-3 & 1e-2 & 2 & 5000 & 2 & sigmoid & 0.01 & 0.05 & 0.01 &0.01\\
\hline
\multirow{4}{*}{\begin{tabular}[c]{@{}c@{}}GT-GAN \\ (Dropped 30\%)\end{tabular}}
& Sines & 1e-2 &1e-3 &2 &5000 &1 &softplus &0.01 &0.05& 0.01  &0.01\\
& Stocks  &1e-2 &1e-3& 2& 10000 &1 &softplue &0.01& None& None  &0.05\\
&Energy & 1e-3 & 1e-2 & 2 & 5000 & 2 & sigmoid  & 0.01 & 0.5 & 0.1& 0.01\\
&MuJoCo & 1e-3 & 1e-2 & 2 & 2500 & 2 & sigmoid & 0.01 & 0.5 & 0.1 & 0.01\\
\hline
\multirow{4}{*}{\begin{tabular}[c]{@{}c@{}}GT-GAN \\ (Dropped 50\%)\end{tabular}}
& Sines &1e-2 & 1e-3 & 2 & 5000 & 2 & softplus &0.01 &0.05& 0.01  &0.01\\
& Stocks  & 1e-3 &1e-3 &2& 10000 &1& softplue & None &0.05& 0.01  &0.05\\
&Energy & 1e-3 & 1e-2 & 2 & 5000 & 2 & sigmoid  & 0.01 & 0.5 & 0.1& 0.01\\
&MuJoCo &1e-3 & 1e-2 & 2 & 1500 & 2 & sigmoid  & 0.1 & 0.1 & 0.01& 0.01\\
\hline
\multirow{4}{*}{\begin{tabular}[c]{@{}c@{}}GT-GAN \\ (Dropped 70\%)\end{tabular}}
& Sines & 1e-2 &1e-3 &2& 5000& 1 &softplus& 0.01& 0.05& 0.01&0.01\\
& Stocks  & 1e-2 &1e-3 &1& 10000 &1 &softplue &None &0.05 &0.01 &0.05\\
&Energy & 1e-3 & 1e-2 & 2 & 2500 & 2 & sigmoid  & 0.01 & 0.5 & 0.1 & 0.01\\
&MuJoCo & 1e-3 & 1e-2 & 2 & 2500 & 2 & sigmoid& 0.01 & 0.5 & 0.1  & 0.01 \\
\hline
\end{tabular}
\end{minipage}
\end{table}

\section{Model size \& training time comparison}
In Table~\ref{tab:Statistical comparison of models}, we report the model size and training time of our method and TimeGAN, one of the best performing baseline. As shown, our model has much smaller numbers of parameters than TimeGAN. However, it take much longer time to train our model than TimeGAN. This is mainly because we need to solve various differential equations, which is not needed for TimeGAN. The memory requirements are more or less the same in both models. Therefore, these exist pros and cons for our method in comparison with the state-of-the-art baseline.

\begin{table}[hbt!]
\begin{minipage}{.99\linewidth}
\centering
\small
\setlength{\tabcolsep}{1pt}
\caption{\label{tab:Statistical comparison of models}Comparison of model size and training time}
\begin{tabular}{ccccccccc}
\hline
 & \multicolumn{2}{c}{Sines} & \multicolumn{2}{c}{Stocks} & \multicolumn{2}{c}{Energy} & \multicolumn{2}{c}{MuJoCo} \\ \hline
Model & \multicolumn{1}{c}{GT-GAN} & TimeGAN & \multicolumn{1}{c}{GT-GAN} & TimeGAN & \multicolumn{1}{c}{GT-GAN} & TimeGAN & \multicolumn{1}{c}{GT-GAN} & TimeGAN \\ \hline
Parameter & \multicolumn{1}{c}{41,913} & 34,026 & \multicolumn{1}{c}{41,776} & 48,775 & \multicolumn{1}{c}{57,104} & 1,043,179 & \multicolumn{1}{c}{47,346} & 264,447 \\ 
Memory (MB) & \multicolumn{1}{c}{1,675} & 1,419 & \multicolumn{1}{c}{1,653} & 1,423 & \multicolumn{1}{c}{1,839} & 1,611 & \multicolumn{1}{c}{1,655} & 1,546 \\
Training Time (HH:MM) & \multicolumn{1}{c}{10:12} & 2:56 & \multicolumn{1}{c}{12:20} & 2:59 & \multicolumn{1}{c}{10:39} & 3:37 & \multicolumn{1}{c}{13:12} & 3:10 \\ \hline
\end{tabular}
\end{minipage}
\end{table}

\section{Visualizations with t-SNE and data distribution}\label{a:visualization}
\begin{figure}[ht]
    \centering
    {\includegraphics[width=0.27\columnwidth]{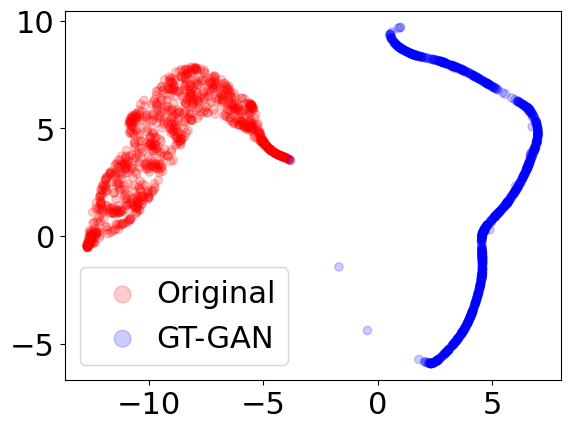}}\hfill
    {\includegraphics[width=0.27\columnwidth]{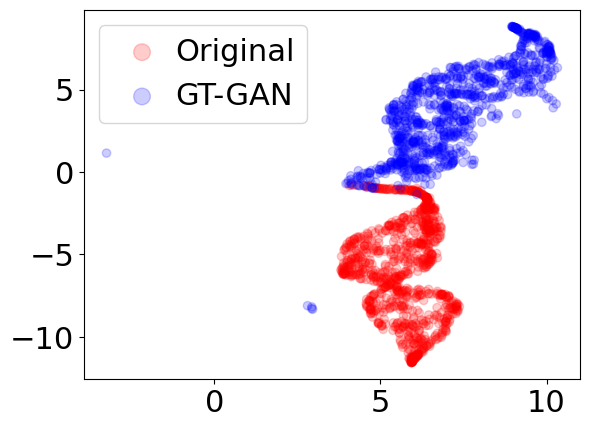}}\hfill
    {\includegraphics[width=0.27\columnwidth]{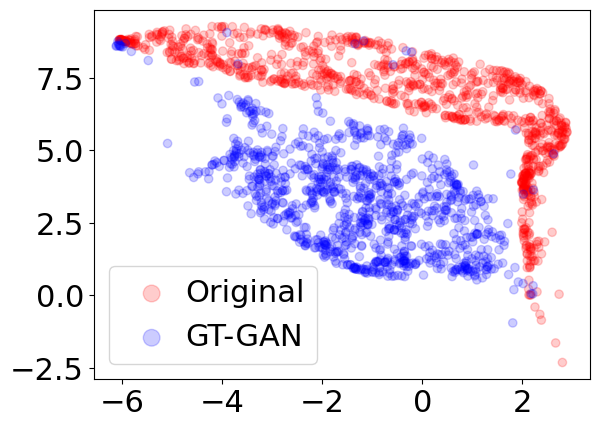}}\hfill
    \newline
    \centering
    {\includegraphics[width=0.27\columnwidth]{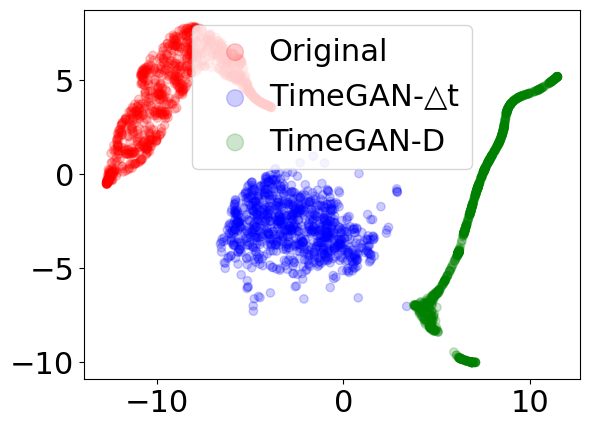}}\hfill
    {\includegraphics[width=0.27\columnwidth]{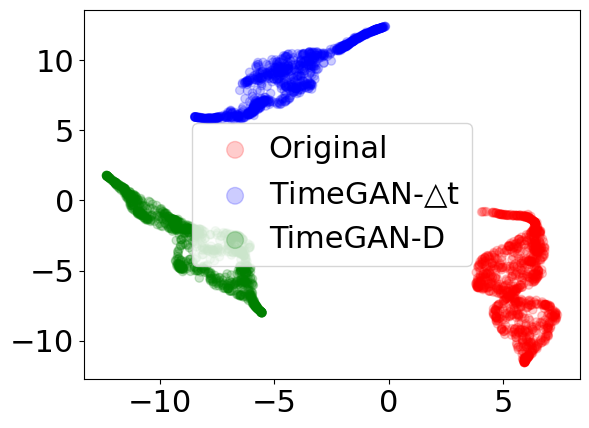}}\hfill
    {\includegraphics[width=0.27\columnwidth]{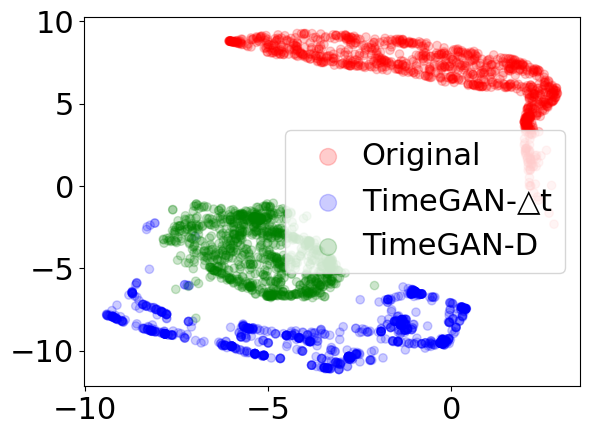}}\hfill
    \newline
    \centering
    {\includegraphics[width=0.27\columnwidth]{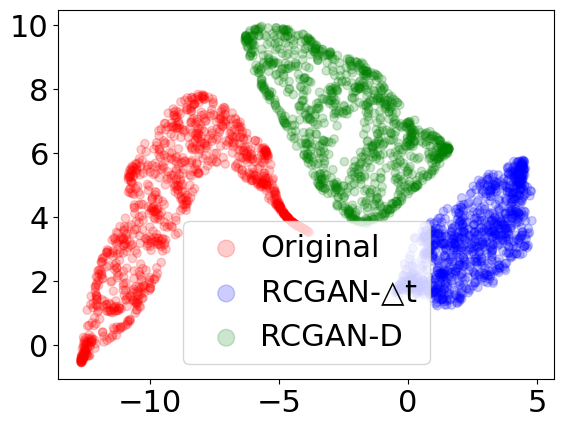}}\hfill
    {\includegraphics[width=0.27\columnwidth]{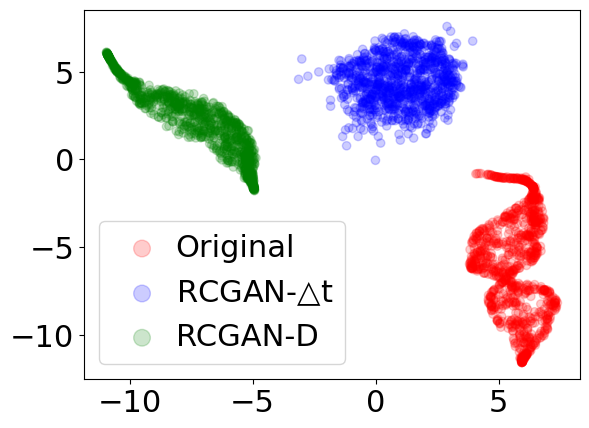}}\hfill
    {\includegraphics[width=0.27\columnwidth]{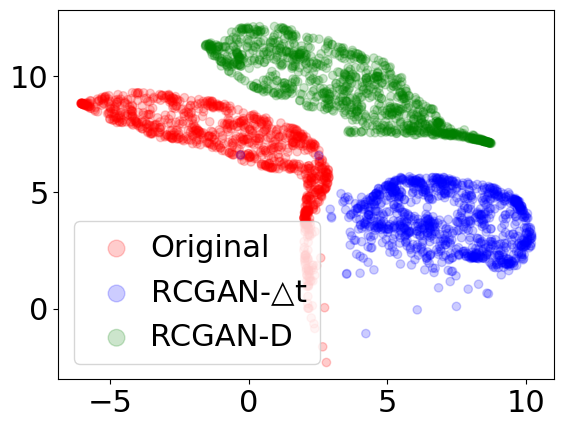}}\hfill
    \newline
    \centering
    {\includegraphics[width=0.27\columnwidth]{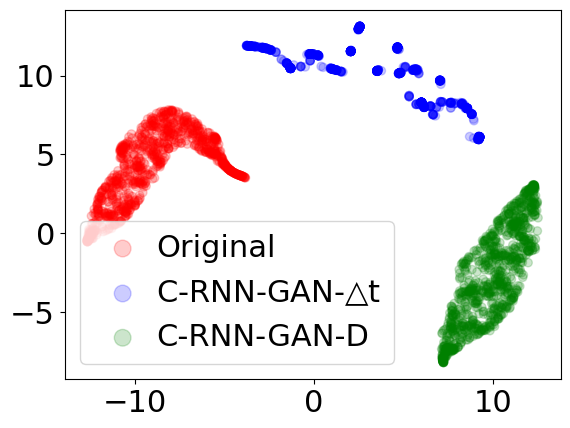}}\hfill
    {\includegraphics[width=0.27\columnwidth]{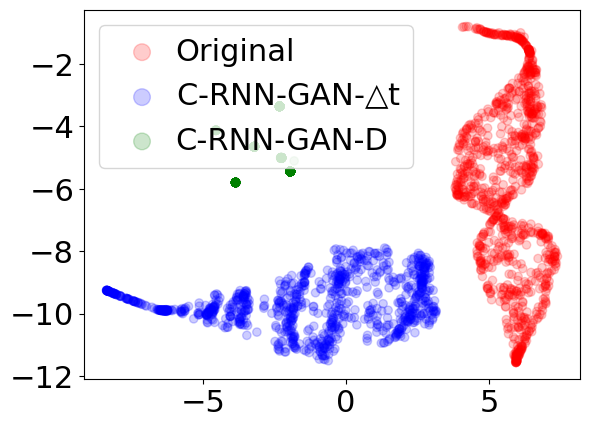}}\hfill
    {\includegraphics[width=0.27\columnwidth]{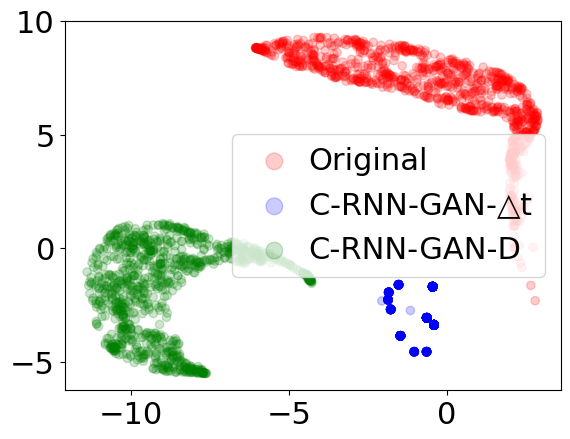}}\hfill
    \newline
    \centering
    {\includegraphics[width=0.27\columnwidth]{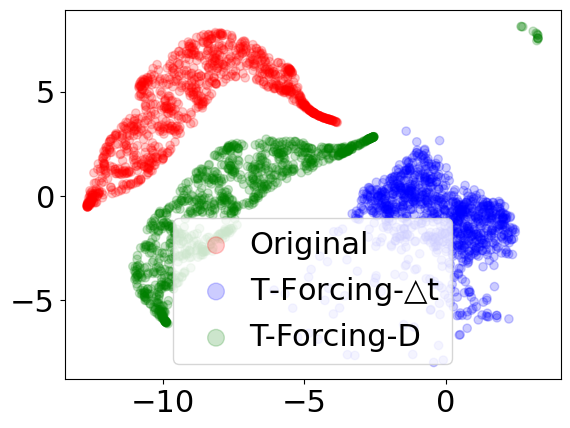}}\hfill
    {\includegraphics[width=0.27\columnwidth]{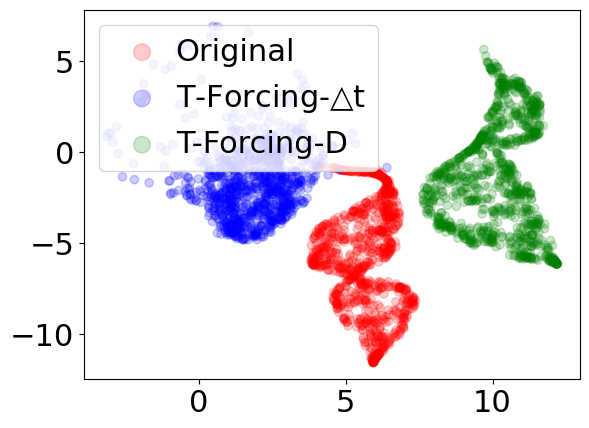}}\hfill
    {\includegraphics[width=0.27\columnwidth]{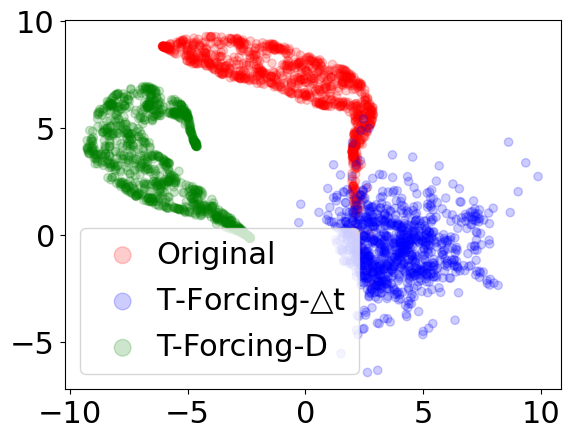}}\hfill
    \newline
    \subfigure[30\%]{\centering\includegraphics[width=0.27\columnwidth]{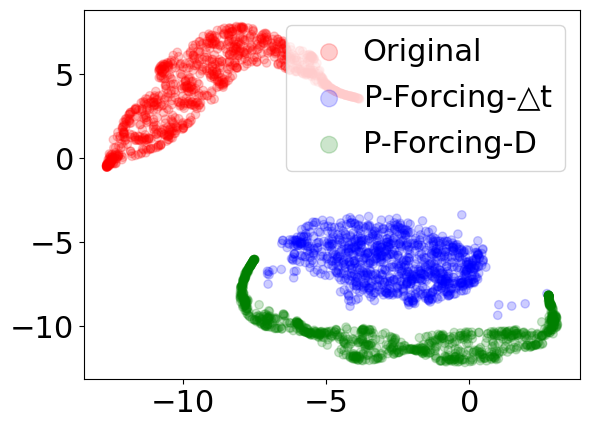}}\hfill
    \subfigure[50\%]{\centering\includegraphics[width=0.27\columnwidth]{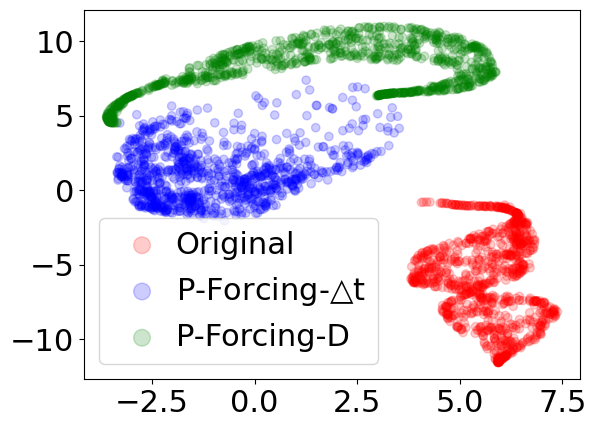}}\hfill
    \subfigure[70\%]{\includegraphics[width=0.27\columnwidth]{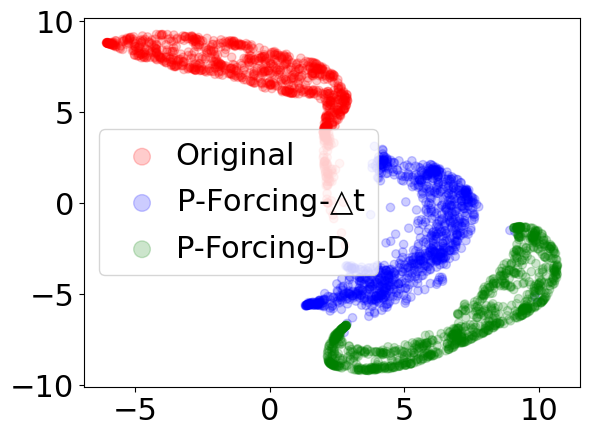}}\hfill
    \caption{t-SNE visualization of recovered irregular Sines data (the 1$^{\text{st}}$ column is for a dropping rate of 30\%, the 2$^{\text{nd}}$ column for a rate of 50\%, and the 3$^{\text{rd}}$ column for a rate of 70\%)}
    \label{fig:tsne_sines}
\end{figure}

\begin{figure}[ht]
    \centering
    {\includegraphics[width=0.27\columnwidth]{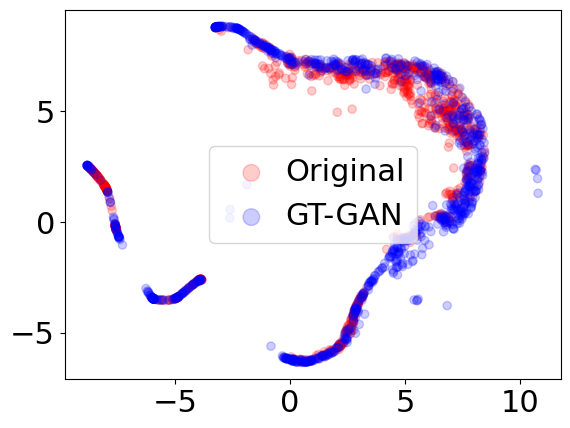}}\hfill
    {\includegraphics[width=0.27\columnwidth]{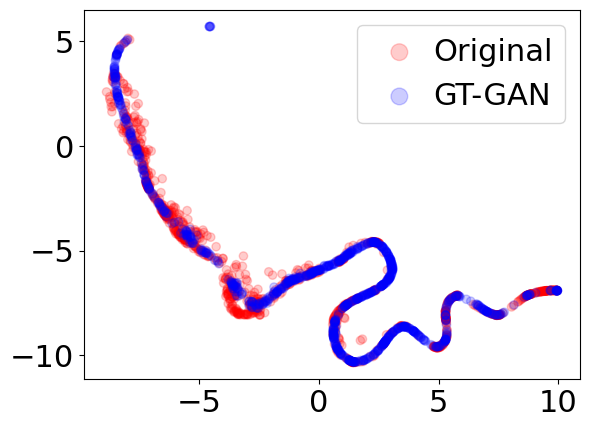}}\hfill
    {\includegraphics[width=0.27\columnwidth]{images/ori_IIT-GAN_stock_0.7.png}}\hfill
    \newline
    \centering
    {\includegraphics[width=0.27\columnwidth]{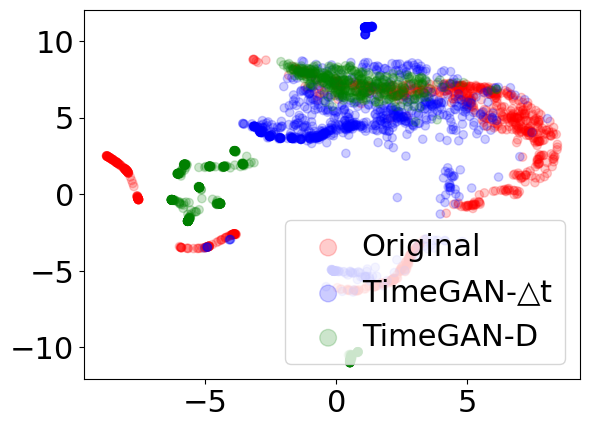}}\hfill
    {\includegraphics[width=0.27\columnwidth]{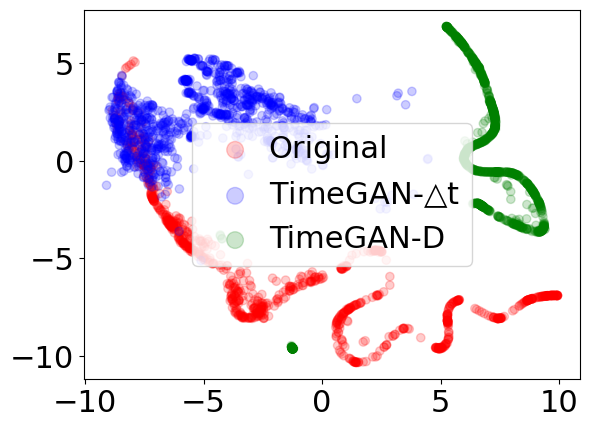}}\hfill
    {\includegraphics[width=0.27\columnwidth]{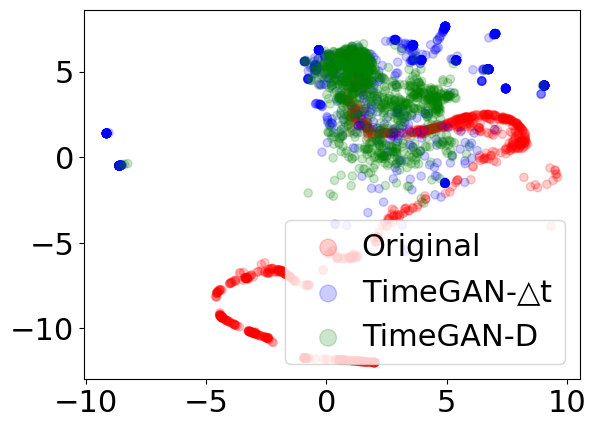}}\hfill
    \newline
    \centering
    {\includegraphics[width=0.27\columnwidth]{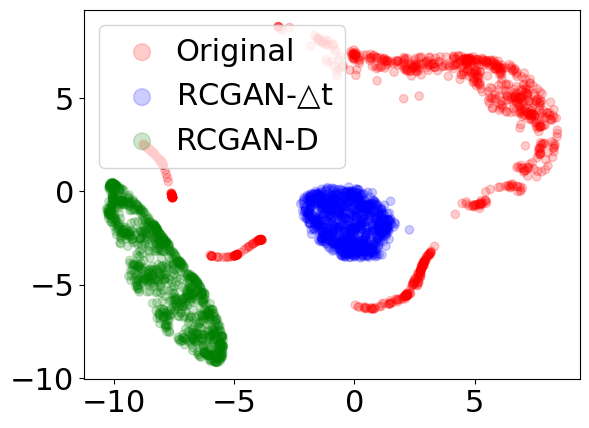}}\hfill
    {\includegraphics[width=0.27\columnwidth]{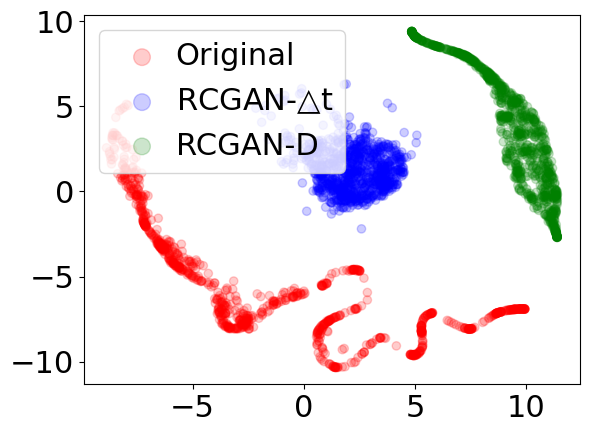}}\hfill
    {\includegraphics[width=0.27\columnwidth]{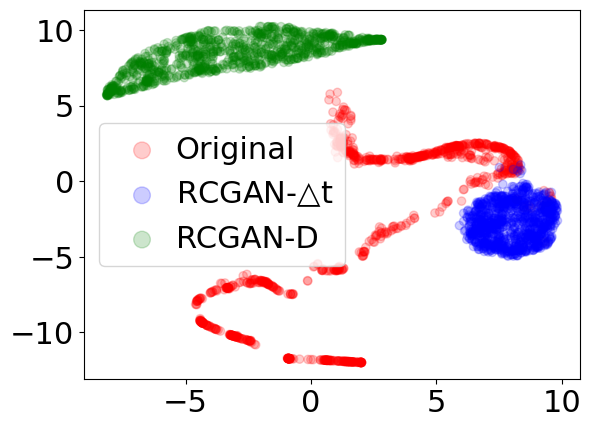}}\hfill
    \newline
    \centering
    {\includegraphics[width=0.27\columnwidth]{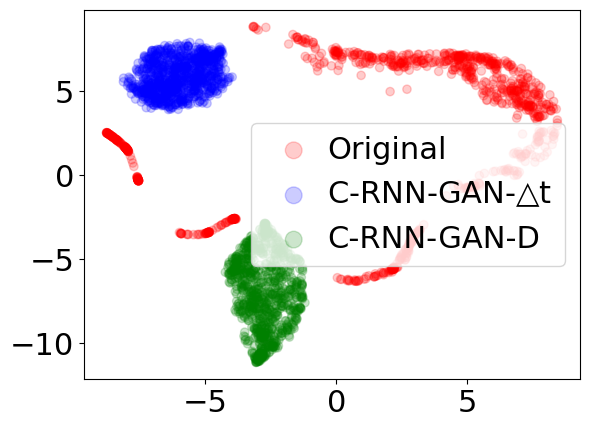}}\hfill
    {\includegraphics[width=0.27\columnwidth]{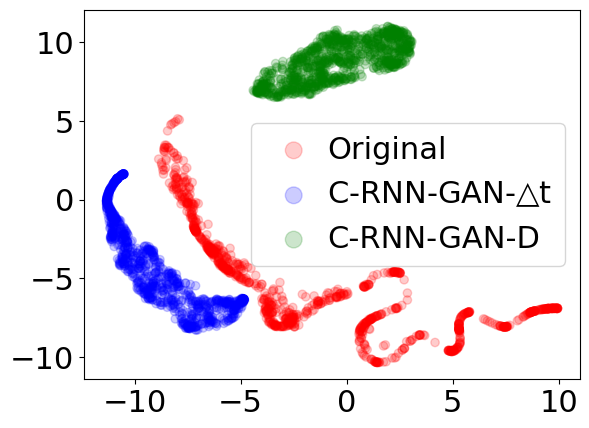}}\hfill
    {\includegraphics[width=0.27\columnwidth]{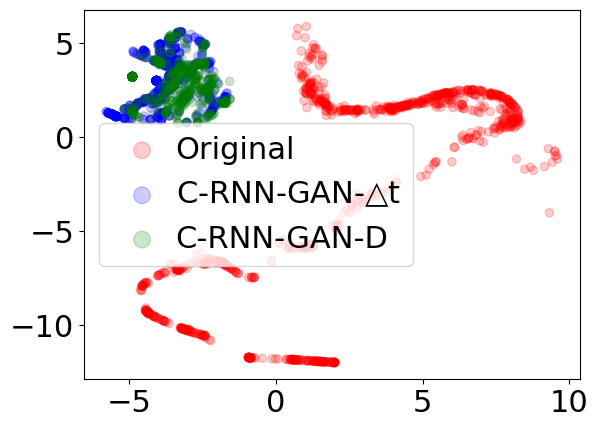}}\hfill
    \newline
    \centering
    {\includegraphics[width=0.27\columnwidth]{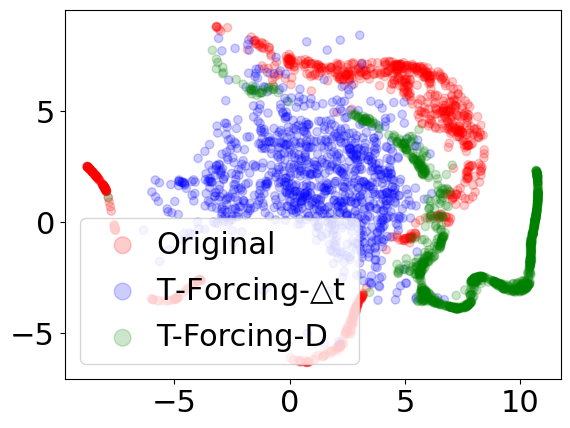}}\hfill
    {\includegraphics[width=0.27\columnwidth]{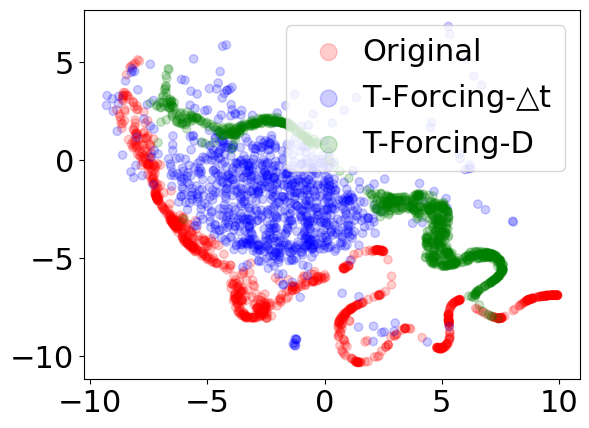}}\hfill
    {\includegraphics[width=0.27\columnwidth]{images/ori_T-Forcing-T_model_stock_0.7.png}}\hfill
    \newline
    \subfigure[30\%]{\centering\includegraphics[width=0.27\columnwidth]{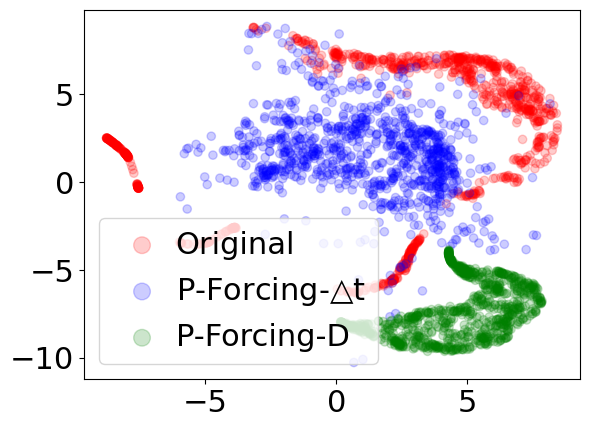}}\hfill
    \subfigure[50\%]{\centering\includegraphics[width=0.27\columnwidth]{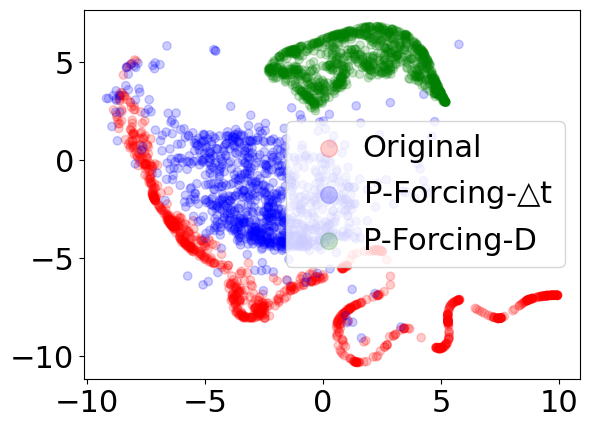}}\hfill
    \subfigure[70\%]{\centering\includegraphics[width=0.27\columnwidth]{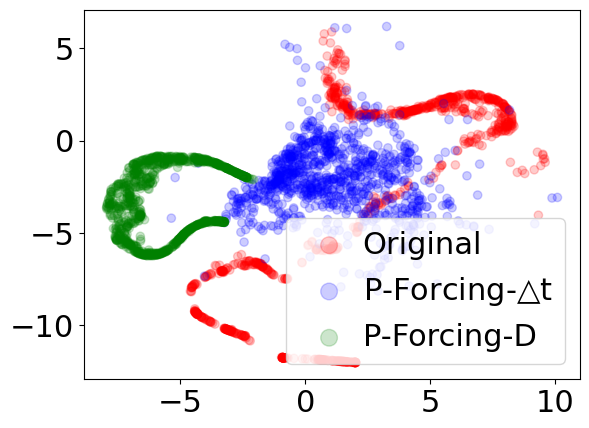}}\hfill
    \caption{t-SNE visualization of recovered irregular Stocks data (the 1$^{\text{st}}$ column is for a dropping rate of 30\%, the 2$^{\text{nd}}$ column for a rate of 50\%, and the 3$^{\text{rd}}$ column for a rate of 70\%)}
    \label{fig:tsne_stocks}
\end{figure}

\begin{figure}[ht]
    \centering
    {\includegraphics[width=0.27\columnwidth]{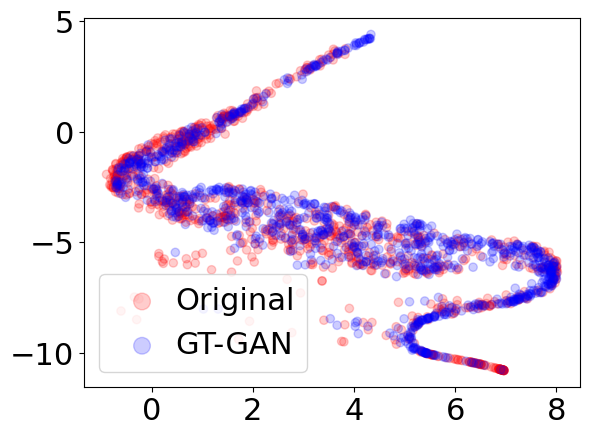}}\hfill
    {\includegraphics[width=0.27\columnwidth]{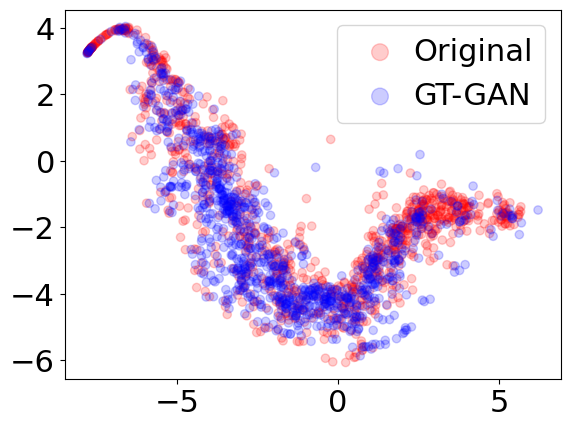}}\hfill
    {\includegraphics[width=0.27\columnwidth]{images/ori_IIT-GAN_energy_0.7.png}}\hfill
    \newline
    \centering
    {\includegraphics[width=0.27\columnwidth]{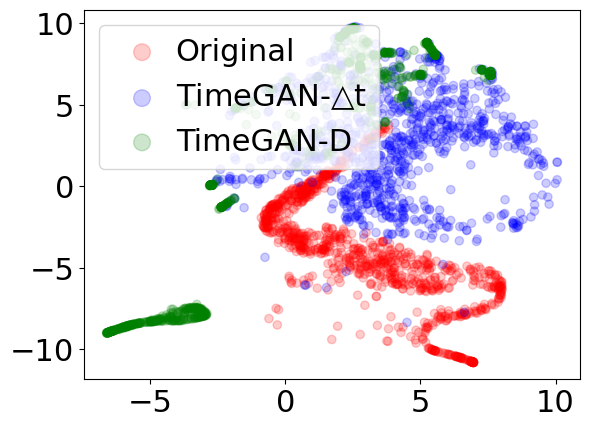}}\hfill
    {\includegraphics[width=0.27\columnwidth]{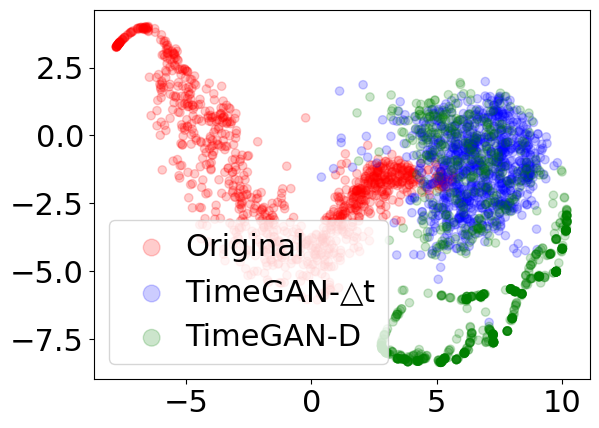}}\hfill
    {\includegraphics[width=0.27\columnwidth]{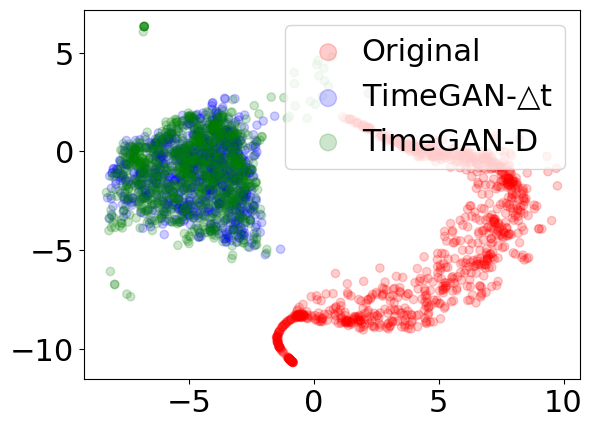}}\hfill
    \newline
    \centering
    {\includegraphics[width=0.27\columnwidth]{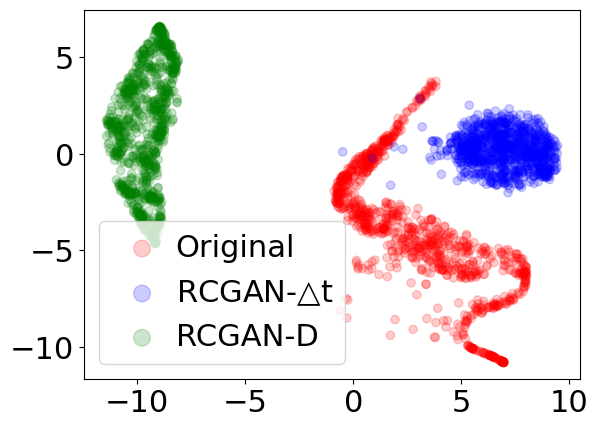}}\hfill
    {\includegraphics[width=0.27\columnwidth]{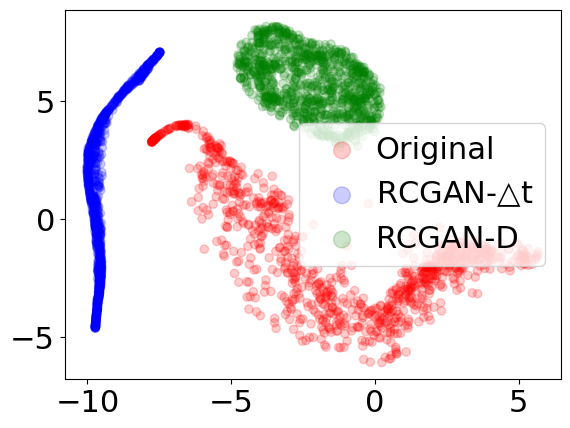}}\hfill
    {\includegraphics[width=0.27\columnwidth]{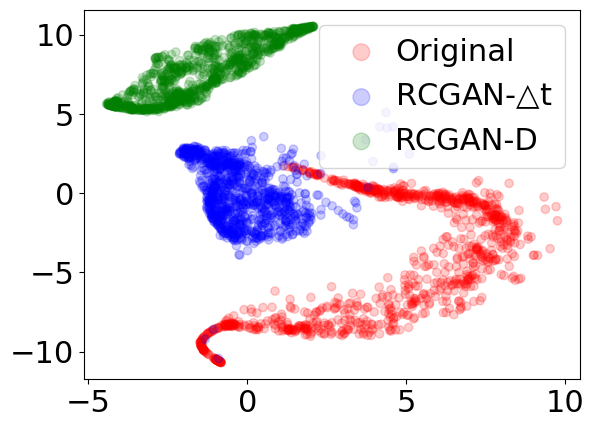}}\hfill
    \newline
    \centering
    {\includegraphics[width=0.27\columnwidth]{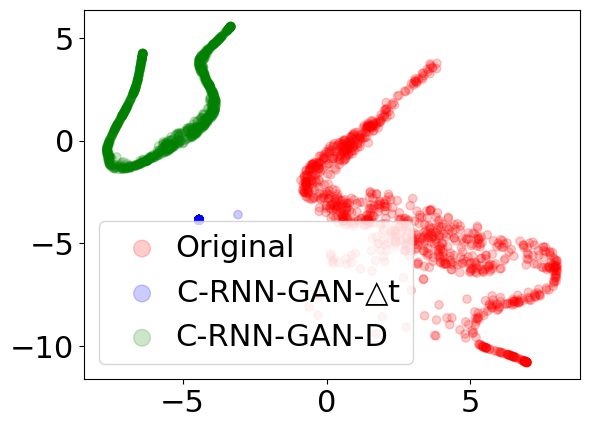}}\hfill
    {\includegraphics[width=0.27\columnwidth]{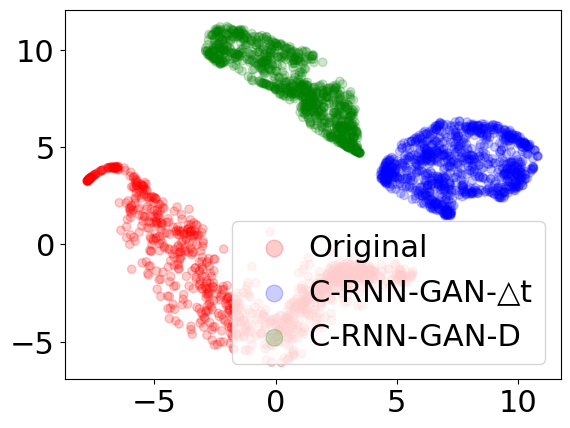}}\hfill
    {\includegraphics[width=0.27\columnwidth]{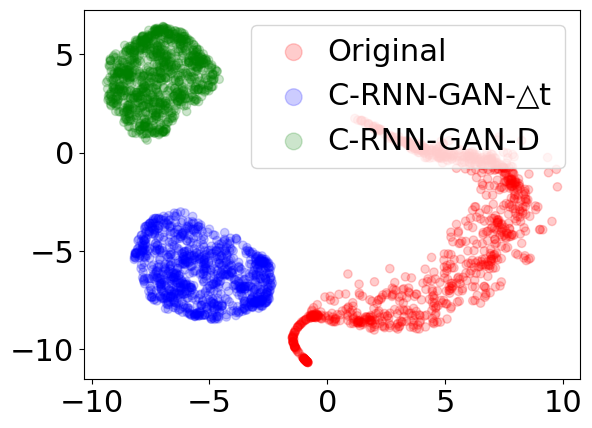}}\hfill
    \newline
    \centering
    {\includegraphics[width=0.27\columnwidth]{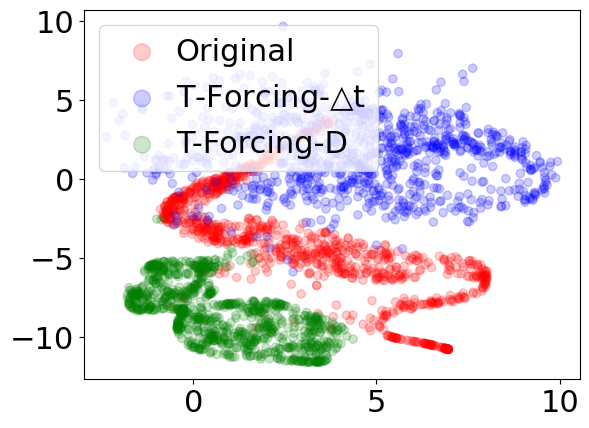}}\hfill
    {\includegraphics[width=0.27\columnwidth]{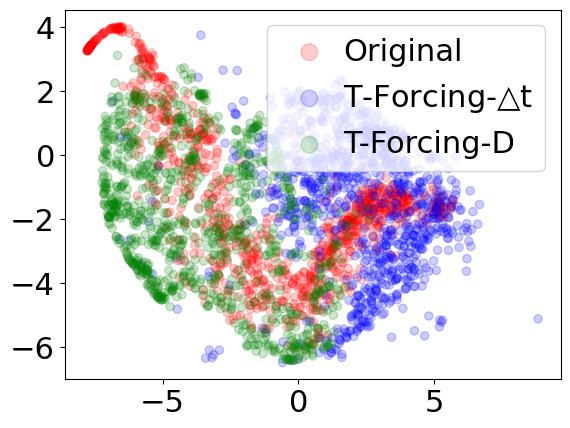}}\hfill
    {\includegraphics[width=0.27\columnwidth]{images/ori_T-Forcing-T_model_energy_0.7.png}}\hfill
    \newline
    \subfigure[30\%]{\centering\includegraphics[width=0.27\columnwidth]{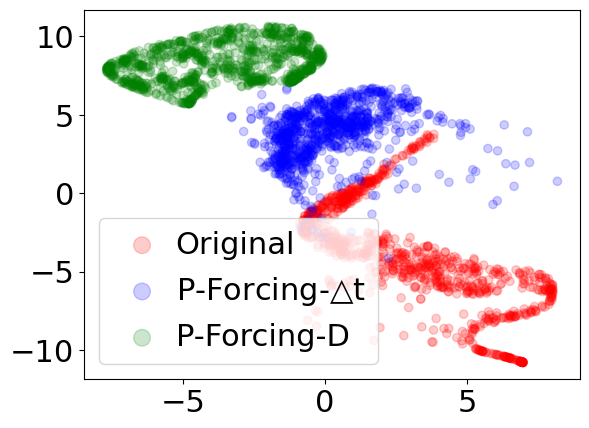}}\hfill
    \subfigure[50\%]{\centering\includegraphics[width=0.27\columnwidth]{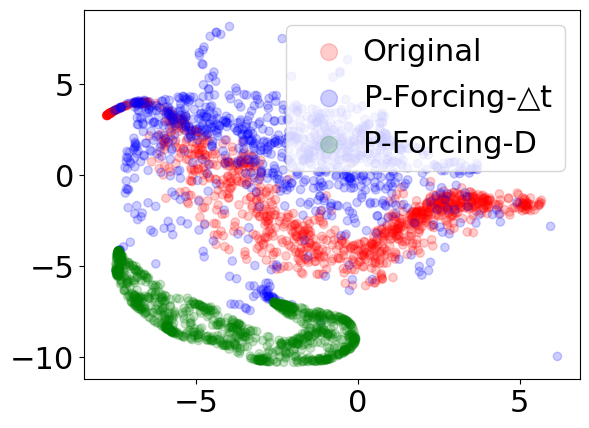}}\hfill
    \subfigure[70\%]{\centering\includegraphics[width=0.27\columnwidth]{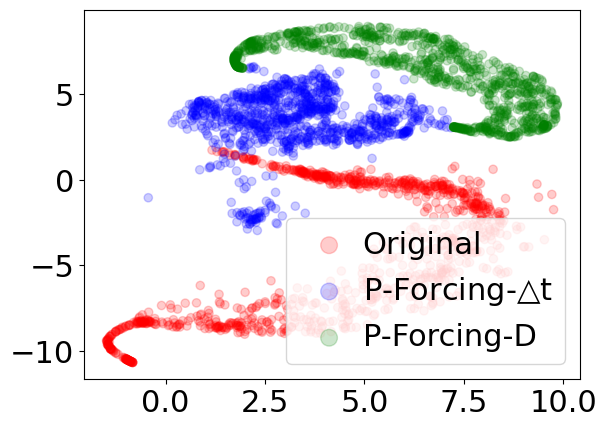}}\hfill
    \caption{t-SNE visualization of recovered irregular Energy data (the 1$^{\text{st}}$ column is for a dropping rate of 30\%, the 2$^{\text{nd}}$ column for a rate of 50\%, and the 3$^{\text{rd}}$ column for a rate of 70\%)}
    \label{fig:tsne_energy}
\end{figure}

\begin{figure}[ht]
    \centering
    {\includegraphics[width=0.27\columnwidth]{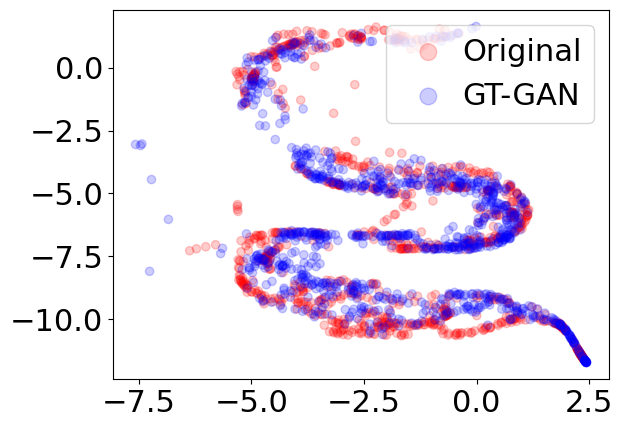}}\hfill
    {\includegraphics[width=0.27\columnwidth]{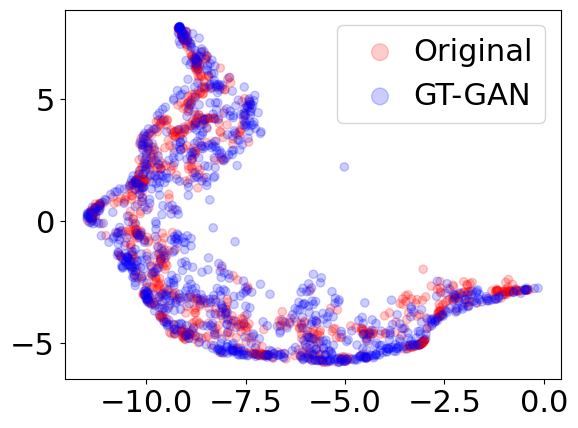}}\hfill
    {\includegraphics[width=0.27\columnwidth]{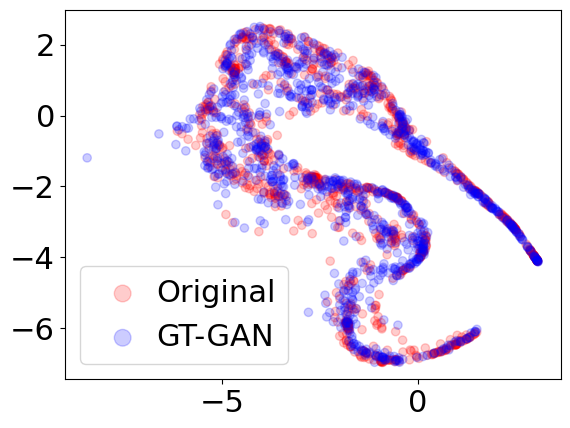}}\hfill
    \newline
    \centering
    {\includegraphics[width=0.27\columnwidth]{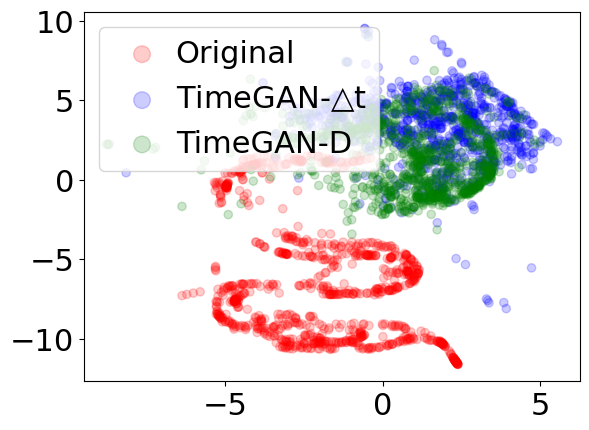}}\hfill
    {\includegraphics[width=0.27\columnwidth]{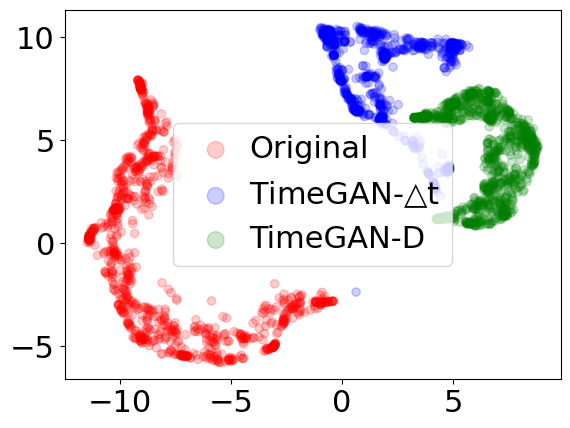}}\hfill
    {\includegraphics[width=0.27\columnwidth]{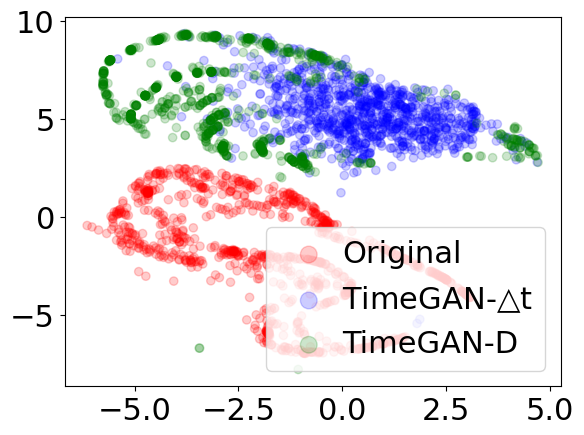}}\hfill
    \newline
    \centering
    {\includegraphics[width=0.27\columnwidth]{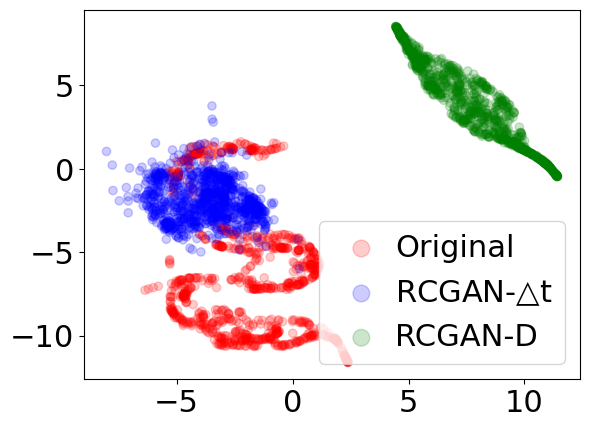}}\hfill
    {\includegraphics[width=0.27\columnwidth]{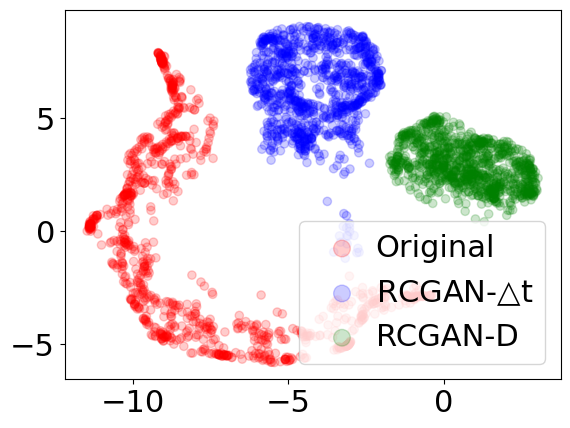}}\hfill
    {\includegraphics[width=0.27\columnwidth]{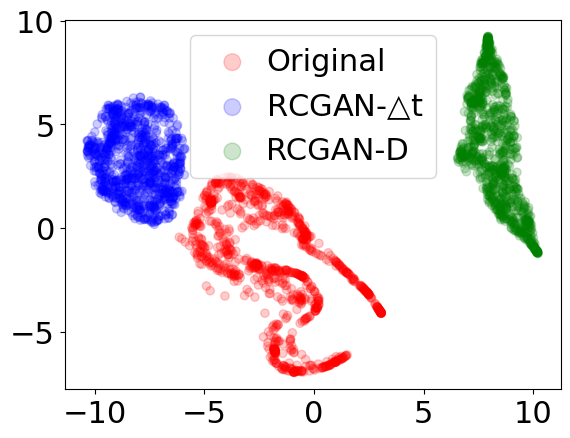}}\hfill
    \newline
    \centering
    {\includegraphics[width=0.27\columnwidth]{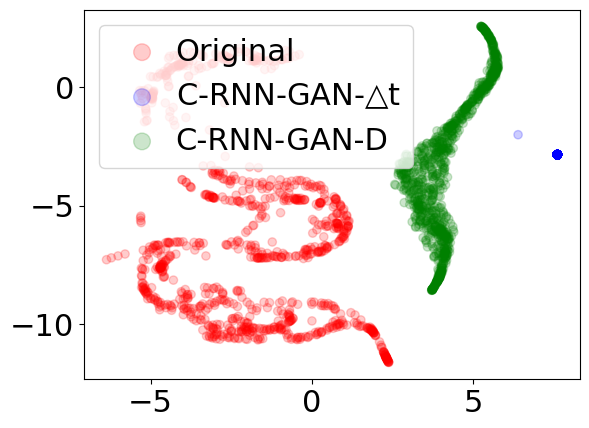}}\hfill
    {\includegraphics[width=0.27\columnwidth]{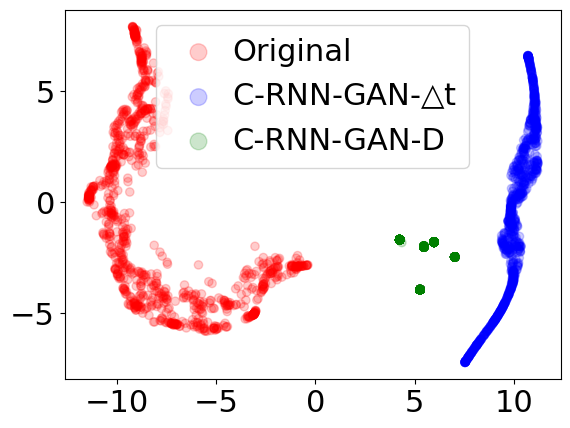}}\hfill
    {\includegraphics[width=0.27\columnwidth]{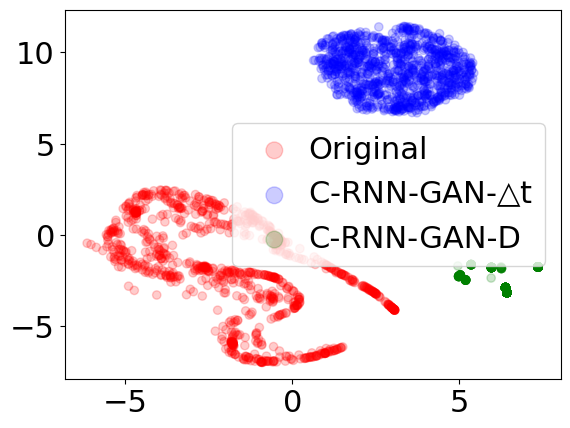}}\hfill
    \newline
    \centering
    {\includegraphics[width=0.27\columnwidth]{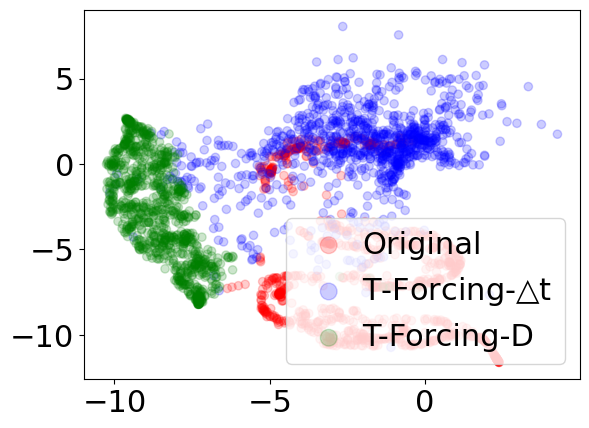}}\hfill
    {\includegraphics[width=0.27\columnwidth]{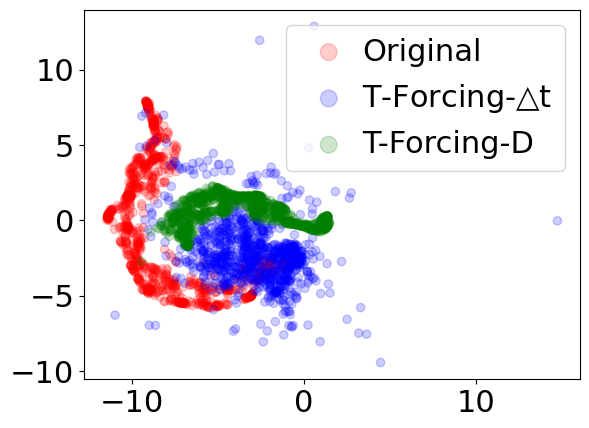}}\hfill
    {\includegraphics[width=0.27\columnwidth]{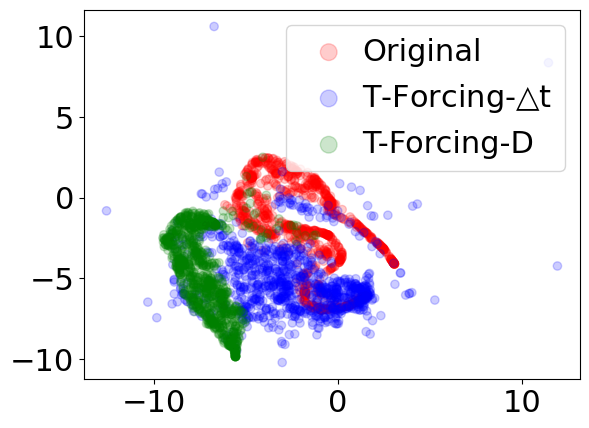}}\hfill
    \newline
    \subfigure[30\%]{\centering\includegraphics[width=0.27\columnwidth]{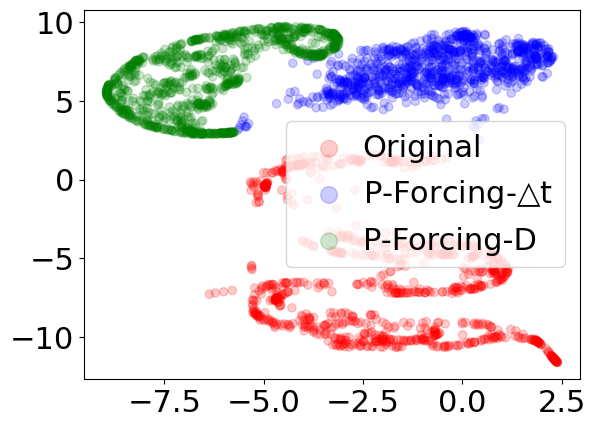}}\hfill
    \subfigure[50\%]{\centering\includegraphics[width=0.27\columnwidth]{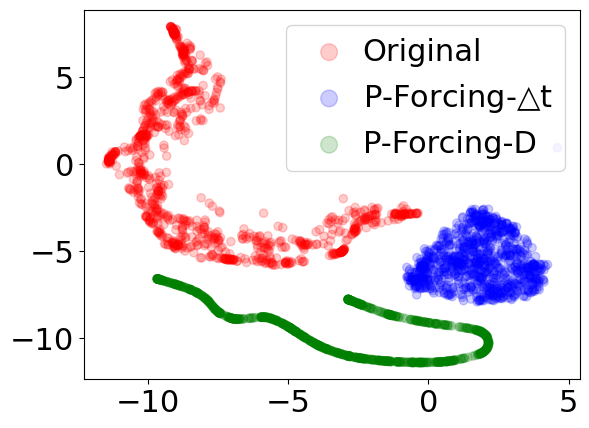}}\hfill
    \subfigure[70\%]{\centering\includegraphics[width=0.27\columnwidth]{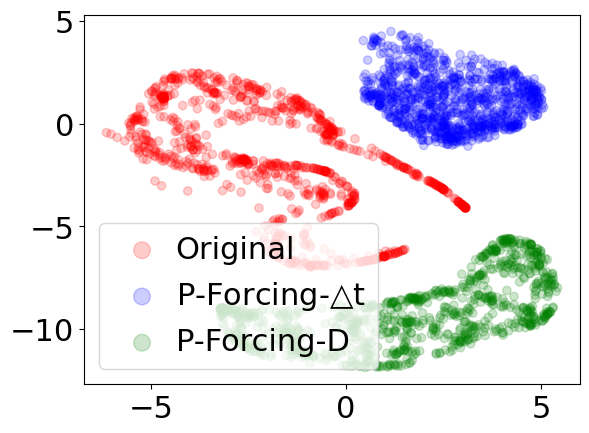}}\hfill
    \caption{t-SNE visualization of recovered irregular MuJoCo data (the 1$^{\text{st}}$ column is for a dropping rate of 30\%, the 2$^{\text{nd}}$ column for a rate of 50\%, and the 3$^{\text{rd}}$ column for a rate of 70\%)}
    \label{fig:tsne_mujoco}
\end{figure}

\begin{figure}[ht]
    \centering
    {\includegraphics[width=0.25\columnwidth]{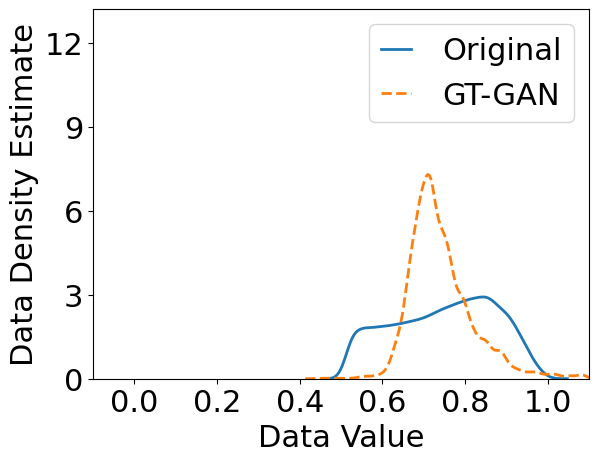}}\hfill
    {\includegraphics[width=0.25\columnwidth]{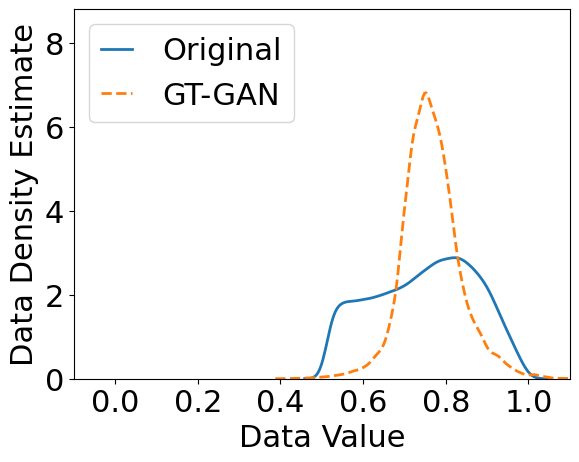}}\hfill
    {\includegraphics[width=0.25\columnwidth]{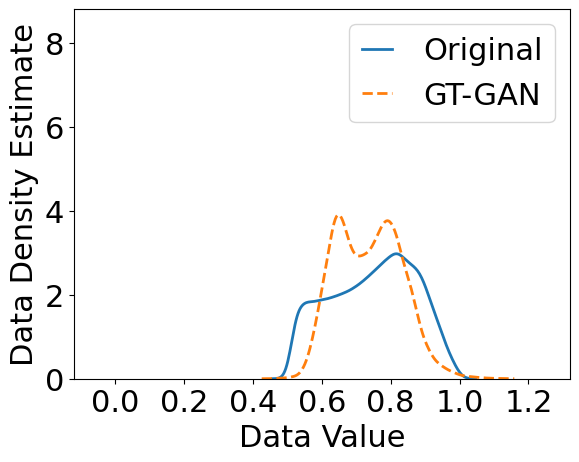}}\hfill
    \newline
    \centering
    {\includegraphics[width=0.25\columnwidth]{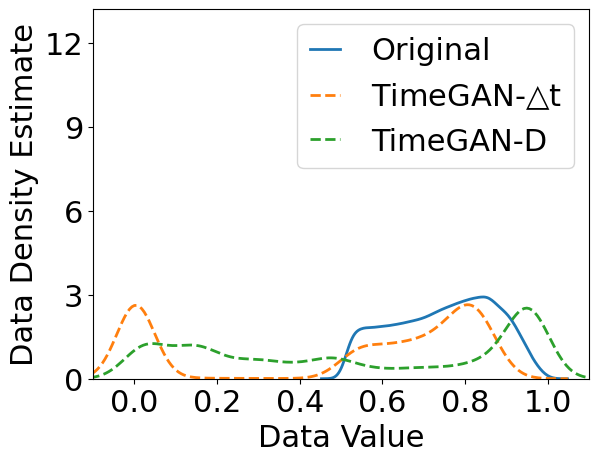}}\hfill
    {\includegraphics[width=0.25\columnwidth]{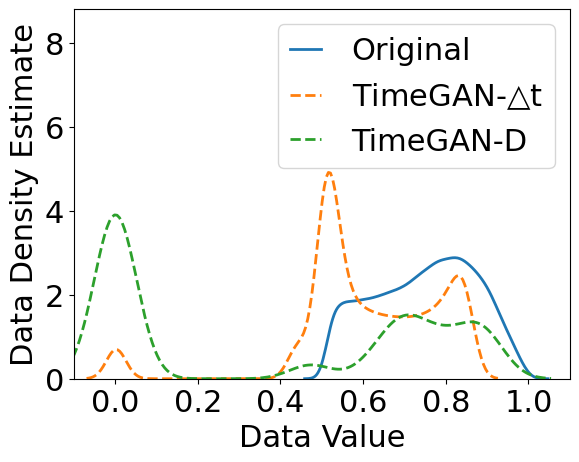}}\hfill
    {\includegraphics[width=0.25\columnwidth]{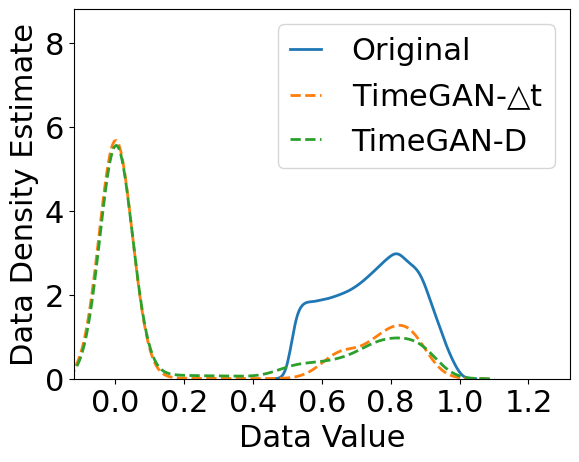}}\hfill
    \newline
    \centering
    {\includegraphics[width=0.25\columnwidth]{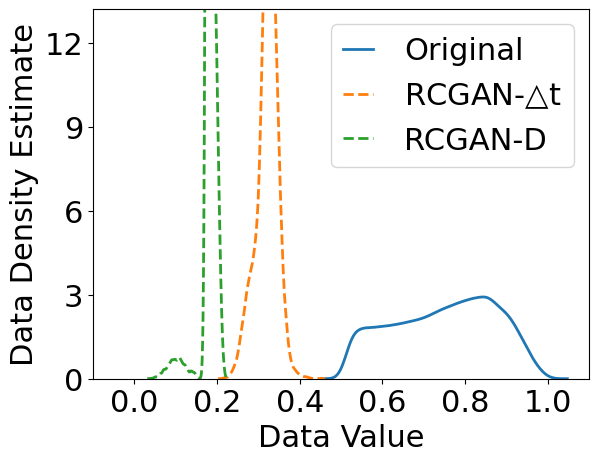}}\hfill
    {\includegraphics[width=0.25\columnwidth]{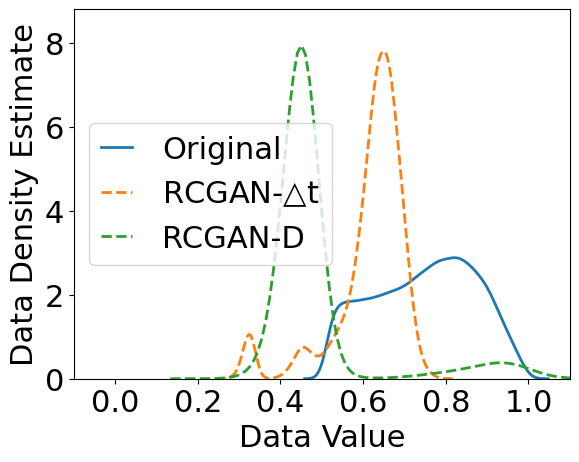}}\hfill
    {\includegraphics[width=0.25\columnwidth]{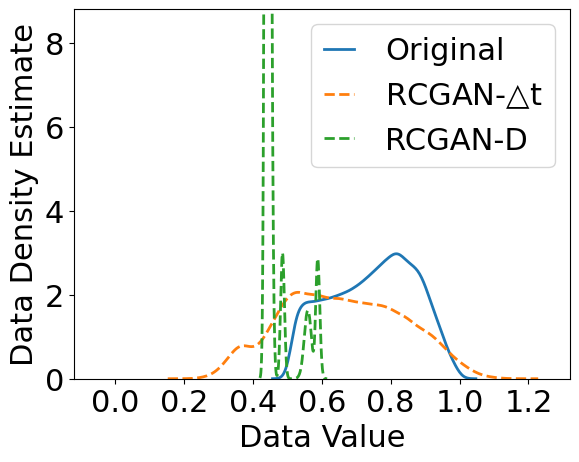}}\hfill
    \newline
    \centering
    {\includegraphics[width=0.25\columnwidth]{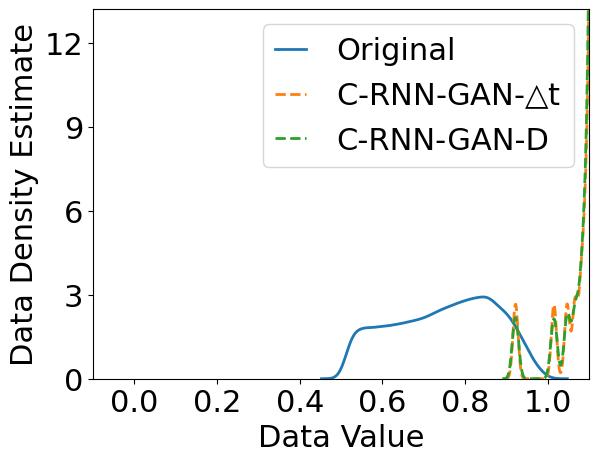}}\hfill
    {\includegraphics[width=0.25\columnwidth]{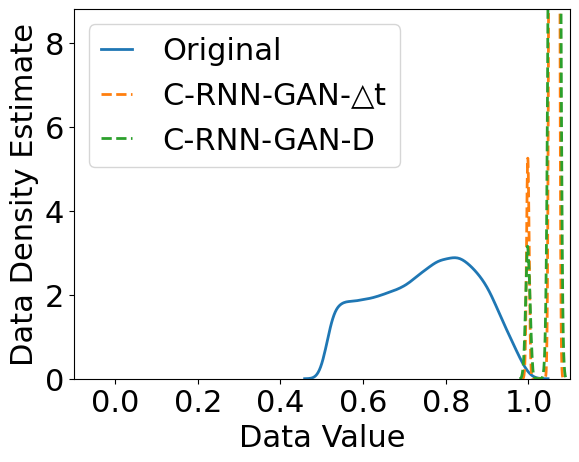}}\hfill
    {\includegraphics[width=0.25\columnwidth]{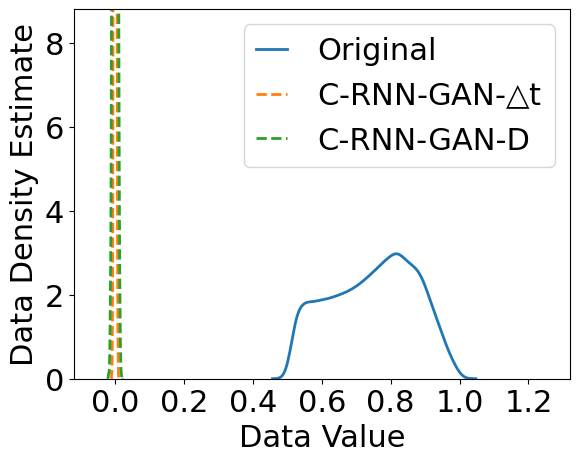}}\hfill
    \newline
    \centering
    {\includegraphics[width=0.25\columnwidth]{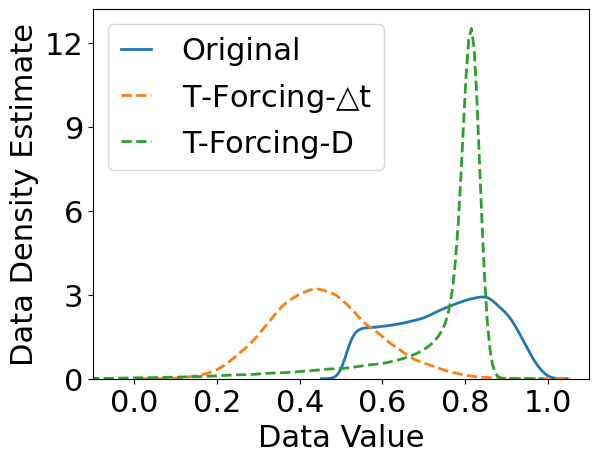}}\hfill
    {\includegraphics[width=0.25\columnwidth]{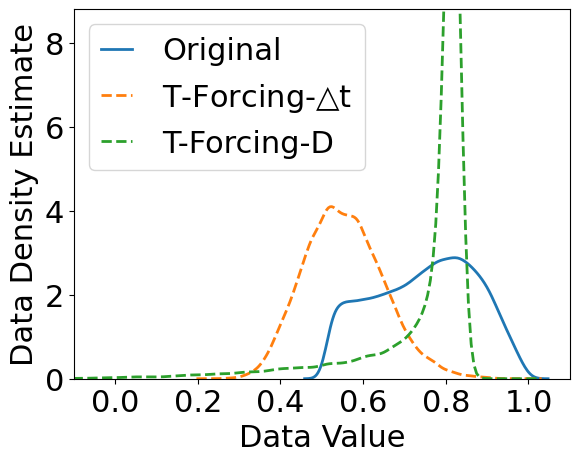}}\hfill
    {\includegraphics[width=0.25\columnwidth]{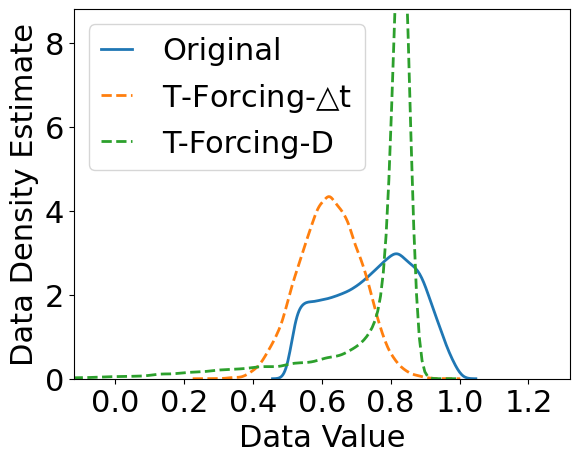}}\hfill
    \newline
    \subfigure[30\%]{\centering\includegraphics[width=0.25\columnwidth]{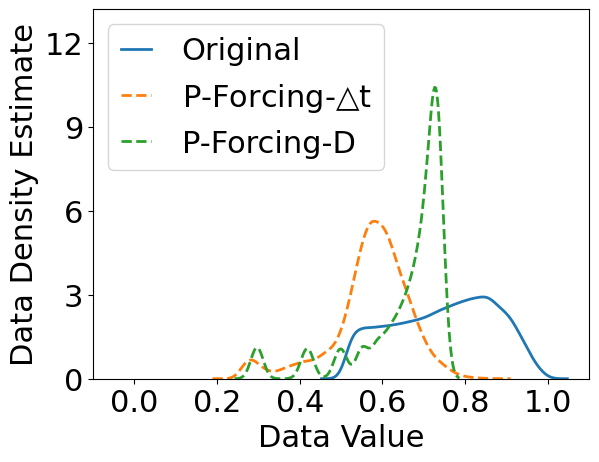}}\hfill
    \subfigure[50\%]{\centering\includegraphics[width=0.25\columnwidth]{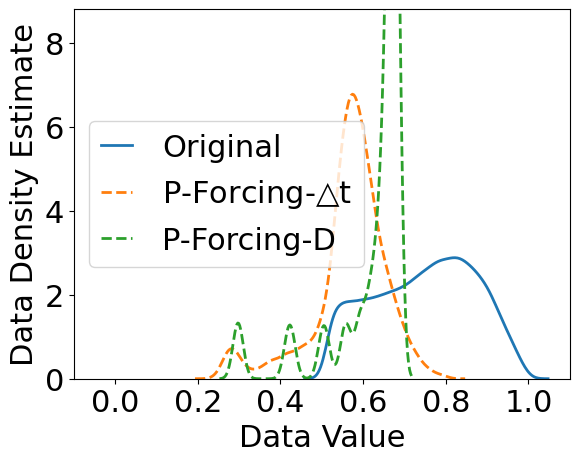}}\hfill
    \subfigure[70\%]{\centering\includegraphics[width=0.25\columnwidth]{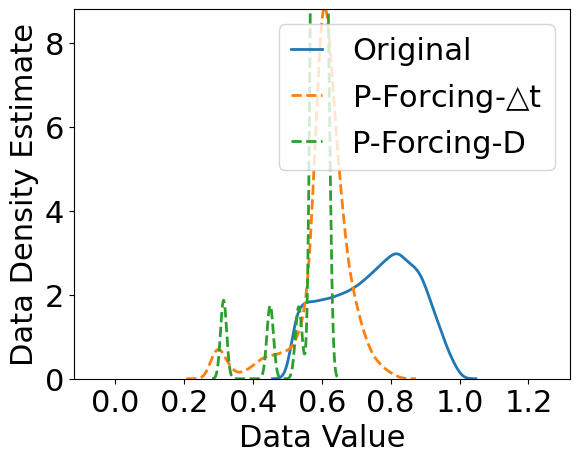}}\hfill
    \caption{Distributions of the Sines data (the 1$^{\text{st}}$ column is for a dropping rate of 30\%, the 2$^{\text{nd}}$ column for a rate of 50\%, and the 3$^{\text{rd}}$ column for a rate of 70\%)}
    \label{fig:histo_sines}
\end{figure}

\begin{figure}[ht]
    \centering
    {\includegraphics[width=0.25\columnwidth]{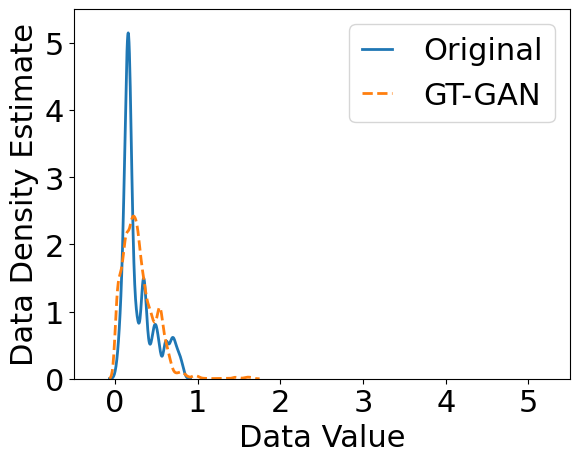}}\hfill
    {\includegraphics[width=0.25\columnwidth]{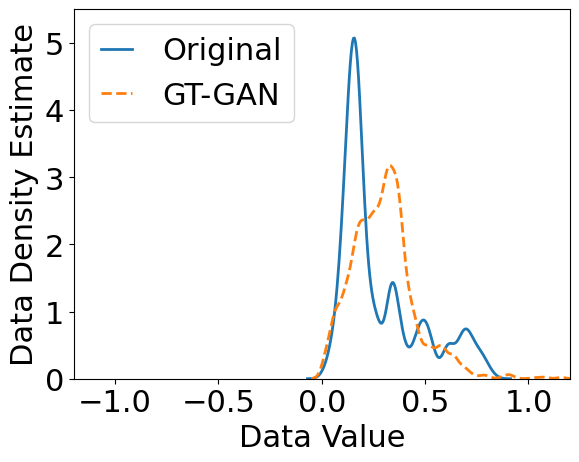}}\hfill
    {\includegraphics[width=0.25\columnwidth]{images/stock_0.7IIT-GAN_histo.png}}\hfill
    \newline
    \centering
    {\includegraphics[width=0.25\columnwidth]{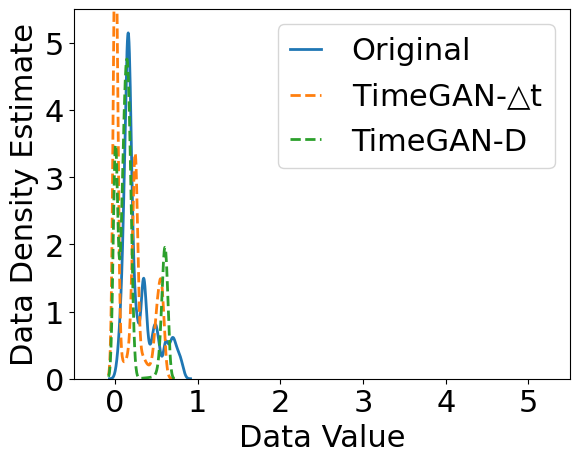}}\hfill
    {\includegraphics[width=0.25\columnwidth]{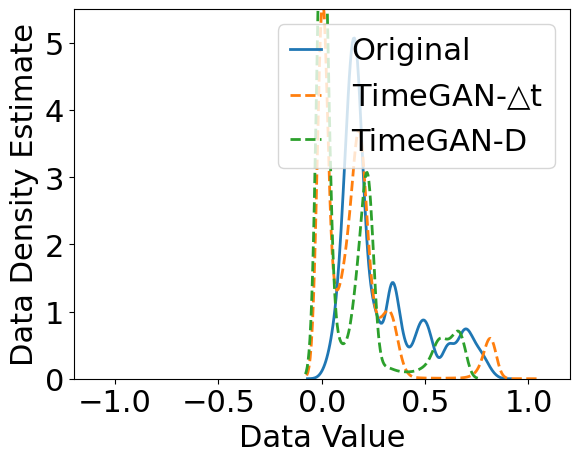}}\hfill
    {\includegraphics[width=0.25\columnwidth]{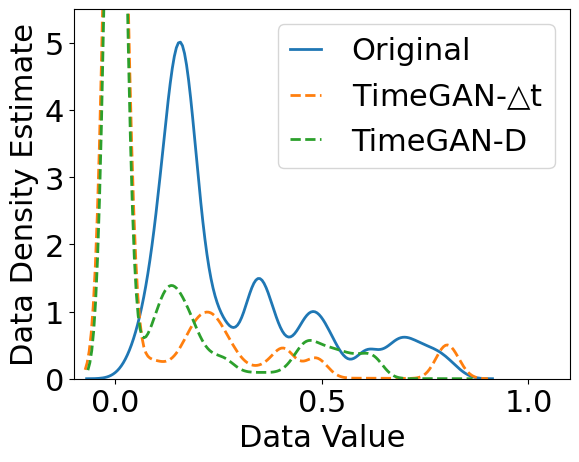}}\hfill
    \newline
    \centering
    {\includegraphics[width=0.25\columnwidth]{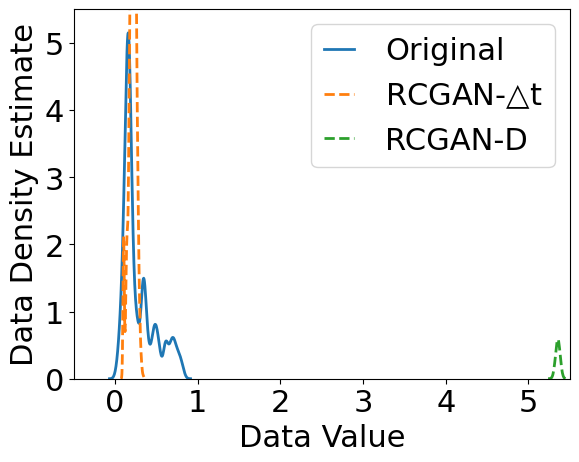}}\hfill
    {\includegraphics[width=0.25\columnwidth]{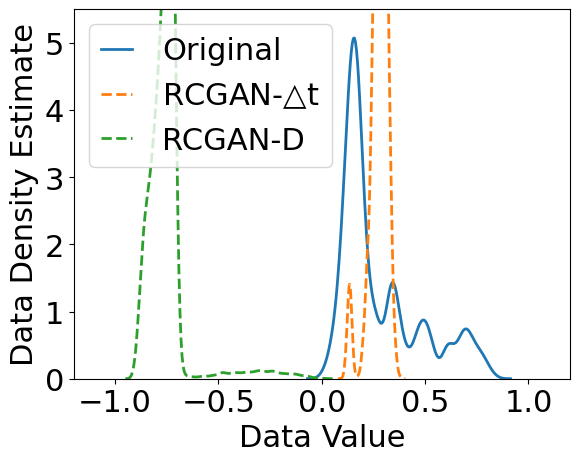}}\hfill
    {\includegraphics[width=0.25\columnwidth]{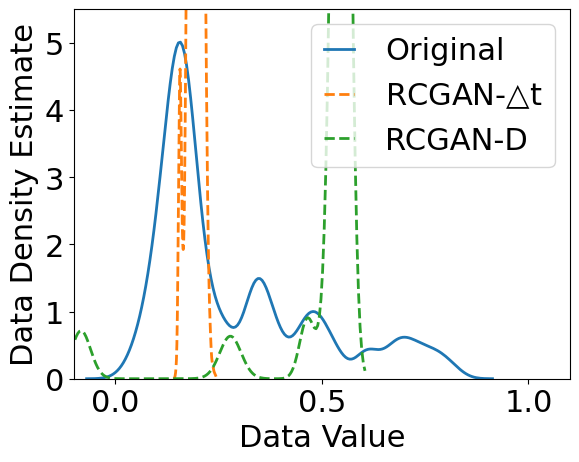}}\hfill
    \newline
    \centering
    {\includegraphics[width=0.25\columnwidth]{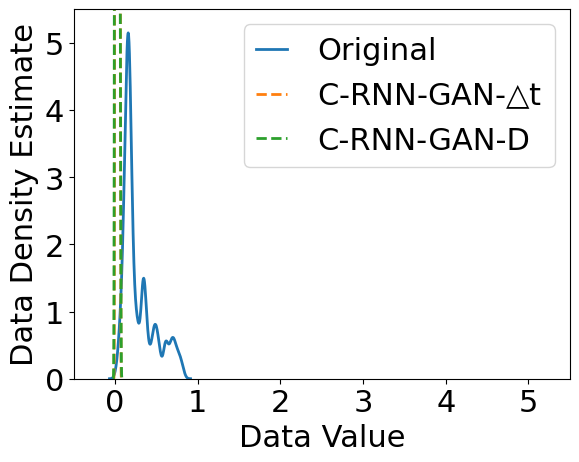}}\hfill
    {\includegraphics[width=0.25\columnwidth]{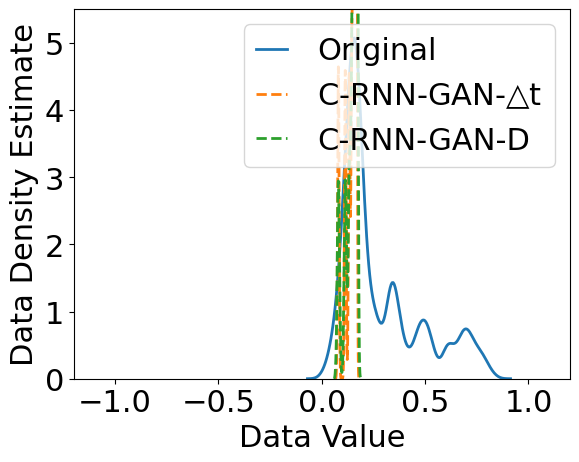}}\hfill
    {\includegraphics[width=0.25\columnwidth]{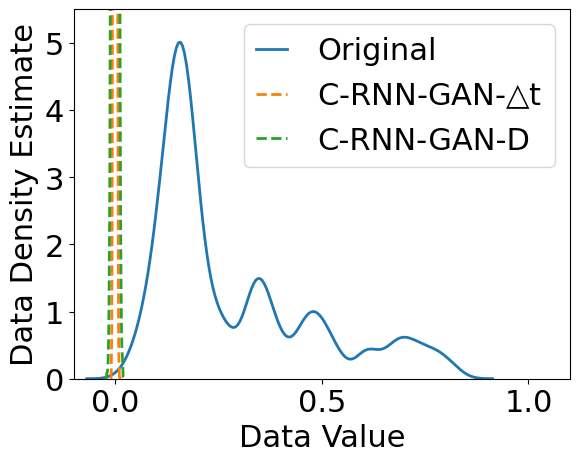}}\hfill
    \newline
    \centering
    {\includegraphics[width=0.25\columnwidth]{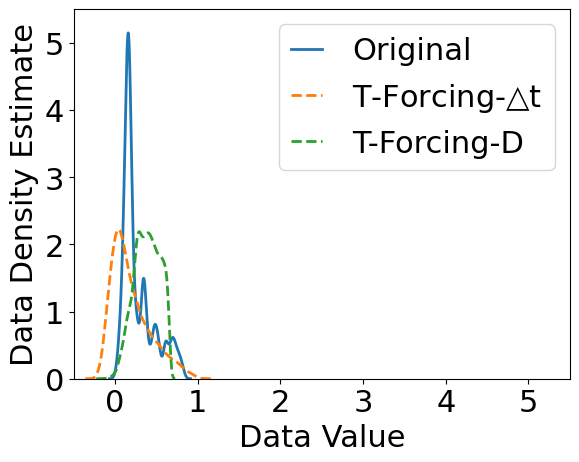}}\hfill
    {\includegraphics[width=0.25\columnwidth]{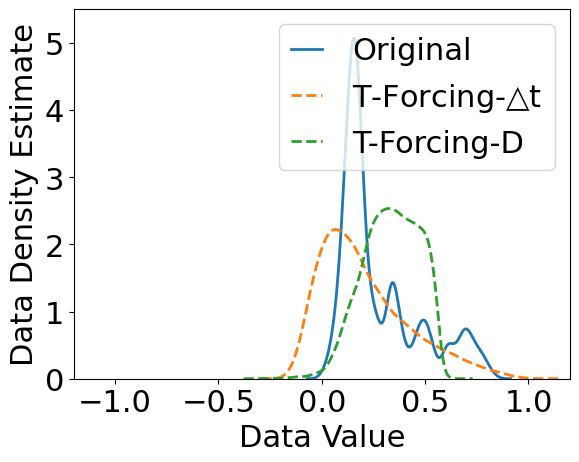}}\hfill
    {\includegraphics[width=0.25\columnwidth]{images/stock_0.7T-Forcing-D_model_histo.png}}\hfill
    \newline
    \subfigure[30\%]{\centering\includegraphics[width=0.25\columnwidth]{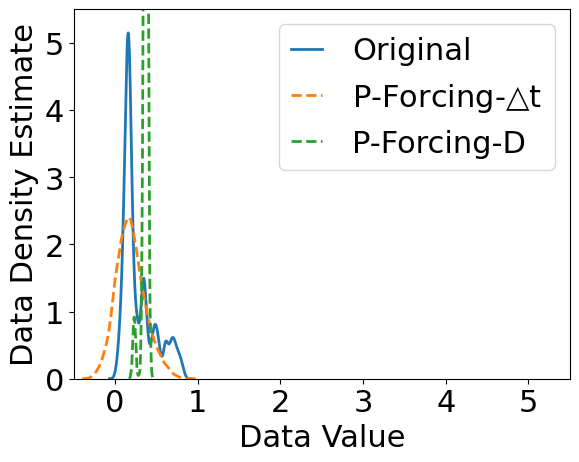}}\hfill
    \subfigure[50\%]{\centering\includegraphics[width=0.25\columnwidth]{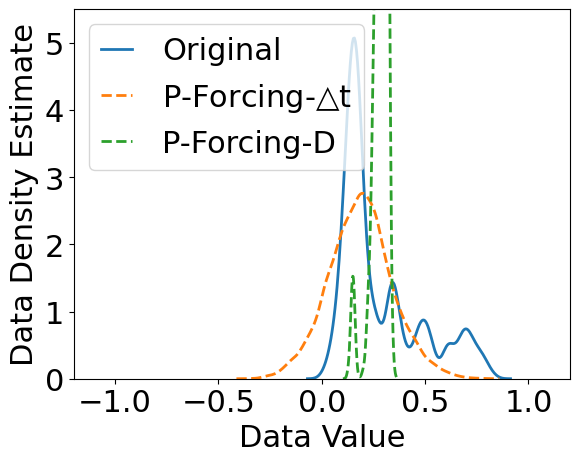}}\hfill
    \subfigure[70\%]{\centering\includegraphics[width=0.25\columnwidth]{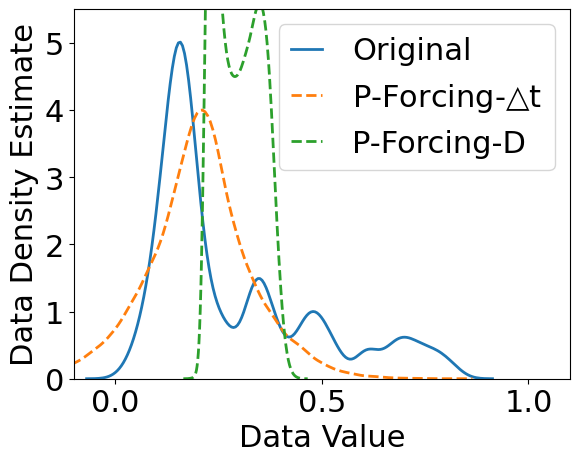}}\hfill
    \caption{Distributions of the Stocks data (the 1$^{\text{st}}$ column is for a dropping rate of 30\%, the 2$^{\text{nd}}$ column for a rate of 50\%, and the 3$^{\text{rd}}$ column for a rate of 70\%)}
    \label{fig:histo_stocks}
\end{figure}

\begin{figure}[ht]
    \centering
    {\includegraphics[width=0.25\columnwidth]{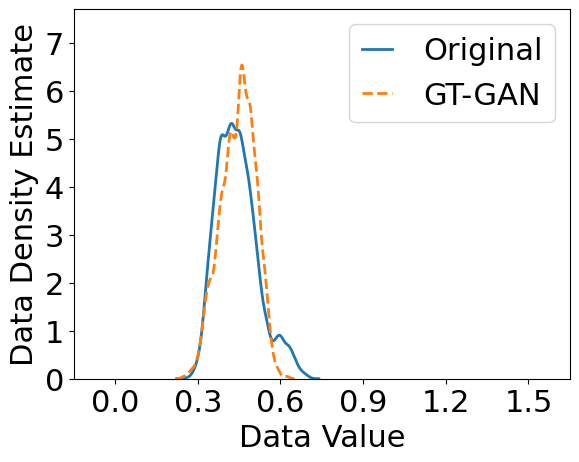}}\hfill
    {\includegraphics[width=0.25\columnwidth]{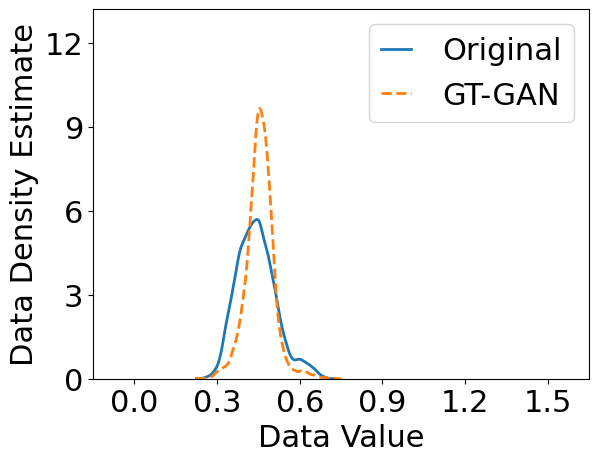}}\hfill
    {\includegraphics[width=0.25\columnwidth]{images/energy_0.7IIT-GAN_histo.png}}\hfill
    \newline
    \centering
    {\includegraphics[width=0.25\columnwidth]{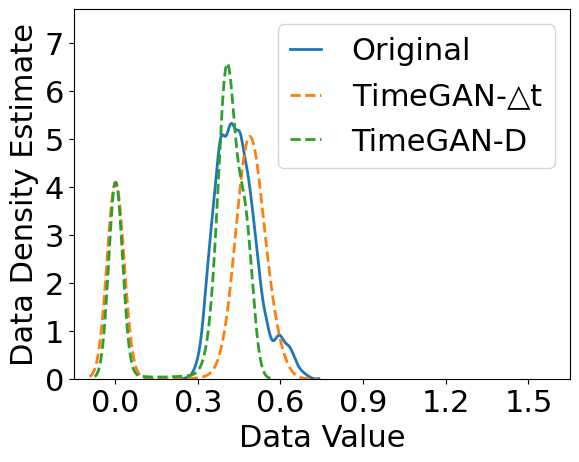}}\hfill
    {\includegraphics[width=0.25\columnwidth]{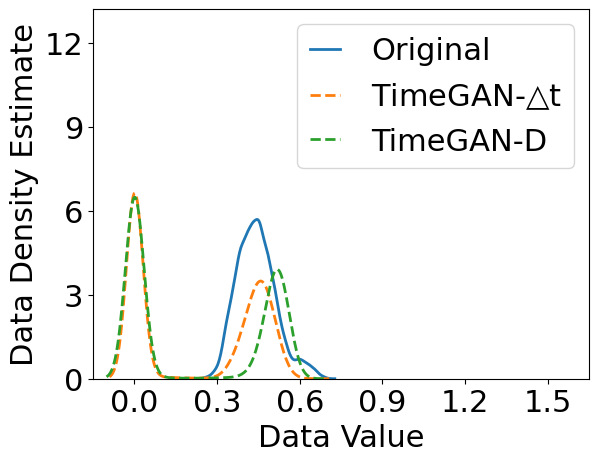}}\hfill
    {\includegraphics[width=0.25\columnwidth]{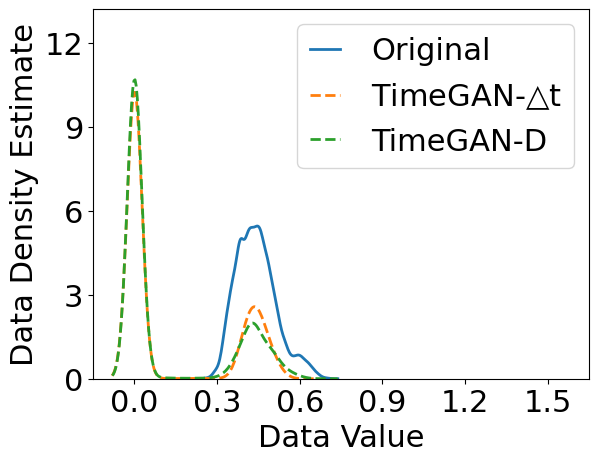}}\hfill
    \newline
    \centering
    {\includegraphics[width=0.25\columnwidth]{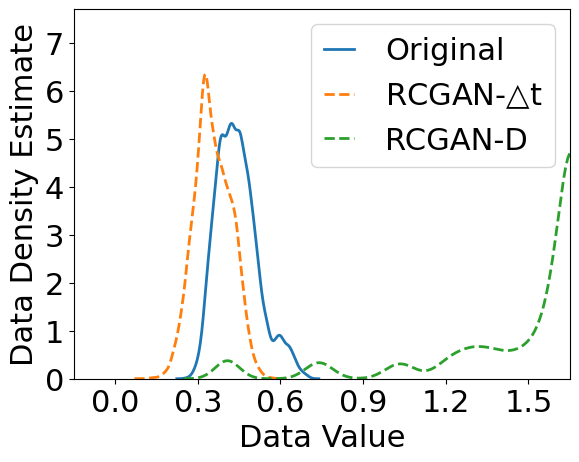}}\hfill
    {\includegraphics[width=0.25\columnwidth]{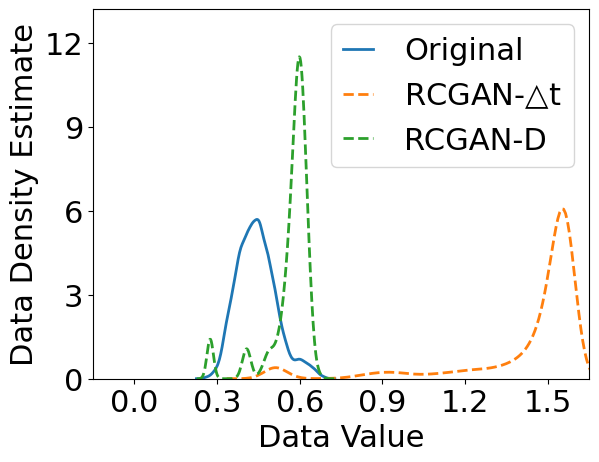}}\hfill
    {\includegraphics[width=0.25\columnwidth]{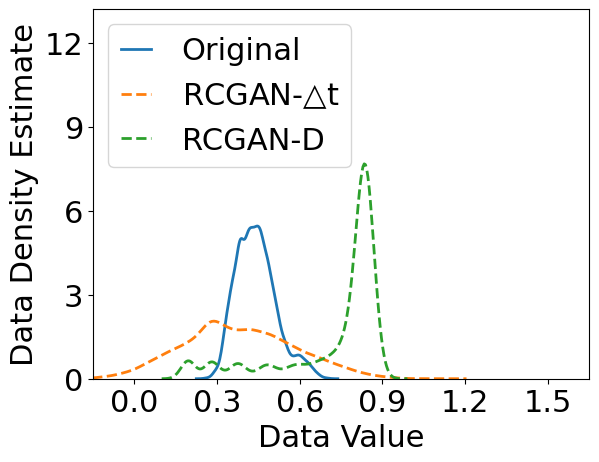}}\hfill
    \newline
    \centering
    {\includegraphics[width=0.25\columnwidth]{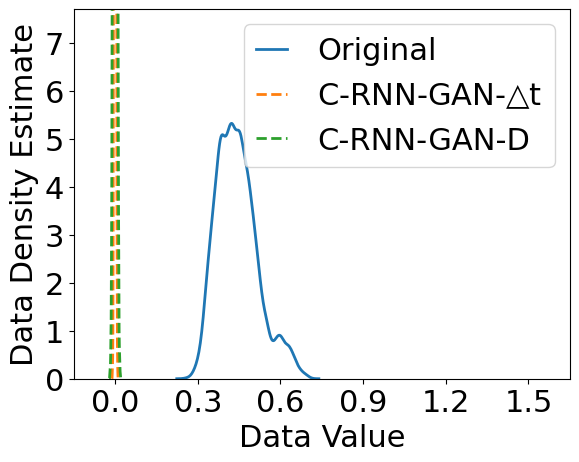}}\hfill
    {\includegraphics[width=0.25\columnwidth]{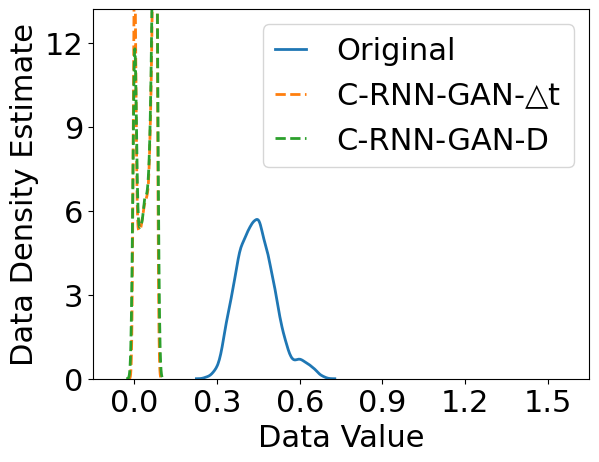}}\hfill
    {\includegraphics[width=0.25\columnwidth]{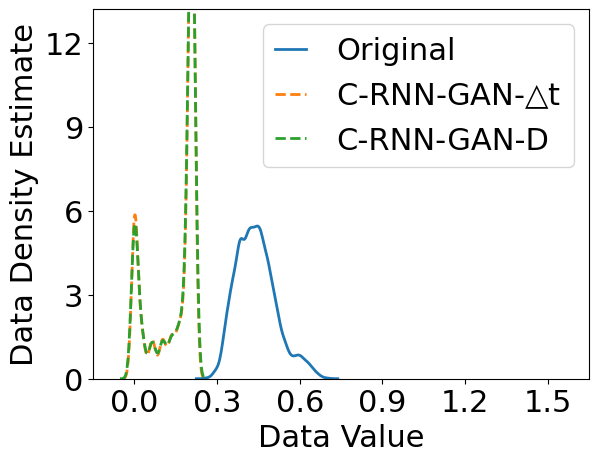}}\hfill
    \newline
    \centering
    {\includegraphics[width=0.25\columnwidth]{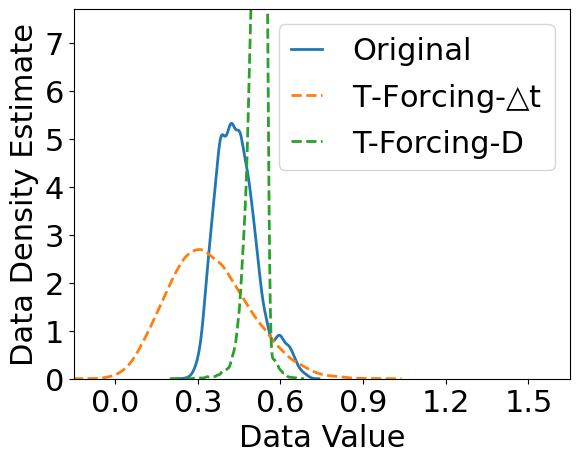}}\hfill
    {\includegraphics[width=0.25\columnwidth]{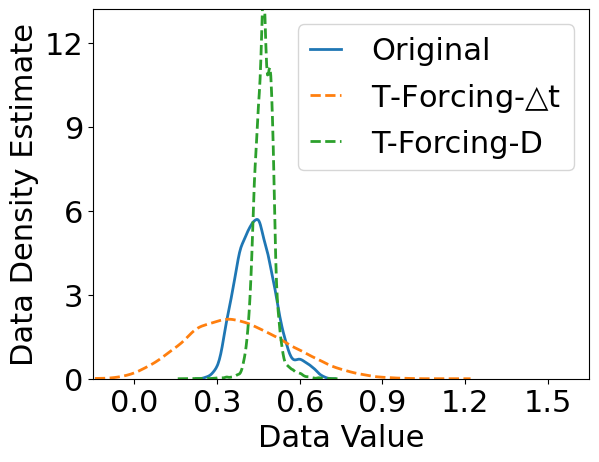}}\hfill
    {\includegraphics[width=0.25\columnwidth]{images/energy_0.7T-Forcing-D_model_histo.png}}\hfill
    \newline
    \subfigure[30\%]{\centering\includegraphics[width=0.25\columnwidth]{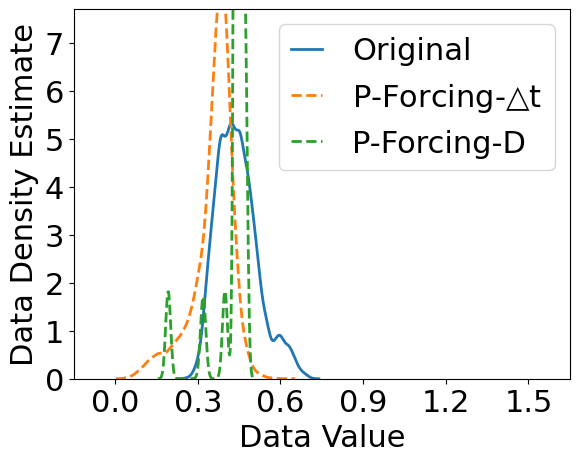}}\hfill
    \subfigure[50\%]{\centering\includegraphics[width=0.25\columnwidth]{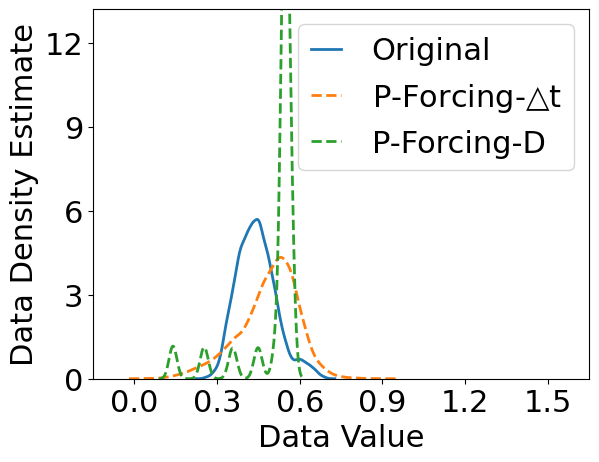}}\hfill
    \subfigure[70\%]{\centering\includegraphics[width=0.25\columnwidth]{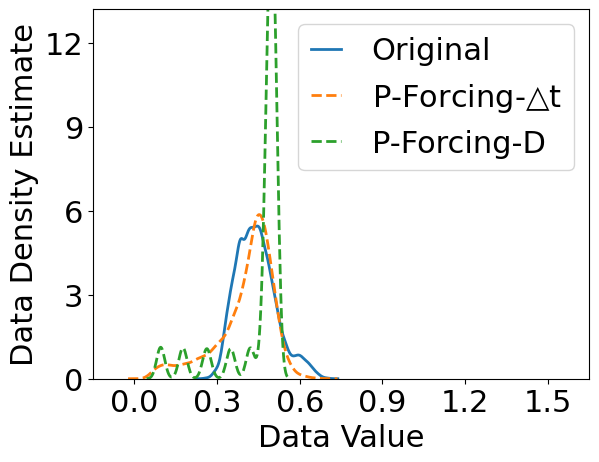}}\hfill
    \caption{Distributions of the Energy data (the 1$^{\text{st}}$ column is for a dropping rate of 30\%, the 2$^{\text{nd}}$ column for a rate of 50\%, and the 3$^{\text{rd}}$ column for a rate of 70\%)}
    \label{fig:histo_energy}
\end{figure}

\begin{figure}[ht]
    \centering
    {\includegraphics[width=0.25\columnwidth]{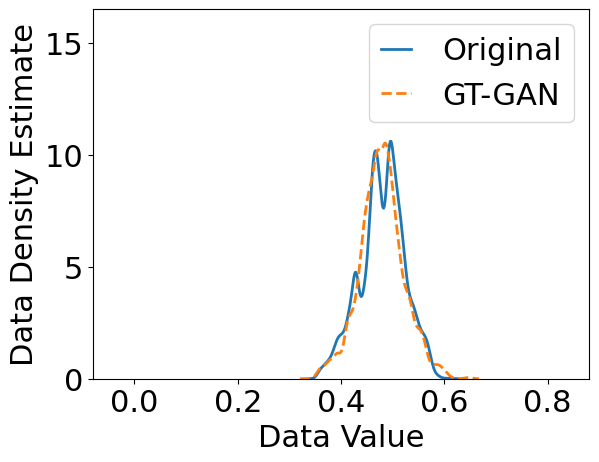}}\hfill
    {\includegraphics[width=0.25\columnwidth]{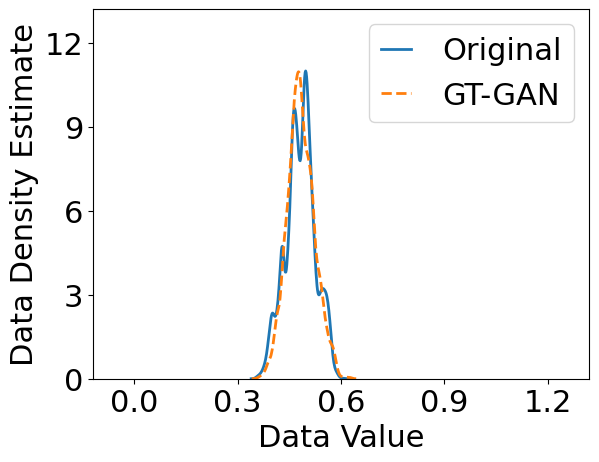}}\hfill
    {\includegraphics[width=0.25\columnwidth]{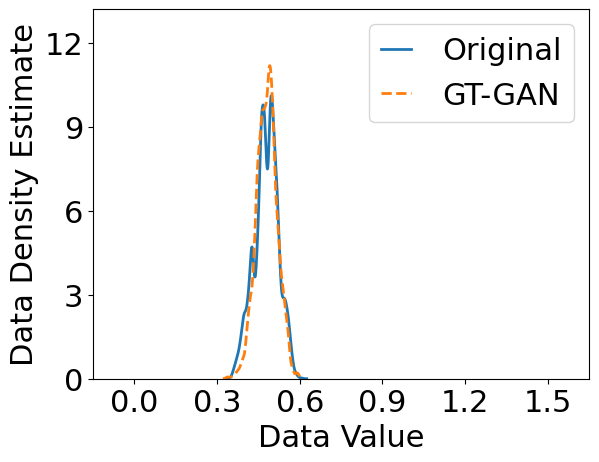}}\hfill
    \newline
    \centering
    {\includegraphics[width=0.25\columnwidth]{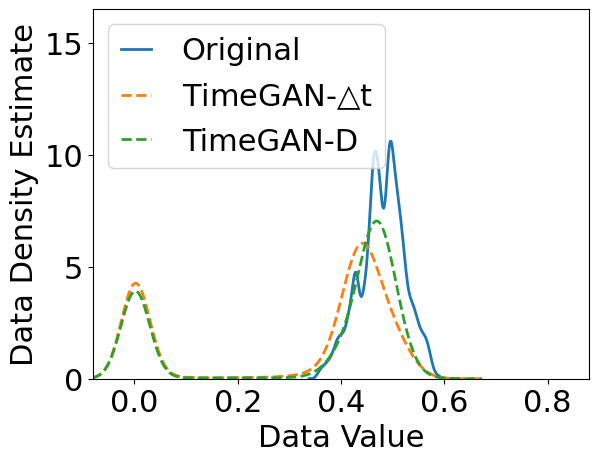}}\hfill
    {\includegraphics[width=0.25\columnwidth]{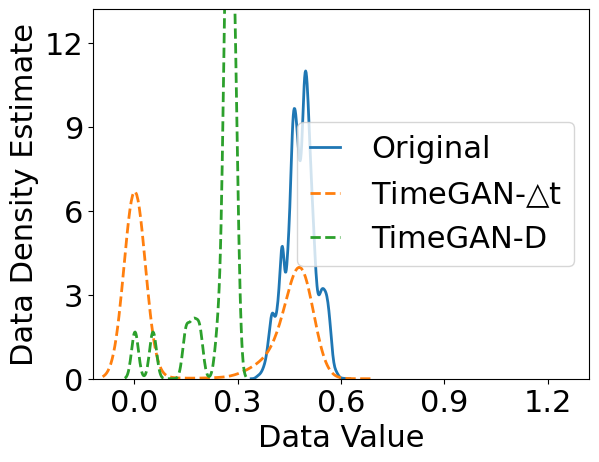}}\hfill
    {\includegraphics[width=0.25\columnwidth]{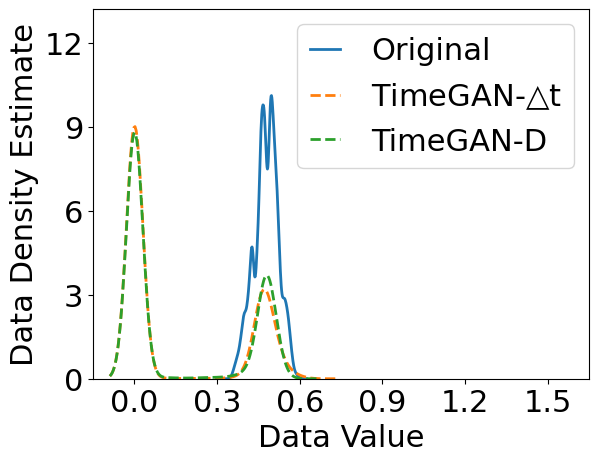}}\hfill
    \newline
    \centering
    {\includegraphics[width=0.25\columnwidth]{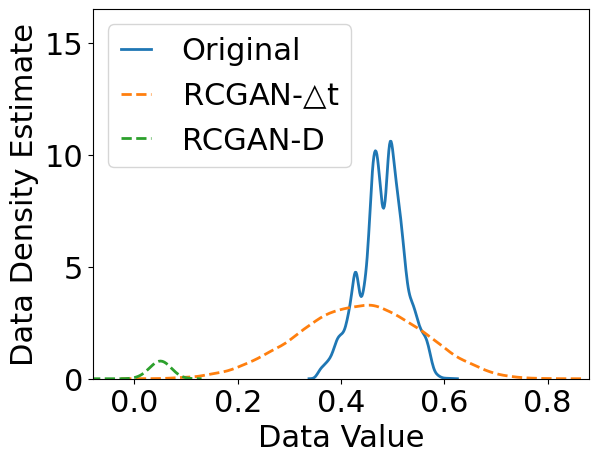}}\hfill
    {\includegraphics[width=0.25\columnwidth]{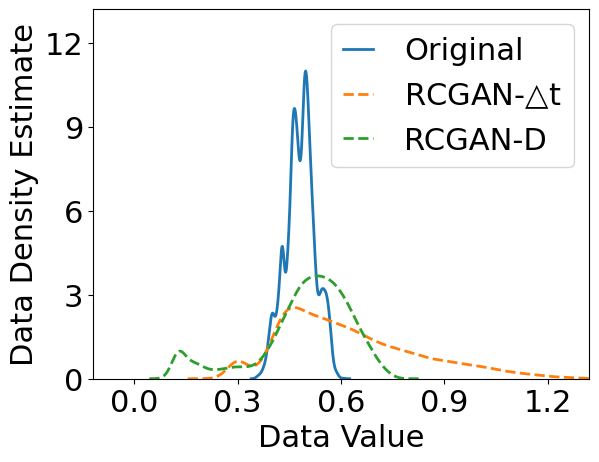}}\hfill
    {\includegraphics[width=0.25\columnwidth]{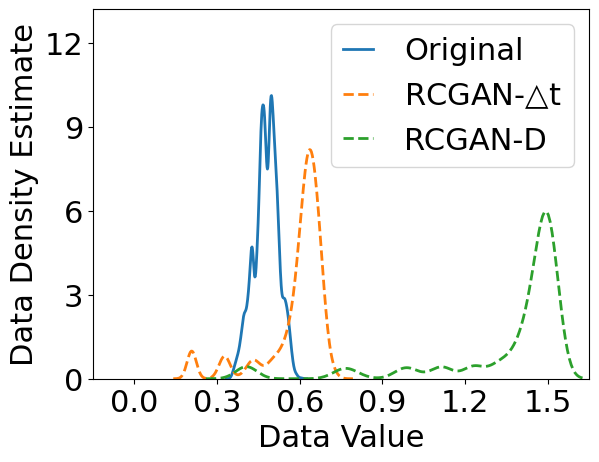}}\hfill
    \newline
    \centering
    {\includegraphics[width=0.25\columnwidth]{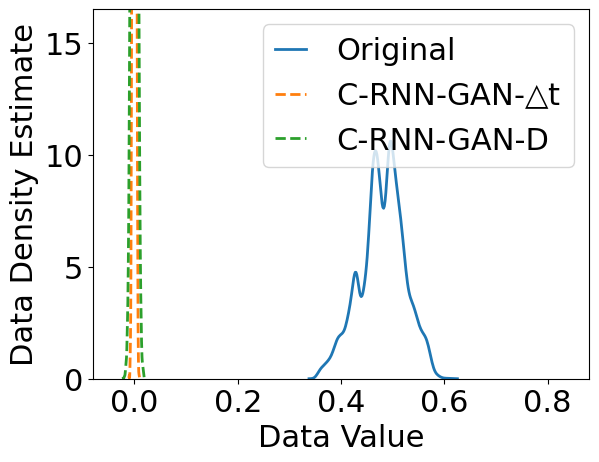}}\hfill
    {\includegraphics[width=0.25\columnwidth]{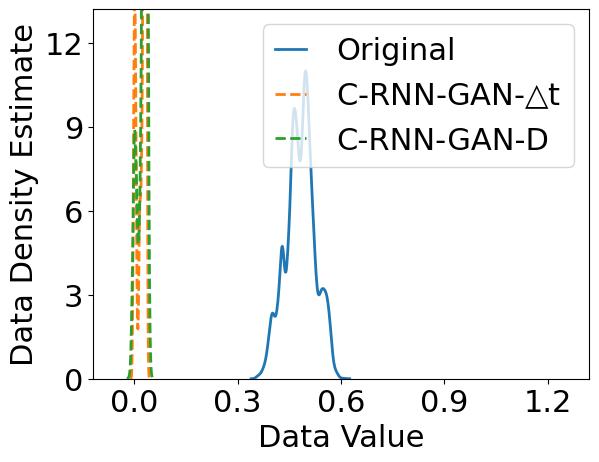}}\hfill
    {\includegraphics[width=0.25\columnwidth]{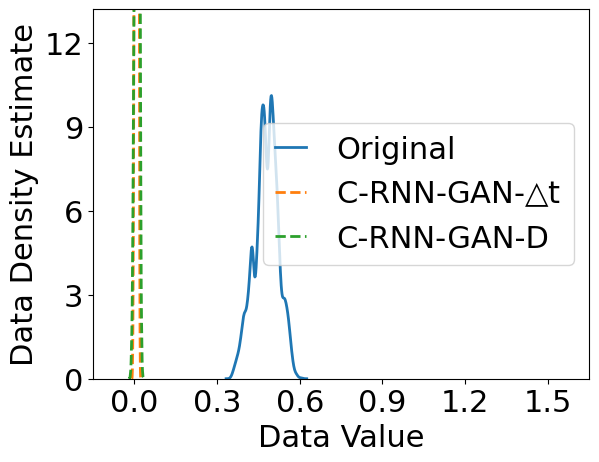}}\hfill
    \newline
    \centering
    {\includegraphics[width=0.25\columnwidth]{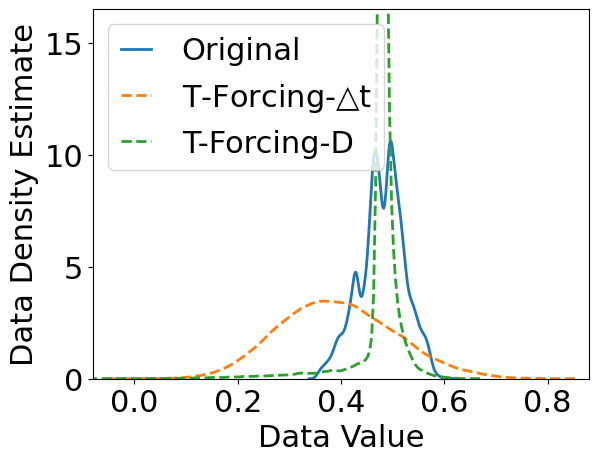}}\hfill
    {\includegraphics[width=0.25\columnwidth]{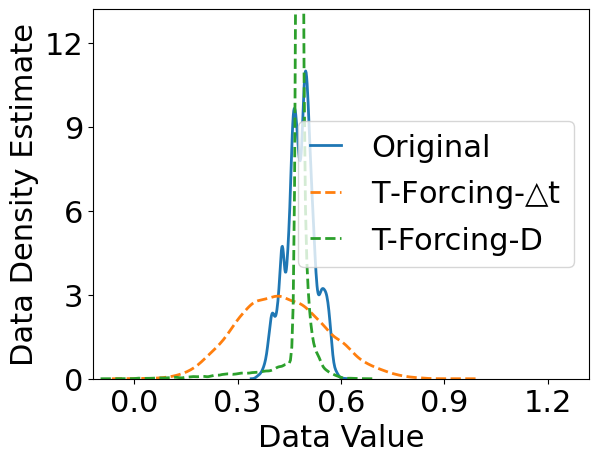}}\hfill
    {\includegraphics[width=0.25\columnwidth]{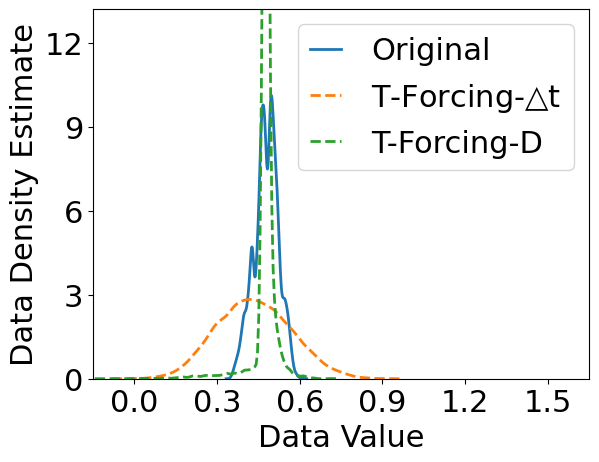}}\hfill
    \newline
    \subfigure[30\%]{\centering\includegraphics[width=0.25\columnwidth]{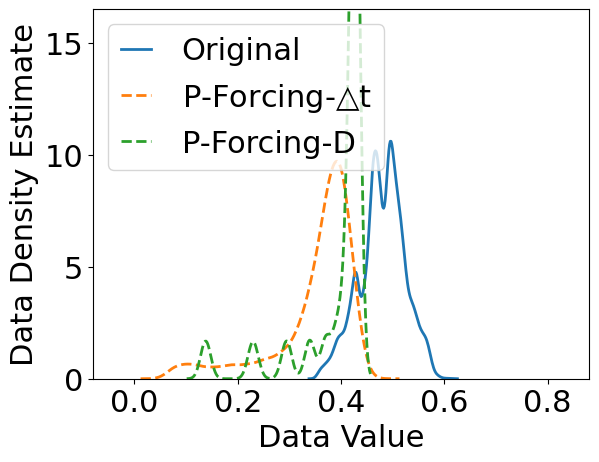}}\hfill
    \subfigure[50\%]{\centering\includegraphics[width=0.25\columnwidth]{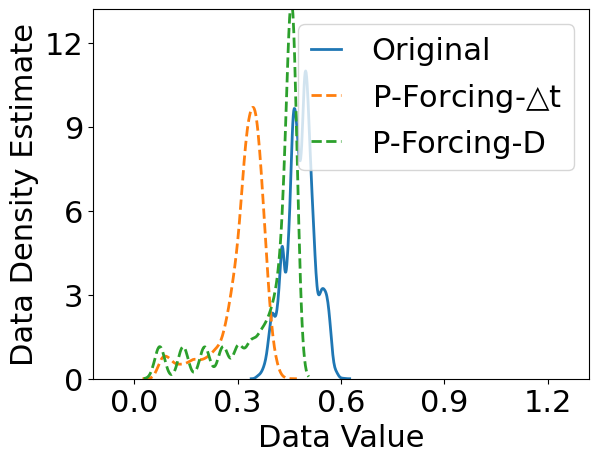}}\hfill
    \subfigure[70\%]{\centering\includegraphics[width=0.25\columnwidth]{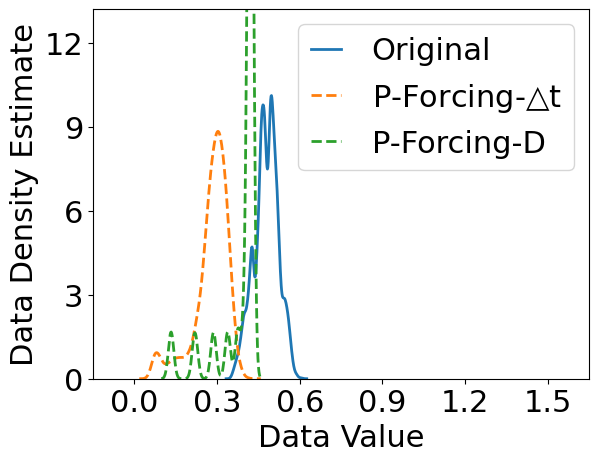}}\hfill
    \caption{Distributions of the MuJoCo data (the 1$^{\text{st}}$ column is for a dropping rate of 30\%, the 2$^{\text{nd}}$ column for a rate of 50\%, and the 3$^{\text{rd}}$ column for a rate of 70\%)}
    \label{fig:histo_mujoco}
\end{figure}

We introduce additional visualization outcomes in Figs.~\ref{fig:tsne_sines} to~\ref{fig:histo_mujoco}.




\clearpage

\section{Algorithm}
\begin{algorithm}[ht]
\caption{How to train \texttt{GT-GAN}}\label{alg:train}
\KwIn{Pre-train iteration number $K_{AE}$, Joint-train iteration number $K_{JOINT}$, MLE train period $P_{MLE}$, Encoder $\boldsymbol{\theta}_f$, Decoder $\boldsymbol{\theta}_g$, Generator $\boldsymbol{\theta}_r$, and Discriminator $\boldsymbol{\theta}_q$ }
Initialize $\boldsymbol{\theta}_f$, $\boldsymbol{\theta}_g$, $\boldsymbol{\theta}_r$ and $\boldsymbol{\theta}_q$\;
$k \gets 0$ \;
\While {$k < K_{AE}$}{
    $\mathbf{h}_{real}$ $\gets$ Encoder($\mathbf{x}_{real}$;$\boldsymbol{\theta}_f$)\;
    $\hat{\mathbf{x}}_{real}$ $\gets$ Decoder($\mathbf{h}_{real}$;$\boldsymbol{\theta}_g$)\;
    Update $\boldsymbol{\theta}_f$ and $\boldsymbol{\theta}_g$ with $\| \mathbf{x}_{real} - \hat{\mathbf{x}}_{real}\|^{2}$\label{alg:ae}\;
    $k$ $\gets$ $k+1$\;
}
$k \gets 0$ \;
\While {$k < K_{JOINT}$}{
        $\mathbf{h}_{real}$ $\gets$ Encoder($\mathbf{x}_{real}$;$\boldsymbol{\theta}_f$)\;
        $\hat{\mathbf{x}}_{real}$ $\gets$ Decoder($\mathbf{h}_{real}$;$\boldsymbol{\theta}_g$)\;
        Update $\boldsymbol{\theta}_f$ and $\boldsymbol{\theta}_g$ with $\| \mathbf{x}_{real} - \hat{\mathbf{x}}_{real}\|^{2}$\;
        \If{$k \mod P_{MLE} \equiv 0$}{\label{alg1:1}
        $\hat{\mathbf{z}}$ $\gets$ $\text{Generator}^{-1}$($\mathbf{h}_{real}, \boldsymbol{\theta}_r$) \label{alg1:2}\;
        $\hat{\mathbf{h}}_{real}$ $\gets$ Generator($\hat{\mathbf{z}}, \boldsymbol{\theta}_r$) \label{alg1:3}\;
        Update $\boldsymbol{\theta}_r$ with $-\log \Pr(\hat{\mathbf{h}}_{real})$\label{alg1:4}\;
        }
        $\mathbf{h}_{fake}$ $\gets$ Generator($\mathbf{z}, \boldsymbol{\theta}_r$)\;
        $\mathbf{x}_{fake}$ $\gets$ Decoder($\mathbf{h}_{fake}, \boldsymbol{\theta}_g$)\;
        Update $\boldsymbol{\theta}_r$ and $\boldsymbol{\theta}_q$ with the adversarial loss with Discriminator($\mathbf{x}_{fake}, \mathbf{x}_{real}, \boldsymbol{\theta}_q$)\;
        $k \gets k + 1$\;
}
\end{algorithm}

We describe the training method in Alg.~\eqref{alg:train}. We first pre-train the autoencoder in the first while loop, followed by the second while loop for the main training step. The main training step consists of i) fine-tuning the autoencoder, ii) training the generation with the log-density loss, iii) training the GAN part with the adversarial loss. 

\section{Efficacy of the log-density training}
In order to see the efficacy of the log-density training, we conduct two more studies. The first model GT-GAN (w/o Eq.~\eqref{eq:for}) is a model in which the generator is trained only with adversarial loss. The second model GT-GAN (supervised loss) replacing the log-density loss to a supervised loss. To obtain the supervised loss, like TimeGAN, we added a supervisor network between the encoder and decoder.
\begin{table}[hbt!]
\begin{minipage}{.99\linewidth}
\centering
\small
\setlength{\tabcolsep}{1pt}
\caption{\label{tab:additional_ablation}Ablation study for log-likelihood training}
\begin{tabular}{|c|c|c|}
\hline
Stocks (Regular) & Discriminative Score & Predictive Score \\ \hline
GT-GAN & .077 & .040 \\ \hline
GT-GAN (w/o Eq.~\eqref{eq:for}) & .159 & .043 \\ \hline
GT-GAN (supervised loss) & .124 & .037 \\ \hline
\end{tabular}
\end{minipage}
\end{table}

According to the above results, it was confirmed that even if TimeGAN's supervised loss is used, no better results than those of our original design are obtained (the predictive score is slightly improved though). In other words, this experiment confirms the importance of the log-density path in our model.

\section{Efficacy of the NCDE-based encoder}
We execute two experiments to justify using an NCDE-based encoder. First, we experiment by replacing the encoder of TimeGAN with our NCDE-based encoder. Second, the NCDE-based encoder of GT-GAN is changed to GRU-$\bigtriangleup t$. The results are in Tables~\ref{tab:encoder_ablation_regular} and~\ref{tab:encoder_ablation_irregular}. Our model shows the best outcomes when we use the NCDE-based encoder.

\begin{table*}[t]
\begin{minipage}{.5\linewidth}
\centering
\scriptsize
\setlength{\tabcolsep}{1pt}
\captionof{table}{\label{tab:encoder_ablation_regular}Regular time series}
\begin{tabular}{|c|c|c|}
\hline
Stocks (Regular) & Discriminative Score & Predictive Score \\ \hline
GT-GAN & .077 & .040 \\ \hline
TimeGAN (NCDE) & .183 & .036 \\ \hline
GT-GAN (GRU-$\bigtriangleup t$) & .184 & .041 \\ \hline
\end{tabular}
\end{minipage}
\begin{minipage}{.5\linewidth}
\centering
\scriptsize
\setlength{\tabcolsep}{1pt}
\captionof{table}{\label{tab:encoder_ablation_irregular}Irregular time series (30\% dropped)}
\begin{tabular}{|c|c|c|}
\hline
Stocks (30\% dropped) & Discriminative Score & Predictive Score \\ \hline
GT-GAN & .077 & .021 \\ \hline
TimeGAN (NCDE) & .430 & .036 \\ \hline
GT-GAN (GRU-$\bigtriangleup t$) & .345 & .022 \\ \hline
\end{tabular}
\end{minipage}
\vspace{-0.7em}
\end{table*}

\section{Role of each network}\label{a:role}
Although our model looks complicated, we use an appropriate network for each part to fit its role. The role of each part is as follows:

\textbf{Encoder} The neural CDE-based encoder is able to encode a regular/irregular time series sample into a regular/irregular hidden vector sequence. Neural CDEs are sometimes called continuous RNNs and are specially designed for the representation learning of irregular time series. As reported in our first email, generation quality is severely degraded when this network is substituted with GRU-$\bigtriangleup t$.

\textbf{Decoder} The GRU-ODE-based decoder is able to decode a regular/irregular hidden vector sequence into a regular/irregular time series sample. One beauty of this decoder is, as shown in Fig.~\ref{fig:sampling} in our main paper, that the sampling time point and the sample length can be freely determined by users.

\textbf{Generator} The CTFP-based invertible generator was intentionally selected by us since we can perform both the log-likelihood and the adversarial training together. Since this network is a key part of our model, we wanted to use the two different training paradigms. Our ablation studies about the log-likelihood and supervised-learning training in Table.~\ref{tab:additional_ablation} justify our design selection.

\textbf{Discriminator} The GRU-ODE-based discriminator is able to process regular/irregular time series. Unlike the encoding task of the neural CDE-based encoder, we observed faster and better results with the GRU-ODE-based discriminator. Moreover, neural CDEs require interpolation of input as a pre-processing. We can do this for real data before training. However, it is hard to perform dynamically for the fake hidden vector sequence due to its excessive computation amount. In particular, it significantly delays the overall training process if we use a neural CDE-based discriminator.

In general, our key design points lie in utilizing i) the continuous-time method-based autoencoder, and ii) the CTFP-based generator. Therefore, we can stabilize the generating performance for complicated irregular time series as well.

\begin{wrapfigure}{r}{6cm}
\vspace{-1.5em}
\centering
\includegraphics[width=0.28\columnwidth]{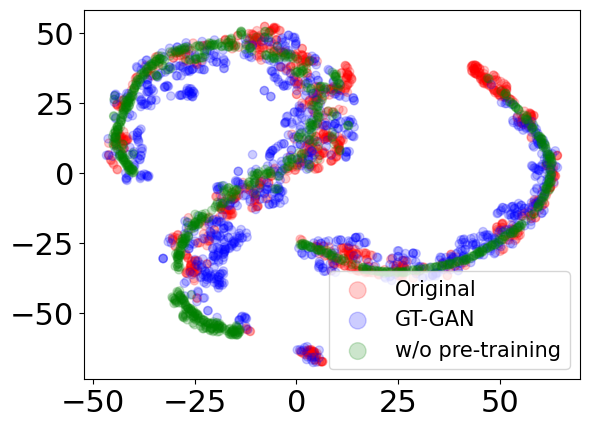}
\caption{\label{fig:ablation_vis}t-SNE visualization of GT-GAN and GT-GAN (w/o pre-training)}
\end{wrapfigure}

\section{Discriminative vs. predictive score}
In Table~\ref{tab:ablation}, GT-GAN without pre-training performs better than GT-GAN in terms of the predictive score. In Fig.~\ref{fig:ablation_vis}, however, GT-GAN generates samples a little out of the original data distribution whereas GT-GAN w/o pre-training has a severe mode-collapse problem (i.e., generating in a narrow region). We conjecture that those samples a little outside the original data distribution make the prediction tasks' scores a little low. However, note that GT-GAN can successfully recall almost the entire data region.

\section{Discussions}\label{a:societal impacts}

\noindent\textbf{Limitations}
Our model shows the best performance in both regular and irregular time series synthesis. However, since our model has a complicated architecture, many hyperparameters exist. Sometimes it is hard to train such large models, which involves a large scale hyperparameter search.

\noindent\textbf{Societal impacts}
Time series data is one of the most widely used data in the field of machine learning. In many cases, time series data carries sensitive personal information, in which case one can use our method to synthesize fake time series and protect privacy. Likewise, we believe that our method has much more positive impacts on our society than negative ones.

\end{document}